\documentclass{article}
\usepackage{ijcai25}
\usepackage{comment}
\usepackage{amsmath}
\usepackage{CJKutf8}
\usepackage{tabularx}
\usepackage{amssymb}
\usepackage{multirow}
\usepackage{graphicx}
\usepackage{array}

\usepackage{times}
\usepackage{soul}
\usepackage{url}
\usepackage[hidelinks]{hyperref}
\usepackage[utf8]{inputenc}
\usepackage[small]{caption}
\usepackage{graphicx}
\usepackage{amsmath}
\usepackage{amsthm}
\usepackage{booktabs}
\usepackage{algorithm}
\usepackage{algorithmic}
\usepackage[switch]{lineno}

\usepackage{multirow}
\usepackage{float}
\usepackage{colortbl}
\usepackage[table]{xcolor}
\usepackage{subfigure}
\usepackage{amssymb}
\usepackage{arydshln}
\usepackage{csquotes}

\title{Always Clear Depth: Robust Monocular Depth Estimation under Adverse Weather}

\author{
Kui Jiang$^1$\and
Jing Cao$^1$\and
Zhaocheng Yu$^1$\and
Junjun Jiang$^1$\footnote{Corresponding Author}\And
Jingchun Zhou$^2$\\
\affiliations
$^1$Harbin Institute of Technology\\
$^2$Dalian Maritime University\\
\emails
jiangkui@hit.edu.com,
\{caojing,yuzhaocheng\}@stu.hit.edu.cn,
jiangjunjun@hit.edu.cn,
zhoujingchun@dlmu.edu.cn
}
\begin{document}

\maketitle

\begin{abstract}
Monocular depth estimation is critical for applications such as autonomous driving and scene reconstruction. While existing methods perform well under normal scenarios, their performance declines in adverse weather, due to challenging domain shifts and difficulties in extracting scene information. 
To address this issue, we present a robust monocular depth estimation method called \textbf{ACDepth} from the perspective of high-quality training data generation and domain adaptation. 
Specifically, we introduce a one-step diffusion model for generating samples that simulate adverse weather conditions, constructing a multi-tuple degradation dataset during training. 
To ensure the quality of the generated degradation samples, we employ LoRA adapters to fine-tune the generation weights of diffusion model. Additionally, we integrate circular consistency loss and adversarial training to guarantee the fidelity and naturalness of the scene contents. 
Furthermore, we elaborate on a multi-granularity knowledge distillation strategy (MKD) that encourages the student network to absorb knowledge from both the teacher model and pretrained Depth Anything V2. 
This strategy guides the student model in learning degradation-agnostic scene information from various degradation inputs. 
In particular, we introduce an ordinal guidance distillation mechanism (OGD) that encourages the network to focus on uncertain regions through differential ranking, leading to a more precise depth estimation. 
Experimental results demonstrate that our ACDepth surpasses md4all-DD by 2.50\% for night scene and 2.61\% for rainy scene on the nuScenes dataset in terms of the absRel metric. Code and data are available at \href{https://github.com/msscao/ACDepth}{https://github.com/msscao/ACDepth}.

\end{abstract}

\section{Introduction}

Monocular depth estimation (MDE) is a fundamental task in computer vision that aims to predict the depth from a single image. It has wide-ranging applications in autonomous driving~\cite{schon2021mgnet,xue2020toward}, robot navigation~\cite{hane2011stereo}, and 3D reconstruction~\cite{yu20143d,yin2022towards}. MDE methods can be roughly divided into supervised methods and self-supervised methods ~\cite{arampatzakis2023monocular}. The former~\cite{bhat2023zoedepth,ranftl2020towards,ranftl2021vision} directly learns depth mapping from the RGB input with paired samples (depth maps or 3D sensors like LiDAR). While achieving impressive performance, these methods heavily rely on high-quality ground truth data, which is time-consuming and labor-intensive for collection. By contrast, the latter~\cite{godard2017unsupervised,godard2019digging,zhou2017unsupervised} uses only image sequences captured by a single camera and the corresponding camera parameters, and employs two assumptions (photometric constancy and rigid motion)~\cite{godard2019digging} to produce the supervised signal. Specifically, these methods assume that scene information remains largely unchanged as the viewpoint shifts, relying on the geometric consistency between consecutive frames to provide depth supervision. While existing self-supervised approaches~\cite{zhao2022monovit,zhang2023lite,lyu2021hr} have shown significant success in outdoor scenes, their performance deteriorates under adverse conditions, such as low-light and rain-haze conditions. In these situations, insufficient lighting and significant motion perturbations (reflections of raindrops and streetlights) violate the assumptions mentioned above, leading to an unreliable estimation.

\begin{figure}[!htp]
	\centering
	\includegraphics[width=1\linewidth]{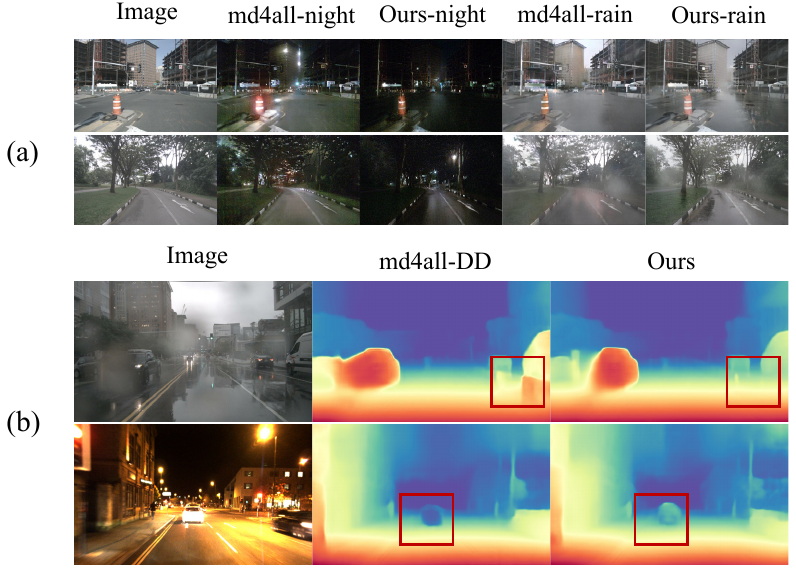}
	\caption{{\textbf{(a) Comparison of training data in challenging scenes:}} Compared to the data generation method used in md4all~\protect\cite{gasperini2023robust}, the samples generated by our approach are more realistic, providing a better simulation of challenging real-world conditions. {\textbf{(b) Estimation results in challenging scenes:}} Our method consistently produces more accurate results than the existing md4all~\protect\cite{gasperini2023robust} method, particularly in handling complex issues such as ground water reflections and nighttime object recognition.}	\vspace{-2mm}
	\label{Fig:1}

\end{figure}
Recent studies~\cite{gasperini2023robust,saunders2023self,tosi2025diffusion} have explored ways to enhance the robustness of models under adverse conditions. Some technologies elaborate modules to promote robustness in specific scenes, such as nighttime and rain-haze  conditions~\cite{shi2023even,zheng2023steps,yang2024self}. However, due to the limitation for scene-specific representation, these methods struggle to generalize to complex and diverse environments, frequently encountered in real-world applications. 
To promote generalization, some efforts harmonize the merits of distillation learning, contrastive learning, and data augmentation strategies to create more generalized models that improve model performance across varied scenes~\cite{wang2024digging,wang2024weatherdepth,saunders2023self}. 
For example, md4all~\cite{gasperini2023robust} generates the [normal, degraded] sample pairs with a GAN-based model and uses distillation learning to extract supervised signals from clear depth maps to train the student model. 
However, the degraded characteristics in the translated image are unrealistic and unnatural (shown in Figure.~\ref{Fig:1}(a)), which significantly affects the generalization from normal scenarios to real adverse weather. 
In addition, existing methods are often vulnerable to disturbances in complex scenes due to slack constraints between the teacher and student models, leading to incomplete knowledge transfer. 
Consequently, as shown in Figure.~\ref{Fig:1}(b), md4all 
suffers from nonnegligible performance decline when there are significant domain differences between scenes, making it far from generalizing to various degraded scenarios. 

Overall, two critical issues are imperative to train a generalizable and reliable depth estimation model: (1) the lack of a high-quality multi-tuple dataset covering diverse degradation types; (2) slack constraints between the teacher and student model, resulting in incomplete knowledge transfer. 

To mitigate the problem of data scarcity, the author in~\cite{tosi2025diffusion} utilize depth maps and text descriptions as control conditions to generate [normal, degraded] sample pairs with T2I-Adapter~\cite{mou2024t2i}. 
This provides an alternative solution with diffusion models for multi-tuple dataset generation~\cite{zhu2017unpaired,parmar2024one,sauer2025adversarial}. 
However, the translated image shows obvious inconsistencies and less authenticity of scene contents, which is unacceptable for training generalizable depth estimation models. 

To address these issues, we propose to optimize the data generation and domain adaptation learning, and construct the degradation-agnostic robust monocular depth estimation method (ACDepth). 
Specifically, we fully explore the potential generation capability of diffusion model and employ LoRA adapters to encourage the network to generalize to diverse scene generation. Meanwhile, we integrate circular consistency loss and adversarial training to guarantee the naturalness and consistency of translated images. 
To achieve full transfer and alignment of capabilities, we propose the multi-granularity knowledge distillation strategy (MKD), which borrows priors from the teacher model to provide comprehensive supervision and guidance to the student model. 
Besides the commonly used feature and result distillation learning, we pioneer the ordinal guidance distillation mechanism (OGD).  
In summary, the contributions are as follows: 
\begin{itemize}
    \item We propose a practical multi-tuple degradation dataset generation scheme, and develop a novel robust monocular depth estimation framework, termed as  ACDepth for high-quality depth estimation under adverse weather. 

    \item We propose the multi-granularity knowledge distillation strategy (MKD) to achieve the complete transfer and alignment of capabilities from the teacher model to student model. In addition, we introduce the ordinal guidance distillation mechanism (OGD) to heartens the network to focus on uncertain regions through differential ranking. 

    \item Extensive experiments demonstrate the effectiveness of ACDepth, surpassing md4all-DD by 2.50\% for night scene and 2.61\% for rainy scene on the nuScenes dataset in terms of absRel metric.
\end{itemize}

\section{Related work}

\subsection{Monocular Depth Estimation}

Before the advent of deep neural networks, traditional depth estimation methods primarily rely on handcrafted priors to explore the limited physical and geometric properties. 
Deep learning methods~\cite{sun2012depth,liu2015learning,liu2015deep} have emerged as a preferable alternative due to their ability to learn generalizable priors from large-scale data, such as depth maps from LiDAR or RGB-D cameras.
arning. 
However, due to the high cost to obtain the high-quality annotation, 
self-supervised learning technologies, deriving depth information from stereo pairs~\cite{garg2016unsupervised,godard2017unsupervised} or video sequences~\cite{zhou2017unsupervised,godard2019digging,bian2019unsupervised} have drawn growing interest. 
Unfortunately, a significant portion of the effort is focused on the normal scene. The depth estimation under adverse weather, such as low-light and rain-haze
conditions, is barely explored, which is the common scenario in  autonomous driving.  
Some studies
~\cite{zheng2023steps,wang2021regularizing,guo2020zero} divide depth estimation in adverse scenes into denoising and estimation, but these methods show poor generalization under unknown degradations. 
Further, 
researchers achieve more robust depth estimation through data generation and knowledge distillation~\cite{zhu2023ec,gasperini2023robust}, and integrate multi-level contrastive learning with diffusion model for robust feature representation~\cite{wang2024digging}. 
However, the data quality and completeness of constraints in these methods significantly affect the generalization from normal scenarios to real adverse weather.

\subsection{Distillation Learning}
Early distillation learning methods primarily focus on model compression and acceleration. A classic example 
of distillation learning is to guide the training of student models by softening the output probability distribution of the teacher model~\cite{hinton2015distilling}. 
Subsequently, several studies have successfully applied distillation learning to monocular depth estimation. 
For example, 
Song et al.~\cite{song2022learning} selectively distill stereo knowledge for monocular depth estimation, using learned binary masks to pick the best disparity or pixel-wise depth map. More recently, md4all~\cite{gasperini2023robust} trains the teacher model on clear samples by self-supervised learning, and transfers the ability or priors to the student model under adverse weather. 
However, during the distillation process, the student model struggles to fully 
reproduce the ability of the teacher model, relying solely on depth-derived pseudo-labels or specific priors. 
To address these challenges, we propose a multi-granularity knowledge distillation strategy that enhances the knowledge transfer process by borrowing priors from multiple teachers to provide comprehensive supervision and guidance. 
\section{Method}

\begin{figure*}[!htbp]
    \hsize=\textwidth
    \centering
    \includegraphics[width=0.95\linewidth]{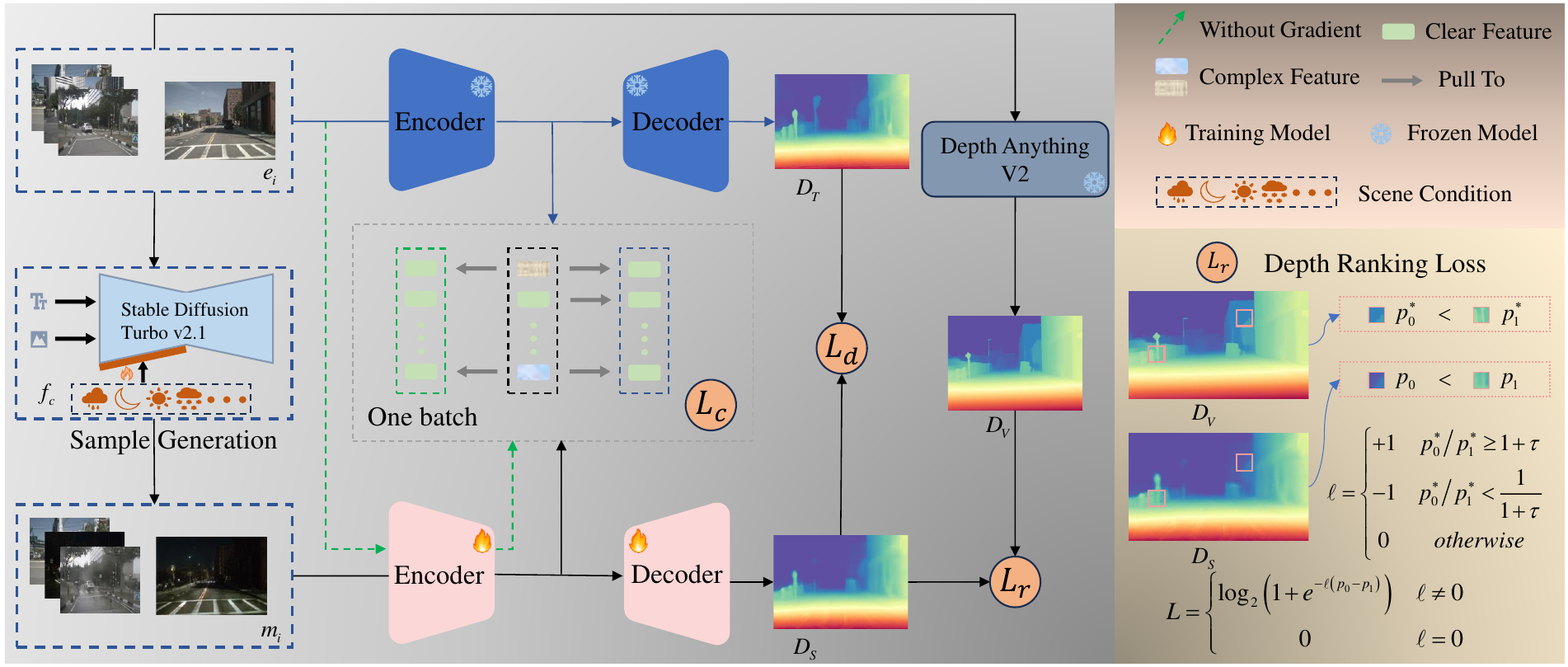}
    \caption{
    {\textbf{Overview of our ACDepth for robust monocular depth estimation}.The teacher model is trained on simple samples using self-supervised learning, and the student model is trained on a mixed dataset of simple and complex samples using distillation learning. To provide the student model with supervisory signals beyond those from the teacher model, we designed a depth ranking loss $L_r$ leveraging ordinal information from the Depth Anything V2 model. To improve the student model's generalization across diverse scenarios, we incorporated a feature constraint loss $L_c$.}\vspace{-3mm}
    }
    \label{Fig:2}
\end{figure*}

\subsection{Preliminary}
MDE aims to estimate the depth map $D_t$ from a single RGB image $I_t$. Restricted by high-quality paired samples in real-world scenarios, especially for adverse weather, the self-supervised learning strategy~\cite{zhu2023ec,wang2024digging} provides an alternative solution to obtain supervised signals of the target image $I_{t}$ from adjacent frames $I_{t^\prime}\in[I_{t-1},I_{t+1}]$. 
Specifically, the pose network and depth network are jointly optimized. The former  
estimates the relative pose of camera motion $T_{t\rightarrow t^\prime}$, from the target image to the adjacent frame ($I_{t^\prime}$). 
Combining with the intrinsic parameters $K$ of camera, it allows the network 
to synthesize 
the reconstruction image ($I_{t^\prime\rightarrow t}$) of $I_{t}$ 
using the adjacent frame $I_{t^\prime}$, depicted as:
\begin{equation}
I_{{t^\prime} \to t} = {I_{{t^\prime}}}\left\langle {proj\left( {{D_t},{T_{t \to {t^\prime}}},K} \right)} \right\rangle, 
\end{equation}
where the 
$\left\langle \right\rangle $ 
denotes the pixel sampling operator, and we constrain the depth by calculating the photometric reconstruction loss between $I_t$ and $I_{t^\prime\rightarrow t}$, formulated by:  
\begin{equation}
L_p=\min_{t^\prime}{pe\left(I_t,I_{t^\prime\rightarrow t}\right)},
\end{equation}
\begin{equation}
pe\left(I_a,I_b\right)=\frac{\theta}{2}\left(1-SSIM\left(I_a{,I}_b\right)\right)+\left(1-\theta\right)I_a-I_b,
\end{equation}
where SSIM is the structural similarity index measure. 
Compared with the pixel-wised L2 loss, it can better reflect the structural similarity between images. 
Additionally, edge-aware smoothness loss is also used to constrain the continuity of the depth:
\begin{equation}
L_e\left(D\right)=\left|\partial_xD'\right|e^{\partial_xI}+\left|\partial_yD'\right|e^{\partial_yI},
\end{equation}
where $D'$ refers to the normalized inverse depth of $D$. $\partial_x$ and $\partial_y$ represent the horizontal and vertical gradients. 
Similar to~\cite{gasperini2023robust}, the above theoretical foundations of self-supervised learning are used to train the teacher model  in this work. 

\subsection{Overall Architecture} 
Our ultimate goal is to achieve the robust MDE under adverse weather. Thus, we elaborate an ACDepth approach, as shown in Figure.~\ref{Fig:2}, which involves the data generation and robust model training.
In the first part, we utilize the LoRA adapters to fine-tune the pretrained diffusion model, where the circular consistency loss and adversarial training are used to guarantee the naturalness and consistency of the translated image. In this way, the trained generator can produce a multi-tuple degradation dataset under different weather conditions.  
And then, ACDepth takes the multi-tuple degradation dataset as input, and proposes the multi-granularity knowledge distillation strategy (MKD) to borrow priors from the teacher model to optimize the student network. The process is depicted as:
\begin{equation}
L=L_d+{\lambda_1L}_r+{\lambda_2L}_c,
\end{equation}
where $L_d$ denotes the distillation loss between the teacher and student models, $L_r$ refers to the ordinal guidance distillation between the Depth Anything V2 model and our ACDepth,  
and $L_c$ is the feature consistency loss between the teacher encoder and the student encoder. 
$\lambda_1$ and $\lambda_2$ are the weight parameters to balance the loss components. Detailed explanation of losses ($L_d$, $L_r$ and $L_c$) is provided below. 

\subsection{Data Generation}
Given a normal input ($e_i$), which is captured under good lighting and visibility conditions, the previous studies employ GAN-based or diffusion-based models to generate the degraded images ($h_i^c$). $c$ represents various weather scenes, such as rain, night, and fog. However, besides requiring a large number of real multi-tuple pairs for training, the translated images generated by these technologies~\cite{gasperini2023robust,tosi2025diffusion} exhibit obvious differences and unnaturalness from the real samples, leading to poor generalization of depth estimation models across different scenes. 

Inspired by recent image translation technologies~\cite{parmar2024one}, we explore the content generation capability of stable diffusion, while employing adversarial learning and cycle consistency loss to train LoRA~\cite{hu2021lora} adapters to promote the naturalness and consistency of translated images.
Specifically, we take $e_i$ as input, along with the corresponding text prompt $P_c$ to learn the specific LoRA adapters for each scene transformation, which can complete the conversion from the source domain to the target domain. 
The aforementioned process is depicted as:
\begin{equation}
h_i^c=F_c (SDT(e_i,P_c )),
\end{equation}
where $SDT(\cdot)$ refers to the Stable Diffusion Turbo model, and $F_c(\cdot)$ is the translator that transforms the normal sample into the corresponding adverse scene with the condition $c$ and text prompt $P_c$. 
Detailed experiments related to data generation can be found in the appendix.

\subsection{Robust Model Training}

To achieve full transfer and alignment of capabilities between the teacher model and student model, we propose the multi-granularity knowledge distillation strategy (MKD) to achieve the robust model training  of the student model, detailed as follows. 
\paragraph{Distillation Learning.}
Similar to the existing technology~\cite{gasperini2023robust}, the commonly used multi-scale feature distillation loss between the teacher model ($F_T(\cdot)$) and student model ($F_S(\cdot)$) is employed to facilitate the perception of the student model to the adverse weather. For the given normal input $e_i$ and the produced degraded sample ($h_i^c$), the distillation loss function is defined as:
\begin{equation}
L_d=\frac{1}{S}\sum_{s=1}^{S}{\frac{1}{N_s}\sum_{j=1}^{N_s}\frac{\left|{F_T\left(e_i\right)}_{js}-{F_S\left(m_i\right)}_{js}\right|}{{F_S\left(m_i\right)}_{js}}},
\end{equation}
where $S$ denotes the number of different scales and $m_i$ represents a sample randomly selected from the mixed training set. 

\paragraph{Ordinal Guidance Distillation.}
1) Uncertain Region Definition: We use the output from Depth Anything V2 model as the supervisory signal because it ensures accurate depth prediction while maintaining acceptable inference speed. $D_T$ and $D_S$ represent the inverse depth of the teacher model and student model, respectively, and $D_v$ represents the output depth of Depth Anything V2. We first identify the regions with significant discrepancies between $D_T$ and $D_S$ for refinement:
\begin{equation}
\bar{D}=\left|D_T-D_S\right|,
\end{equation}
where $\bar{D}$ reflects the difference between the output of the teacher model and the student model. We normalize the depth difference $\bar{D}$ to obtain $\hat{D}$, which takes values between 0 and 1. We define $U$ as the region of focus for the loss, and the region of focus is determined by the following equation:
\begin{equation}
U=\left\{\begin{matrix}1&d_i>\gamma\\0&d_i\le\gamma,\\\end{matrix}\right.
\end{equation}
where $\gamma$ is the threshold for dividing the key regions, set to 95\% of $\hat{D}$. These key regions correspond to areas with impaired perception in challenging scenes, as shown in Fig 3(e).

2) Depth Ordinal Strategy: We sample a pixel from $U \ast D_S$, with the corresponding depth value $p_0$, and another pixel from $\left(\sim U\right)\ast D_S$, with the corresponding depth value $p_1$. The ranking loss~\cite{xian2020structure,sun2023sc} for the pair $\left[p_0,p_1\right]$ is computed as follows:
\begin{equation}
\psi\left(p_0,p_1\right)=\left\{\begin{matrix}\log_2{\left(1+e^{-\ell\left(p_0-p_1\right)}\right)}&\ell\neq0\\\ell\left(p_0-p_1\right)^2&\ell=0,\\\end{matrix}\right.
\end{equation}
The ordinal label $\ell$ is calculated as follows:
\begin{equation}
\ell  = \left\{ {\begin{array}{*{20}{c}}
{ + 1}&{\frac{{p_0^ * }}{{p_1^ * }} \ge 1 + \tau }\\
{ - 1}&{\frac{{p_0^ * }}{{p_1^ * }} < \frac{1}{{1 + \tau }}}\\
0&{otherwise,}
\end{array}} \right.
\end{equation}
where $\tau$ is a hyperparameter used to control the selection of pixel pairs for the sorting. $p_0^\ast$ is sampled from the region $U\ast D_v$, and $p_1^\ast$ is sampled from the complementary region $\left(\sim U\right)\ast D_v$. These two pixels, $\left[p_0^\ast,p_1^\ast\right]$, form an ordinal pair.
\begin{figure}[!htp]
	\centering
	\includegraphics[width=0.98\linewidth]{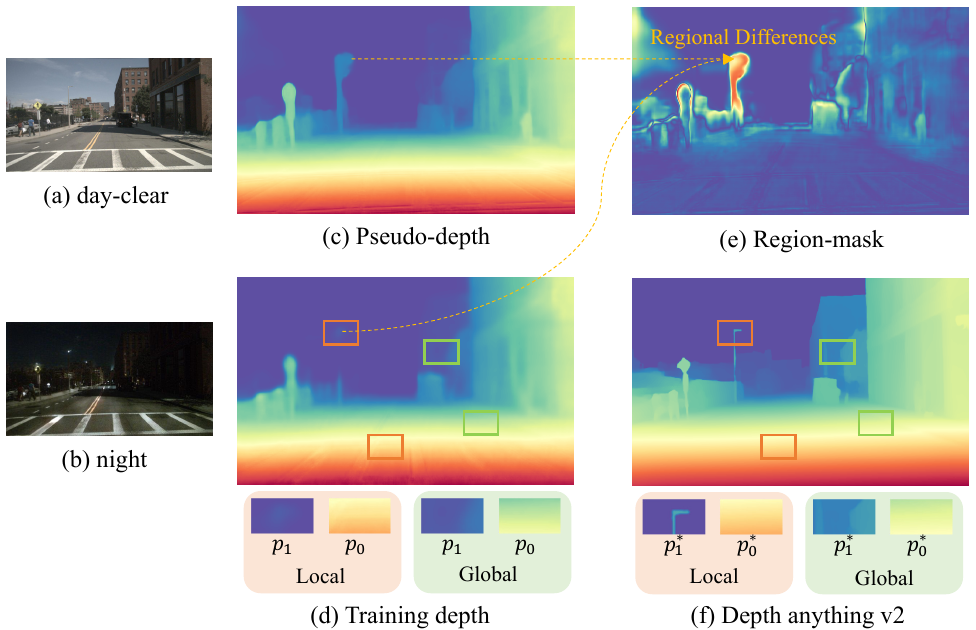}
	\caption{Ordinal Pair Sampling: During training, we use two strategies to efficiently sample ordinal pairs. First, we employ the teacher and student models to respectively predict the depth maps (c and d) from the normal (a) and degraded (b) input, and then compute the pixel-wise errors (e) among them. 
    Based on (e), we sample the local (uncertain regions with large errors) and global (random sampling regions) patches from the depth maps (d) and (f) to compute the total ordinal pair loss. 
    }
	\label{Fig:3}
\end{figure}
The ranking loss aims to enhance the student model's performance by focusing on poorly predicted regions. However, we found that the constraints imposed by the distillation loss sometimes conflicted with the ranking loss. To mitigate this issue, after identifying each ordinal pair, we compute the average depth value of the surrounding 5×5 pixel region for each sample point and use this average value to calculate the ranking loss.  Considering the increase in computational overhead, we control the number of selected points in each iteration by setting $\tau=0.15$ to select fewer sample points. Similarly, we set the proportion of sample pairs selected from $U$ to 0.05. Finally, we randomly sample ordinal pairs from the global depth map by a ratio of 0.01, and the ranking loss calculated from these sampled points is used to refine the global depth information. The sampling process is illustrated in Figure.~\ref{Fig:3}. Thus, our final loss becomes:
\begin{equation}
\begin{split}
L_r=\frac{1}{\left|Z_g\right|}\sum_{p_0,p_1\in Z_g}\psi\left(p_0,p_1\right)+\frac{1}{\left|Z_l\right|}\sum_{p_0,p_1\in Z_l}{\psi\left(p_0,p_1\right),}
\end{split}
\end{equation}
where $Z_g$ and $Z_l$ represent the global and local ordinal sampling sample sets, respectively.

\paragraph{Feature Consistency Constraint.}
Since a robust model can provide the precise perception on both the normal and degraded scenarios, we propose the feature consistency constraint to mitigate the impact of image degradation, thus generalizing to various adverse weather. 
Specifically, we dynamically adjust the feature alignment strategy based on the input image $e_i$. If $e_i$ is the degraded image $h_i^c$, we minimize the error between the feature maps $F_S(h)$ and $F_T(e)$, as well as the error between $F_S(h)$ and $F_S(e)$, where $F_T$ and $F_S$ represent the feature extraction by the teacher and student models, respectively.
If $e_i$ is a clear sample $e$, we minimize the error between the feature maps $F_T(e)$ and $F_S(e)$, respectively extracted by the teacher and student models. This form can be written as:
\begin{equation}
\begin{split}
L_c=\left\{\begin{matrix}f\left(F_S\left(h\right),\bar{F_T\left(e\right)}\right)+f\left(F_S\left(h\right),\bar{F_S\left(e\right)}\right)&if\ \ i=h\\f\left(F_S\left(e\right),\bar{F_T\left(e\right)}\right)&if\ \ i=e,\\\end{matrix}\right.\
\end{split}
\end{equation}
where $f$ represents L1 loss for aligning features, and the horizontal line denotes that no gradient backpropagation is applied. The above process enables the student model to not only acquire semantic features from the teacher model but also progressively improve its feature extraction capabilities across both normal and challenging scenarios, ensuring a seamless transition from clear to complex environments.
\section{Experiments}

\begin{figure*}[!hbt]
\begin{center}
\includegraphics[width=1.00\textwidth]{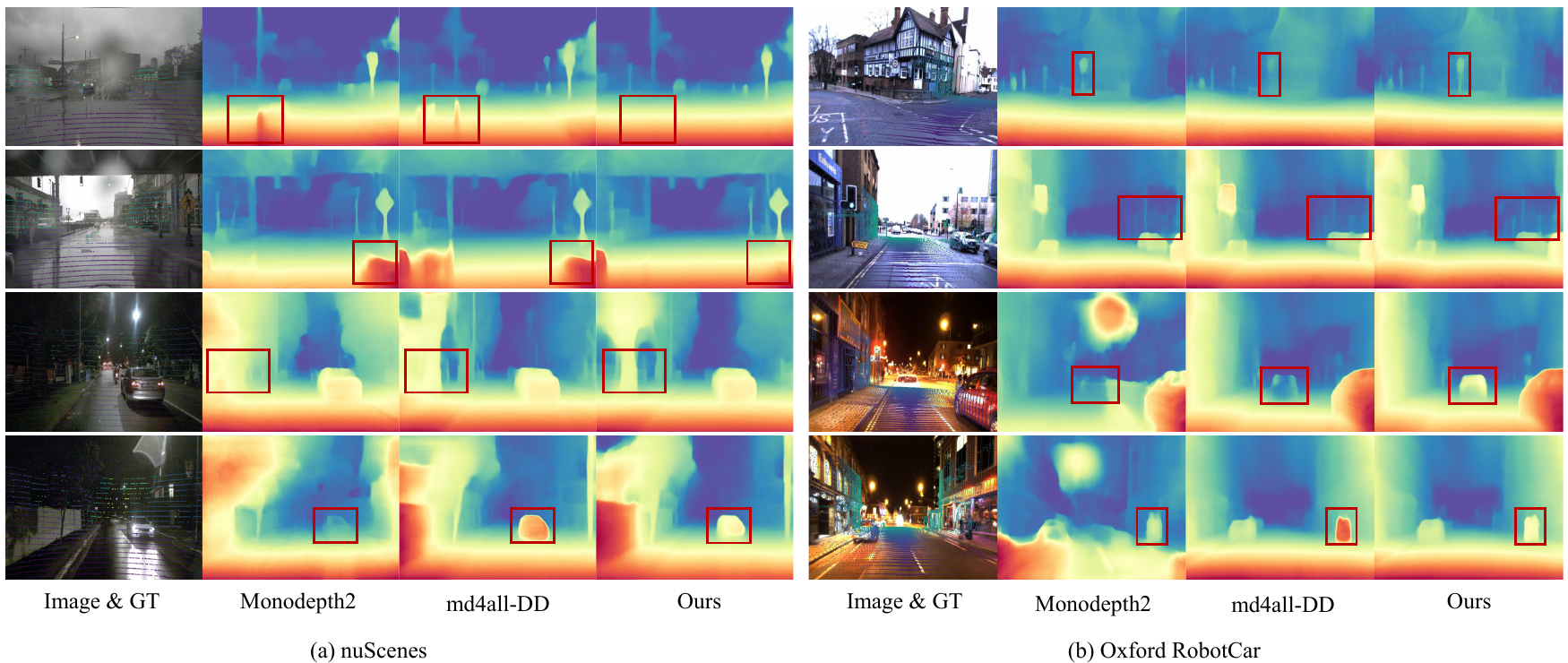}
\end{center}\vspace{-2mm}
   \caption{Qualitative results on nuScenes~\protect\cite{caesar2020nuscenes} and RobotCar~\protect\cite{maddern20171}. We compare the ACDepth approach with md4all-DD and MonoDepth2, all of which use the same backbone. To better illustrate the results, the real point cloud is projected onto the original image, and no ground truth (GT) is required during training.}\vspace{-2mm}
\label{fig:4}
\end{figure*}

\begin{table*}[!htbp]
	\begin{center}
    \scalebox{1.0}{
    	\scriptsize
		\begin{tabular}{lll|ccc|ccc|ccc}
			\toprule
			&&& \multicolumn{3}{c|}{\textit{day-clear} -- nuScenes} & \multicolumn{3}{c|}{\textit{night} -- nuScenes} & \multicolumn{3}{c}{\textit{day-rain} -- nuScenes} \\
			Method & sup. & tr.data & absRel & RMSE & $\delta_1$   & absRel & RMSE & $\delta_1$   & absRel & RMSE & $\delta_1$  \\
			\midrule
			
			Monodepth2~\cite{godard2019digging} & M$^*$ & a: dnr & 0.1477 & 6.771 & \textbf{85.25} & 2.3332 & 32.940 & 10.54 & 0.4114 & 9.442 & 60.58 \\
			Monodepth2~\cite{godard2019digging} & M$^*$ & d & 0.1374 & 6.692 & 85.00 & 0.2828 & {9.729} & 51.83 & 0.1727 & 7.743 & 77.57 \\
			PackNet-SfM~\cite{guizilini20203d} & Mv & d & 0.1567 & 7.230 & 82.64 & 0.2617 & 11.063 & 56.64 & 0.1645 & 8.288 & 77.07\\
			RNW~\cite{wang2021regularizing} & M$^*$ & dn & 0.2872 & 9.185 & 56.21 & 0.3333 & 10.098 & 43.72 & 0.2952 & 9.341 & 57.21 \\

			md4all-AD~\cite{gasperini2023robust} & Mv & dT\dag(nr) & 0.1523 & 6.853 & 83.11 & {0.2187} & {9.003} & {68.84} & 0.1601 & 7.832 & 78.97 \\

			md4all-DD~\cite{gasperini2023robust} & Mv & dT\dag(nr) & 0.1366 & {6.452} & 84.61 & {0.1921} & {8.507} & \underline{71.07} & {0.1414} & {7.228} & {80.98} \\

			DMMDE v1~\cite{tosi2025diffusion} & Ms & dT\ddag(nr) & 0.1370 & \underline{6.318} & 85.05 & \underline{0.1880} & \underline{8.432} & {69.94} & {0.1470} & {7.345} & {79.59} \\

			DMMDE v2~\cite{tosi2025diffusion} & Ms & dT\ddag(nr) & 0.1400 & {6.573} & 83.51 & {0.1970} & {8.826} & {69.65} & {0.1430} & {7.317} & {80.28} \\

			DMMDE v3~\cite{tosi2025diffusion} & Ms & dT\ddag(nr) & \textbf{0.1280} & {6.449} & 84.03 & {0.1910} & {8.433} & \textbf{71.14} & \underline{0.1390} & \underline{7.129} & \textbf{81.36} \\

			\textbf{ACDepth (Ours)} & Mv & dT(nr) & \underline{0.1340} & \textbf{6.284} & \underline{85.07} & \textbf{0.1873} & \textbf{8.125} & \textbf{71.14} & \textbf{0.1377} & \textbf{6.970} & \underline{81.32} \\

			\bottomrule
		\end{tabular}}
	\end{center}
	\caption{Evaluation of self-supervised methods on nuScenes~\protect\cite{caesar2020nuscenes} validation set. v1,v2,v3: depth maps from md4all-DD, DPT, Depth Anything in ~\protect\cite{tosi2025diffusion}. Supervisions (sup.): M: via monocular videos, *: test-time median-scaling via LiDAR, v: weak velocity, s: test-time scaling via LSE criterion~\protect\cite{ranftl2020towards}. Training data (tr.data): d: day-clear, n: night, r: rain, a: all. T\dag: Translated via GAN, T\ddag: Translated via Diffusion. Visual support: \textbf{1st} and \underline{2nd}.}
	\label{Table:Nuscenes}
	\vspace{-2mm}
\end{table*}
\subsection{Datasets}
In this study, the commonly used nuScenes~\cite{caesar2020nuscenes} and RobotCar~\cite{maddern20171} datasets are used for training and comparison. 
NuScenes is a comprehensive dataset featuring diverse outdoor scenes, specifically tailored for autonomous driving research, where the multi-frame sequence is supplemented with detailed radar annotations. Following~\cite{gasperini2023robust}, we adopt 15,129 generated samples (day-clear, day-rain, night) for training and 6,019 samples (including 4449 day-clear, 1088 rain, and 602 night) for testing. 
RobotCar is a large outdoor dataset collected in Oxford, UK. Following~\cite{gasperini2023robust}, we adopt 16,563 generated samples (day, night) for training and 1,411 samples (including 702 day 709 night) for testing. 
During the data preprocessing stage, images in the nuScenes dataset are resized to $320\times576$, while those in the RobotCar dataset are resized to $320\times544$.
We record metrics within a range of 0.1m to 50m for RobotCar and 0.1m to 80m for nuScenes. 
\subsection{Implementation Details} All experiments are conducted on the same ResNet18 architecture ~\cite{he2016deep}.
We train the student model and teacher model on a single NVIDIA 3090 GPU with a batch size of 16, using the Adam optimizer. We set initial learning rate to 5e-4, reducing it by a factor of 0.1 every 15 epoch. The student model are trained for 25 epochs.
Following the experimental protocol of ~\cite{gasperini2023robust}, we maintain identical hyperparameter settings for self-supervised learning.
Through experimental validation of different parameter combinations, the weights for the loss functions are set to $\lambda_1=0.01$, $\lambda_2=0.02$. 
Considering both the effectiveness of supervision signals and training efficiency, we select the small version of Depth Anything V2~\cite{yang2024depth} as our supervision prior  for ordinal guidance distillation.
Further details on the model training procedure and dataset translation process are provided in the appendix.
\subsection{Comparison with SoTAs}
In this section, we evaluate our model on the nuScenes and RobotCar datasets. Qualitative and quantitative comparisons are shown below.  

\paragraph{Comparison on the nuScenes.} In Table~\ref{Table:Nuscenes}, we compare our method with existing depth estimation models, including the typical MDE technologies (Monodepth2~\cite{godard2019digging}, PackNet-SfM~\cite{guizilini20203d}) and robust depth estimation approaches (RNW~\cite{wang2021regularizing}, md4all~\cite{gasperini2023robust} and DMMDE~\cite{tosi2025diffusion}). 
As shown in Table~\ref{Table:Nuscenes}, our proposed ACDepth shows significant competitiveness, particularly for the night condition, respectively reducing 2.50\% and 4.49\% in terms of absRel and RMSE metrics, when compared to robust depth estimation method md4all-DD. 
DMMDE~\cite{tosi2025diffusion} uses different depth models as prior information to guide image translation, and achieves improvements over md4all-DD. However, due to the scale-and-shift-invariant loss, the scaling relation of depth map is ignored and corrupted during alignment, degrading the final performance. 
For better convincing, Figure.~\ref{fig:4}(a) presents qualitative results. 
md4all-DD is particularly sensitive to water reflections in rainy scenes. For example, in the first and second rows of Figure.~\ref{fig:4}(a), the obvious errors of depth estimation are observed in the regions of water surface reflections. In addition for nighttime scenes, md4all-DD is hard to predict the real depth of regions hidden in the darkness. In contrast, our ACDepth method can produce reasonable and reliable predictions of depth information in night scenes, such as trees and cars under low-light conditions.

\paragraph{Comparison on the RobotCar.} 
To further verify the effectiveness, 
we compare our ACDepth with existing state-of-the-art approaches on the RobotCar benchmark. 
As expected in Table~\ref{robotcar}, 
our method achieves the best performance in almost all metrics.  Compared to md4all-DD, our approach reduces RMSE by 8.80\% on the standard benchmark and 11.99\% on the challenging benchmark, highlighting the superiority of our approach. To demonstrate the effectiveness of our method, Figure.~\ref{fig:4}(b) provides the visual comparisons, showing that our ACDepth can produce more precise depth maps over those of md4all-DD on both day and night scenes. 
We speculate that these considerable improvement of this study stems from the more reliable data generation and comprehensive prior constraints, which contribute more to the robust model training and optimization.  

\begin{table*}[!htbp]
	\begin{center}
    \scalebox{1.1}{
	\scriptsize
		\begin{tabular}{lll|cccc|cccc}
			\toprule
			&&&& \multicolumn{3}{c|}{\textit{day} -- RobotCar} & \multicolumn{4}{c}{\textit{night} -- RobotCar} \\
			Method & sup. & tr.data & absRel & sqRel & RMSE & $\delta_1$   & absRel & sqRel & RMSE & $\delta_1$   \\
			\midrule
			
			Monodepth2~\cite{godard2019digging}  & M$^*$ & d & {0.1196} & {0.670} &  \underline{3.164} & {86.38} & 0.3029 & 1.724 & {5.038} & 45.88 \\
			
			DeFeatNet~\cite{spencer2020defeat}   & M$^*$ & a: dn & 0.2470 & 2.980 & 7.884 & 65.00 & 0.3340 & 4.589 & 8.606 & 58.60 \\
			
			ADIDS~\cite{liu2021self} & M$^*$ & a: dn & 0.2390 & 2.089 & 6.743 & 61.40 & 0.2870 & 2.569 & 7.985 & 49.00 \\
			
			RNW~\cite{wang2021regularizing} & M$^*$ & a: dn & 0.2970 & 2.608 & 7.996 & 43.10 & {0.1850} & {1.710} & 6.549 & {73.30} \\
			
			WSGD~\cite{vankadari2023sun} & M$^*$ & a: dn & 0.1760 & 1.603 & 6.036 & 75.00 & {0.1740} & {1.637} & {6.302} & {75.40} \\

			 md4all-DD~\cite{gasperini2023robust} & Mv & dT\dag(n)&  \underline{0.1128} &  \underline{0.648} & {3.206} & {87.13} &  \underline{0.1219} & {0.784} &  \underline{3.604} & \textbf{84.86} \\
			 
			 DMMDE v1~\cite{tosi2025diffusion}    & Mv & dT\ddag(n) & {0.1190} & {0.676} & {3.239} & {87.20} & {0.1390} &  \underline{0.739} & {3.700} & {82.46} \\

			 DMMDE v2~\cite{tosi2025diffusion} & Mv & dT\ddag(n) & {0.1230} & {0.724} & {3.333} & {86.62} & {0.1330} & {0.824} & {3.712} & {83.95} \\

			DMMDE v3~\cite{tosi2025diffusion} & Mv & dT\ddag(n) & {0.1190} & {0.728} & {3.287} &  \underline{87.17} & {0.1290} & {0.751} & {3.661} & {83.68} \\

			 ACDepth (Ours) & Mv & dT(n) & \textbf{0.1107} & \textbf{0.591} & \textbf{3.084} & \textbf{88.03} & \textbf{0.1206} & \textbf{0.690} & \textbf{3.432} &  \underline{84.47} \\

			\bottomrule
		\end{tabular}}
	\end{center}
	\vspace{-2mm}
	\caption{Evaluation of self-supervised works on the RobotCar~\protect\cite{maddern20171} test set. Trailing 0 added to the values from~\protect\cite{vankadari2023sun} and ~\protect\cite{tosi2025diffusion}. Notation from Table~\ref{Table:Nuscenes}.} 
	\label{robotcar}
	\vspace{-2mm}
\end{table*}

\subsection{Ablation Study}
In this section, we conduct detailed ablation experiments on nuScenes and RobotCar datasets to demonstrate the individual effectiveness of the proposed components.
\begin{figure}[htbp]
	\centering
	\includegraphics[width=0.98\columnwidth]{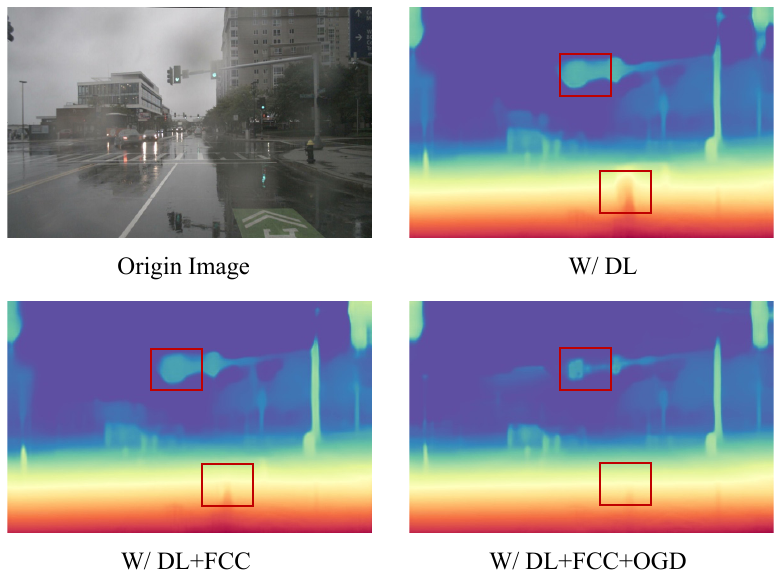}
	\caption{Visualization of ablation study on the distillation learning, feature consistency constraint and ordinal guidance distillation.}
	\label{Figure:4}
\end{figure}

\paragraph{Evaluation on Major Design Components.} 
\begin{table}[!htb]
\centering
\resizebox{0.49 \textwidth}{!}{
\begin{tabular}{cccc|cccccc}
\hline
\multirow{2}{*}{}         & \multirow{2}{*}{DL} & \multirow{2}{*}{OGD} & \multirow{2}{*}{FCC} & \multicolumn{2}{c|}{\textit{day-clear}}      & \multicolumn{2}{c|}{\textit{night}}                                & \multicolumn{2}{c}{\textit{day-rain}} \\ \cline{5-10}
                          &                     &                     &                     & absRel & \multicolumn{1}{c|}{RMSE}  & absRel                      & \multicolumn{1}{c|}{RMSE} & absRel          & RMSE       \\ \hline
\multirow{4}{*}{N} &                     &                     &                     & 0.1333   & \multicolumn{1}{c|}{6.459} & 0.2419                      & \multicolumn{1}{c|}{10.922}  & 0.1572        & 7.453        \\
                          & \checkmark                   &                     &                     & 0.1335 & \multicolumn{1}{c|}{6.408} & 0.1883                      & \multicolumn{1}{c|}{8.398}  & 0.1404        & 7.092        \\
                          & \checkmark                   & \checkmark                   &                     & \textbf{0.1325} & \multicolumn{1}{c|}{\textbf{6.328}} & 0.1879                      & \multicolumn{1}{c|}{8.353}  & 0.1414        & 7.084        \\
                          & \checkmark                   & \checkmark                   & \checkmark                   & 0.1355 & \multicolumn{1}{c|}{6.340} & \textbf{0.1872}                      & \multicolumn{1}{c|}{\textbf{8.125}}  & \textbf{0.1377}        & \textbf{6.999}        \\ \hline
\multirow{2}{*}{}         & \multirow{2}{*}{DL} & \multirow{2}{*}{OGD} & \multirow{2}{*}{FCC} & \multicolumn{3}{c|}{\textit{day}}                                          & \multicolumn{3}{c}{\textit{night}}                                  \\ \cline{5-10} 
                          &                     &                     &                     & absRel & RMSE                       & \multicolumn{1}{c|}{sqRel}  & absRel                      & RMSE          & sqRel        \\ \hline
\multirow{4}{*}{R} &                     &                     &                     & 0.1209 & 3.335                      & \multicolumn{1}{c|}{0.723} & 0.3909                       & 8.227        & 3.547        \\
                          & \checkmark                   &                     &                     & 0.1123 & 3.135                      & \multicolumn{1}{c|}{0.631} &0.1233                        & 3.476        & 0.720        \\
                          & \checkmark                   & \checkmark                   &                     & 0.1117 & 3.115                      & \multicolumn{1}{c|}{0.615} & 0.1224                       & 3.458        & 0.704        \\
                          & \checkmark                   & \checkmark                   & \checkmark                   & \textbf{0.1107} & \textbf{3.084}                      & \multicolumn{1}{c|}{\textbf{0.591}} & \textbf{0.1206}                       & \textbf{3.432}        & \textbf{0.690}        \\ \hline
\end{tabular}}
	\caption{Ablation study of Design Components. N: nuScenes, R: RobotCar, DL: distillation learning, OGD: ordinal guidance distillation, FCC: feature consistency constraint.} \vspace{-2mm}
	\label{Table:3}
\end{table}

Our baseline is trained on daytime scenes~\cite{gasperini2023robust}. 
Based on this, we introduce the distillation learning (DL) to promote training with more robust feature representation, reducing the absRel metric by 22.16\% for night scene and 10.69\% for rainy scene on the nuScenes.
Then we construct another model with additional ordinal guidance distillation (OGD), which provides more precise supervision signals for degraded conditions. By contrast, it achieves the comprehensive improvements in model performance, reducing the absRel metric by 0.54\% for day scene and 0.73\% for night scene on the RobotCar.   
In addition, we introduce the feature consistency constraint (FCC) to evaluate its effect for robust feature representation. As expected, the model with FCC significantly enhances the accuracy of depth estimation in challenging conditions (such as nighttime and rainy scenes), reducing the absRel metric by 0.37\% for night scene and 2.62\% for rain scene on the nuScenes.
We also provide the visual comparisons between the improved versions. As shown in Figure.~\ref{Figure:4}, the complete model equipped with DL, OGD and FCC strategies shows significant superiority over its imperfect versions. These optimization strategies not only provide reliable supervised signals but also guide the network to focus on uncertain regions, resulting in more robust depth estimation.

\paragraph{Evaluation on Ordinal Guidance Distillation.} In this section, we investigate the impact of different sampling methods on ordinal guidance distillation, with the experimental results presented in Table~\ref{Table:4}. 
Our OGD strategy employs two sampling methods: global and local sampling. We observed that model performance became more reliable as both sampling methods were progressively incorporated. 
Additionally, our experiments demonstrate that using windowed sampling regions enables the model to learn more robust feature representations, significantly reducing the risk of overfitting.

\begin{table}[!htb]
\centering
\resizebox{0.49 \textwidth}{!}{
\begin{tabular}{cccc|cccccc}
\hline
\multirow{2}{*}{}         & \multirow{2}{*}{G} & \multirow{2}{*}{L} & \multirow{2}{*}{W} & \multicolumn{2}{c|}{\textit{day-clear}}      & \multicolumn{2}{c|}{\textit{night}}                                & \multicolumn{2}{c}{\textit{day-rain}} \\ \cline{5-10}
                          &                     &                     &                     & absRel & \multicolumn{1}{c|}{RMSE}  & absRel                      & \multicolumn{1}{c|}{RMSE} & absRel          & RMSE       \\ \hline
\multirow{4}{*}{N} &                     &                     &                     & 0.1371 & \multicolumn{1}{c|}{6.560} & 0.1884                      & \multicolumn{1}{c|}{8.209}  & 0.1412        & 7.178        \\
                          & \checkmark                   &                     &                     & 0.1343 & \multicolumn{1}{c|}{6.384} & 0.1876                      & \multicolumn{1}{c|}{8.161}  & 0.1401        & 7.004        \\
                          & \checkmark                   & \checkmark                   &                     & \textbf{0.1330} & \multicolumn{1}{c|}{\textbf{6.318}} & 0.1875                      & \multicolumn{1}{c|}{8.141}  & \textbf{0.1377}        & \textbf{6.982}        \\
                          & \checkmark                   & \checkmark                   & \checkmark                   & 0.1355 & \multicolumn{1}{c|}{6.340} & \textbf{0.1872}                      & \multicolumn{1}{c|}{\textbf{8.125}}  & \textbf{0.1377}        & 6.999        \\ \hline
\multirow{2}{*}{}         & \multirow{2}{*}{G} & \multirow{2}{*}{L} & \multirow{2}{*}{W} & \multicolumn{3}{c|}{\textit{day}}                                          & \multicolumn{3}{c}{\textit{night}}                                  \\ \cline{5-10} 
                          &                     &                     &                     & absRel & RMSE                       & \multicolumn{1}{c|}{sqRel}  & absRel                      & RMSE          & sqRel        \\ \hline
\multirow{4}{*}{R} &                     &                     &                     & 0.1144 & 3.184                      & \multicolumn{1}{c|}{0.642} & 0.1230                       & 3.448        & 0.721        \\
                          & \checkmark                   &                     &                     & 0.1128 & 3.157                      & \multicolumn{1}{c|}{0.623} & 0.1228                       & 3.443        & 0.691        \\
                          & \checkmark                    & \checkmark                    &                     & \textbf{0.1107} & 3.095                      & \multicolumn{1}{c|}{0.600} & 0.1223                       & 3.453        & \textbf{0.685}        \\
                          &\checkmark                   & \checkmark                   & \checkmark                    & \textbf{0.1107} & \textbf{3.084}                      & \multicolumn{1}{c|}{\textbf{0.591}} & \textbf{0.1206}                       & \textbf{3.432}        & 0.690       \\ \hline
\end{tabular}}
	\caption{Ablation study of sampling method for ranking loss. G: global sampling, L: local sampling, W: window sampling.}
	\label{Table:4}
\end{table}


\section{Conclusion}
In this paper, we propose a novel approach named ACDepth for robust monocular depth estimation under adverse weather conditions. In addition, we introduce an elaborate data generation scheme to produce the multi-tuple depth dataset with diverse degradations, which significantly mitigates the problem of data scarcity. Meanwhile, we construct an effective multi-granularity knowledge distillation (MKD) strategy to achieve the robust model training, which facilitates the complete transfer and alignment of capabilities between the teacher model and student model. Extensive experimental results and comprehensive ablation demonstrate the effectiveness of our ACDepth, demonstrating its superior performance compared to SoTA solutions.

\section{Acknowledgments}
This research was financially supported by the Natural Science Foundation of Heilongjiang Province of China for Excellent Youth Project (YQ2024F006), the National Natural Science Foundation of China (U23B2009) and Open Research Fund from Guangdong Laboratory of Artificial Intelligence and Digital Economy (SZ) (GML-KF-24-09).

\bibliographystyle{named}
\bibliography{main}

\clearpage
\appendix
\section{Details of data Translation}

In this section, we provide a detailed discussion of dataset generation, focusing on the transitions from day to night and day to rainy conditions in the nuScenes dataset, as well as from day to night in the RobotCar dataset.

For our image generation process, we build upon the original CycleGAN-Turbo~\cite{parmar2024one} framework to generate challenging scene images from clear input images. CycleGAN-Turbo not only supports traditional RGB three-channel inputs but also incorporates a text prompt adjustment mechanism to enhance the scene adaptability of the generated images. Specifically, we set distinct text prompts for different scenes: ``outdoor picture of clear day'' for clear samples, ``outdoor picture of night'' for nighttime scenes, and ``outdoor picture of rainy day'' for rainy scenes.

In the study~\cite{gasperini2023robust}, the authors used ForkGAN~\cite{zheng2020forkgan} to perform cross-domain translation tasks for datasets. This GAN-based translation model can generate images that closely resemble real-world scene distributions, but its training requires a large number of real-world scene images. For example, in the translation task on the RobotCar~\cite{maddern20171} dataset, the authors used 34,128 daytime images and 32,585 nighttime images for training. Similarly, for the nuScenes~\cite{caesar2020nuscenes} dataset, additional datasets were needed. Specifically, for the day-to-rainy-day translation task, the authors used the nuImages dataset~\cite{caesar2020nuscenes}, which contains 19,857 rainy-day images and 19,685 daytime images; for the day-to-night translation task, they first trained a day-to-night conversion model on the BDD100K~\cite{yu2020bdd100k} dataset and then fine-tuned it on the nuScenes dataset. The conditions required for training the translator are summarized in Table~\ref{Table:5}. This process highlights that GAN-based translation models require large-scale data to achieve effective cross-domain translation tasks. However, in real-world scenarios, acquiring such large-scale multi-scene image data is often impractical, limiting the applicability of GAN-based translation models. Recently, diffusion model-based image translation methods have gained attention. These methods generate images through a step-by-step denoising process, but the generation process of diffusion models lacks precise control over the generated content, limiting their use in image translation tasks. To address these challenges, the core objective of this paper is to explore an image translation method that reduces data dependency while ensuring high consistency in both content and style in the translated images.

\begin{table}[htbp]
\centering
\resizebox{0.48 \textwidth}{!}{
\begin{tabular}{c|cccc|cc}
\hline
 & \multicolumn{4}{c|}{nuScense} & \multicolumn{2}{c}{RobotCar}\\ \cline{2-7} 
       & \textit{day-clear}   & \multicolumn{1}{c|}{\textit{night}} & \textit{day-clear}           & \textit{day-rain}          & \textit{day}            & \textit{night}        \\ \hline
md4all & 36728 & \multicolumn{1}{c|}{27971} & 19685         & 19857         & 34128          & 32585        \\
our    &  500  & \multicolumn{1}{c|}{500}   &     500       &   500         &    500         &   4000       \\ \hline
\end{tabular}}
\caption{Data details for training translators}
\label{Table:5}
\end{table}

In this study, we randomly select 500 images from the clear-day images and night images of nuScenes to train the day-to-night scene translation model. Similarly, we randomly select 500 clear-day images and 500 rain images to train the day-to-rain scene translation model. For the RobotCar dataset, we initially used the same data scale as in the nuScenes. However, the experimental results were suboptimal. We hypothesize that this result is due to the unique lighting distribution characteristics of nighttime samples in the RobotCar. Specifically, the highly uneven light distribution and significant variations in light intensity in night samples make it challenging to train a high-quality translation model with a small number of samples.

To validate our hypothesis, we conducted three experiments: (1) randomly selecting 500 night images as the training set, (2) carefully selecting 500 night images with high-quality lighting conditions, (3) randomly selecting 4,000 night images. The number of the day sample is fixed at 500 across all experiments. As shown in Table~\ref{Table:6}, increasing the dataset scale improved the quality of data generation. Notably, selecting a subset of high-quality nighttime images achieves high-quality translation results with a smaller dataset.

\begin{table}[htbp]
\centering
\resizebox{0.48 \textwidth}{!}{
\begin{tabular}{c|cccc|cccc}
\hline
\multirow{2}{*}{} & \multicolumn{4}{c|}{\textit{day}}  & \multicolumn{4}{c}{\textit{night}} \\
                  & absRel & sqRel & RMSE & $\delta_1$ & absRel & sqRel & RMSE & $\delta_1$ \\ \hline
Random 500        &    0.1157    &   0.650    &   3.180   &  87.40 &    0.1285    &  0.785     &   3.600   &  83.40 \\
Selected 500      &   0.1129     &    0.633   &   3.160   &  87.92 &   0.1239     &   0.701    &   3.436   &  84.39 \\
Random 4000       &     0.1107    &   0.591    &   3.084   & 88.03  &  0.1206      &   0.690    &    3.432  &  84.47 \\ \hline
\end{tabular}}
\caption{Detailed setup of translation for RobotCar dataset}
\label{Table:6} 
\end{table}

To ensure the capability of the translation model, we retain the core components from CycleGAN-Turbo, including skip connections and retraining the first layer of the U-Net. 
These components significantly enhance the preservation of intricate details during cross-domain adaptation. For each pair of scene translation, we independently train a dedicated set of LoRA parameters to adaptively adjust the model’s behavior for domain-specific requirements. 
We take the day-to-rain scene translation on the nuScenes dataset as a example to describe the translation training.
Following CycleGAN-Turbo, we jointly train two bidirectional translation tasks: clear day to rainy and rainy to clear day. In this formulation, $x$ represents image from clear scene and $y$ represents image from rainy scene. 
Text embedding $c_X$ and $c_Y$ as the condition inputs corresponding to the two tasks, where they are set as ``outdoor picture on clear day'' and ``outdoor picture on rainy day''.
To enable the single-step diffusion model to generate realistic and domain-consistent samples, we adopt the same training objective as in~\cite{parmar2024one}, which consists of adversarial loss, cycle consistency loss and identity regularization loss. The cycle-consistency loss ensures content preservation between translated and original images, mathematically formulated as:
\begin{equation}
\begin{split}
L_{\text{cycle}} &= \mathbb{E}_{x \sim p_{\text{data}}(x)} \left[ L_{rec}(F_c(F_c(x, c_Y), c_X), x)  \right] \\
&\quad + \mathbb{E}_{y \sim p_{\text{data}}(y)} \left[ L_{rec}(F_c\left( F_c(y, c_X), c_Y \right), y ) \right],
\end{split}
\end{equation}
where $L_{rec}$ represents the combination of L1 and LPIPS, and $F_c$ is the translator model introduced in Section 3.3. The adversarial loss drives realistic image generation in the target domain through adversarial training, depicted as:
\begin{equation}
\begin{split}
L_{\text{GAN}} &= 
\mathbb{E}_{y \sim p_{\text{data}}(y)} \left[ \log D_Y(y) \right] \\
&\quad+\mathbb{E}_{x \sim p_{\text{data}}(x)} \left[ \log \left(1 - D_Y(F_c(x, c_Y)) \right) \right] \\
&\quad+\mathbb{E}_{x \sim p_{\text{data}}(x)} \left[ \log D_X(x) \right] \\
&\quad+\mathbb{E}_{y \sim p_{\text{data}}(y)} \left[ \log \left(1 - D_X(F_c(y, c_X)) \right) \right].
\end{split}
\end{equation}
The identity regularization loss serves as a critical component for preserving intra-domain characteristics, defined as:
\begin{equation}
\begin{split}
L_{\text{idt}} &= 
\mathbb{E}_{y \sim p_{\text{data}}(y)} \left[ L_{\text{rec}}\left(F_c(y, c_Y), y\right) \right] \\
&\quad + \mathbb{E}_{x \sim p_{\text{data}}(x)} \left[ L_{\text{rec}}\left(F_c(x, c_X), x\right) \right].
\end{split}
\end{equation}
The total loss is formulated as the summation of three components. After training, only the parameters corresponding to the day to rainy translation are retained to constitute our scene transformation.

\section{Other Implement Details}
\subsection{More Training Detail}
For self-supervised learning, following~\cite{gasperini2023robust}, we use Adam as the optimizer, and the batch size is set to 16. We set the learning rate for both the depth and pose networks of the teacher to 2e-4. 
For robust model training, since the network has not learned any useful information in the early stages of training, premature implementation of feature alignment could hinder successful training. Therefore, we initiated feature granularity learning in the 15th epoch on the nuScenes dataset and in the 5th epoch on the RobotCar dataset.
During the data translation and training stage, images in the nuScenes dataset are resized to 320×576, while those in the RobotCar dataset are resized to 320×544.

\subsection{Augmentation on training set}
For nuScenes, we applied color perturbations, added Gaussian noise, and random horizontal flips to easy samples. For complex samples, we only applied random horizontal flips. For RobotCar, we applied color perturbations and random horizontal flips to easy samples. For complex samples, we only applied random horizontal flips.
\subsection{Evaluation Metrics}
We used four evaluation metrics, as described in ~\cite{gasperini2023robust}, including absRel, sqRel, RMSE, and $\delta_1$:
\begin{equation}
    \begin{split}
        & \text{Abs rel} = \frac{1}{|N_d|} \sum_{i \in N_d} 
        \frac{\rvert d_i-d_i^{*} \rvert}{d_i^{*}} , \\
        & \text{Sq rel} =  \frac{1}{|N_d|} \sum_{i \in N_d}
        \frac{\rVert d_i-d_i^{*}\rVert^2}{d_i^{*}}  , \\
        & \text{RMSE} = \sqrt{ \frac{1}{|N_d|} \sum_{i \in N_d}\rVert d_i-d_i^{*} \rVert^2 }, \\
        & \delta_1: \, \% \; \text{of} \; d_i \; s.t. \ \max\left(\frac{d_i}{d_i^{*}}, \frac{d_i^{*}}{d_i}\right) < 1.25, \\  
    \end{split}
    \label{equ:metrics}
\end{equation}

where $d_i$ represents the predicted depth value of pixel $i$, $d_i^{*}$ represents the ground truth depth value of pixel $i$, and $N_d$ is the total number of pixels.

\section{Other Experiments Result}

\subsection{Zero-shot Experiments Result}
To validate the zero-shot generalization capability of our proposed ACDepth, we conducted evaluations on the FogCityScape~\cite{sakaridis2018semantic} and DrivingStereo~\cite{yang2019drivingstereo} datasets. DrivingStereo is a dataset comprising 500 real-world fog and rain scenes, utilized for zero-shot testing under the protocol described in ~\cite{wang2024digging}. FogCityScape is a synthetic dataset based on Cityscapes, containing 1,525 test images following the evaluation setting in ~\cite{saunders2023self}. For all models, we used models trained on the nuScenes dataset. The experimental results in Table~\ref{Table:ZS} validate that our model achieves dominant performance under zero-shot settings.
\begin{table}[htbp]
    \centering
        \resizebox{0.48\textwidth}{!}{
        \begin{tabular}{c | c c c c | ccc}
            \hline
            Method &  absRel $\downarrow$ & sqRel $\downarrow$  &  RMSE $\downarrow$ & RMSE log $\downarrow$  & $\delta_{1} \uparrow$ &  $\delta_{2} \uparrow$ &  $\delta_{3} \uparrow$\\
            \hline
            \multicolumn{8}{c}{DrivingStereo Foggy} \\
            \hline
            Monodepth  & 0.150     	& 1.843     	& 8.727    	& 0.200     	& 0.813     	& 0.954     	& 0.986 \\
            Md4all-DD & {0.135} & {1.357}     	& {7.692}     	& {0.181}     	& 0.839     	& {0.965}     	& {0.991} \\
            ACDepth   & \textbf{0.132}     	& \textbf{1.294}      	& \textbf{7.408}     	& \textbf{0.176}     	& \textbf{0.841}     	& \textbf{0.970}     	& \textbf{0.992} \\
            \hline
            \multicolumn{8}{c}{DrivingStereo Rainy} \\
            \hline
            Monodepth  & 0.198     	& 2.489     	& 10.053    	& 0.243     	& 0.687     	& 0.922     	& 0.981 \\
            Md4all-DD & {0.171} & {1.909}     	& {8.958}     	& {0.227}     	& 0.719     	& {0.938}     	& {0.984} \\
            ACDepth   & \textbf{0.170}     	& \textbf{1.904}     	& \textbf{8.795}     	& \textbf{0.224}     	& \textbf{0.713}     	& \textbf{0.943}     	& \textbf{0.987} \\
            \hline
           \multicolumn{8}{c}{Fogcityscape} \\
           \hline
            Monodepth  & 0.192     	& 3.463     	& 10.210    	& 0.249     	& 0.770     	& 0.915     	& 0.968 \\
            Md4all-DD & {0.171} & {2.497}     	& {8.863}     	& {0.228}     	& 0.788     	& {0.929}     	& {0.976} \\
            ACDepth   & \textbf{0.163}     	& \textbf{2.412}     	& \textbf{8.759}     	& \textbf{0.219}     	& \textbf{0.803}     	& \textbf{0.934}     	& \textbf{0.978}
            \\
            \hline
        \end{tabular}
        }
        \caption{Quantitative results of zero-shot evaluation on FogCityScape and DrivingStereo dataset.
        \vspace{-1mm}
        }
    \label{Table:ZS}
    \vspace{-4mm}
\end{table}

\subsection{Experiments Analysis on $\lambda_1$ and $\lambda_2$}
\begin{figure}[t]
    \centering
    \subfigure[ABS rel]{
    \label{Fig:fdat_a}
    \begin{minipage}[c]{0.32\linewidth}
    \includegraphics[width=1in]{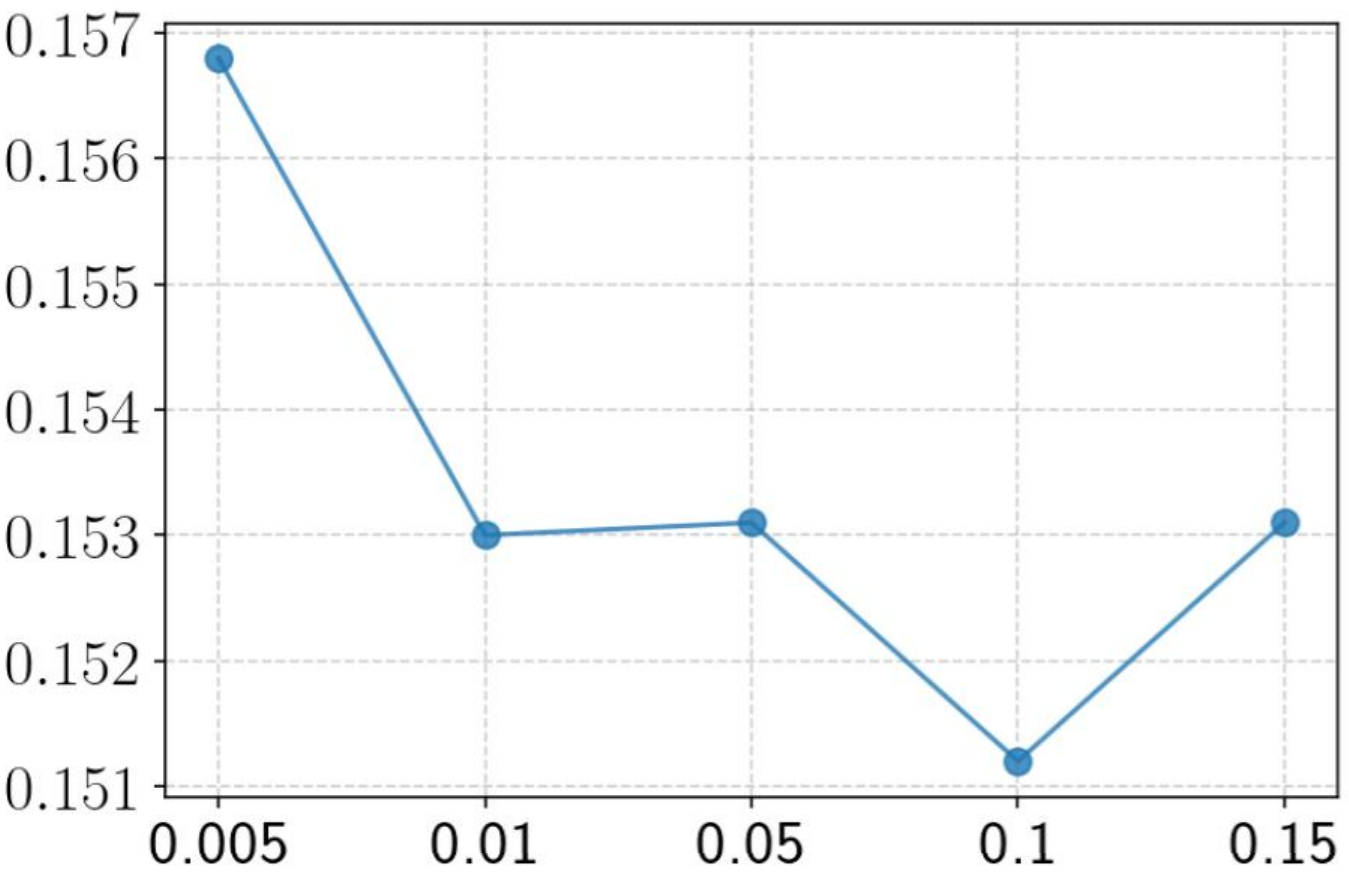}
    
    \end{minipage}%
    }%
    \subfigure[RMSE]{
    \label{Fig:fdat_b}
    \begin{minipage}[c]{0.32\linewidth}
    \includegraphics[width=1in]{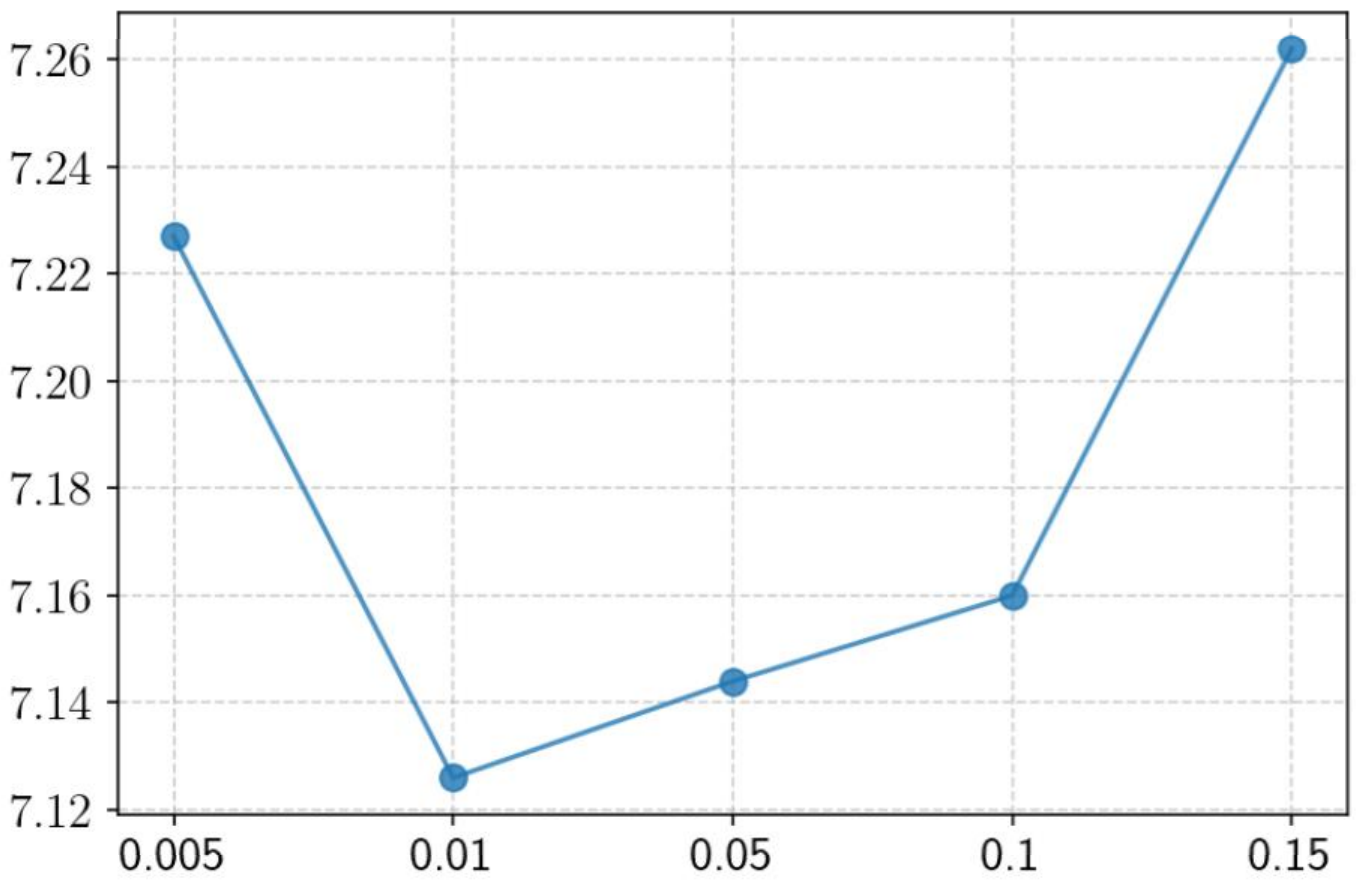}
    \end{minipage}%
    }%
    \subfigure[$ \delta_1 $]{
    \label{Fig:fdat_c}
    \begin{minipage}[c]{0.32\linewidth}
    \includegraphics[width=1in]{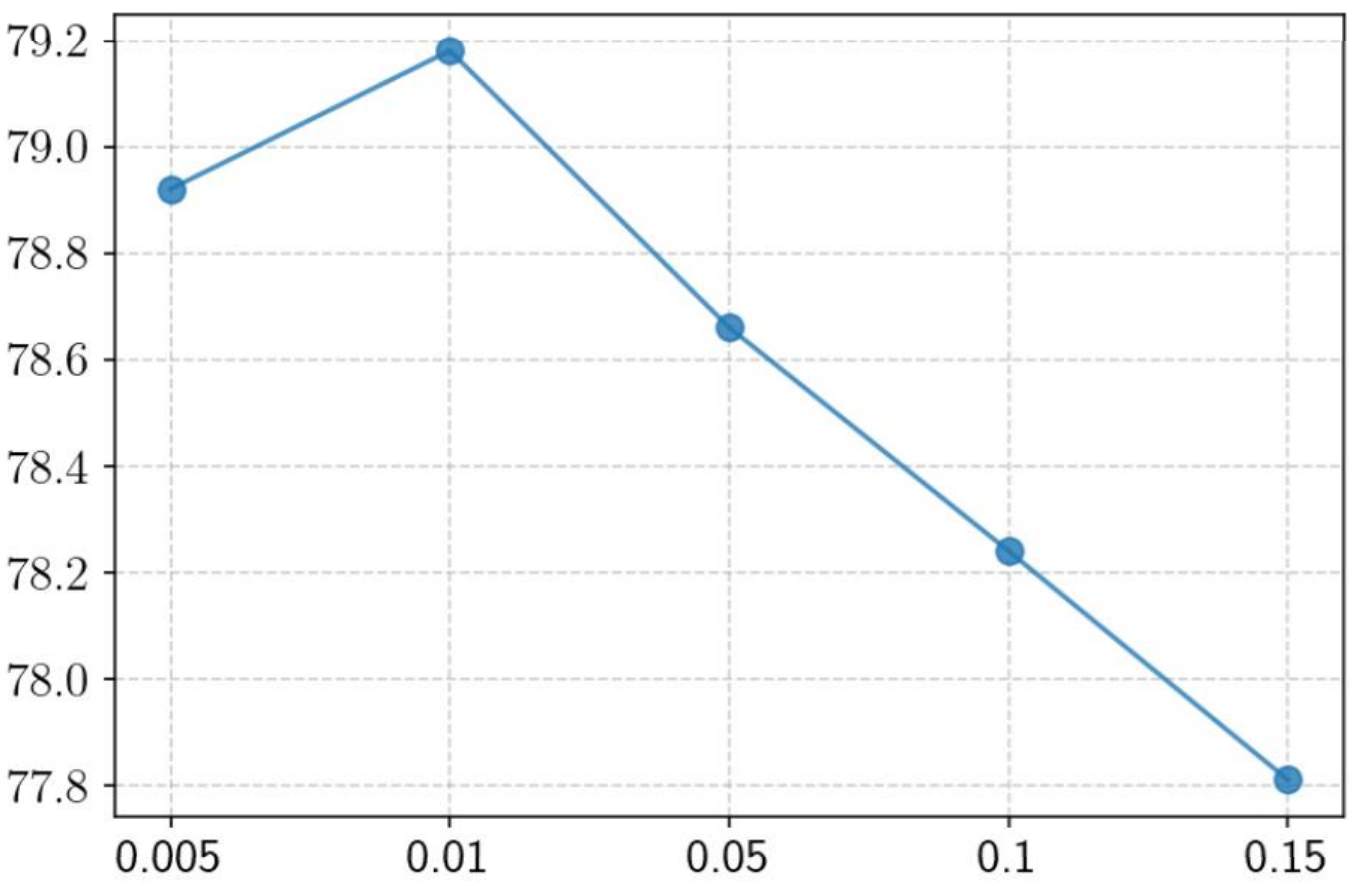}
    \end{minipage}
    }%
    \centering
    \vspace{-3mm}
    \caption{Average results for different $\lambda_1$ on the nuScenes dataset.}
    \label{Fig:loss}
\end{figure}

To determine the optimal weights for the loss components $L_r$ (controlled by $\lambda_1$) and $L_c$ (controlled by $\lambda_2$), we conducted systematic parameter ablation experiments. First, with $\lambda_1$ fixed at 0.01, we evaluated $\lambda_2 \in \{0.01, 0.02, 0.05\}$ on the nuScenes benchmark. As demonstrated in Table~\ref{Table:WL}, $\lambda_2 = 0.02$ achieved robust performance. Subsequently, maintaining $\lambda_2 = 0.02$, we analyzed $\lambda_1 \in \{0.005, 0.01, 0.05, 0.1, 0.15\}$. Fig.~\ref{Fig:loss} reveals that $\lambda_1 = 0.01$ optimally balances all evaluation metrics. The final configuration establishes $\lambda_1 = 0.01$ and $\lambda_2 = 0.02$.

\begin{table}[htbp]
    \centering
        \resizebox{0.30\textwidth}{!}{
        \begin{tabular}{cccc}
            \hline
             $\lambda_2$ &  absRel $\downarrow$  &  RMSE $\downarrow$ & $\delta_{1} \uparrow$ \\
            \hline
            $\lambda_2=0.01$   & 0.1537      	& 7.135     	& \textbf{79.26} \\
            $\lambda_2=0.02$  & \textbf{0.1530} & \textbf{7.126}      	& 79.18\\
            $\lambda_2=0.05$    & 0.1556       	& {7.187}     	& 78.96\\
            \hline
        \end{tabular}
        }
        \caption{Average results for different $\lambda_2$ on the nuScenes dataset.}
    \label{Table:WL}
    \vspace{-4mm}
\end{table}

\section{Additional Translation Qualitative Results}
\begin{figure*}[htbp]
	\centering
	\newcommand{\turnheightnew}{0.35\columnwidth}
	\begin{tabular}{@{\hskip 1mm}c@{\hskip 1mm}c@{\hskip 1mm}c@{\hskip 1mm}c@{\hskip 1mm}c@{\hskip 1mm}c@{}}
		\vspace{-0.5mm}
		{\rotatebox{90}{\hspace{2mm}}} &
		\includegraphics[width=\turnheightnew,keepaspectratio]{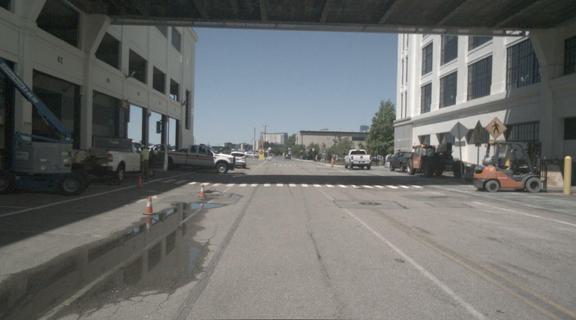} &
		\includegraphics[width=\turnheightnew,keepaspectratio]{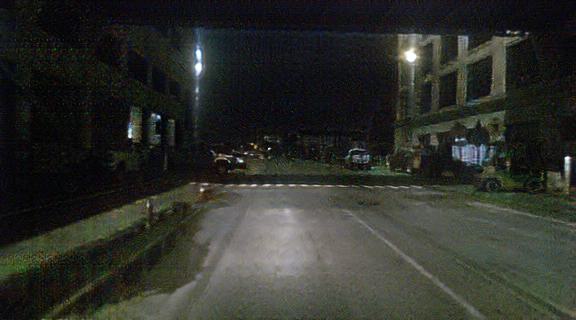} &
		\includegraphics[width=\turnheightnew,keepaspectratio]{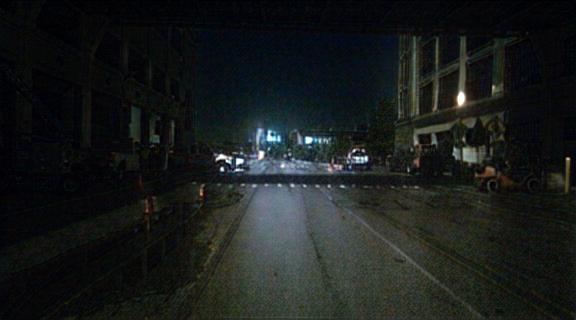} &
		\includegraphics[width=\turnheightnew,keepaspectratio]{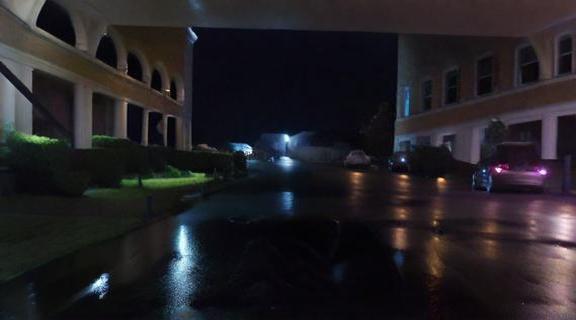} &
		\includegraphics[width=\turnheightnew,keepaspectratio]{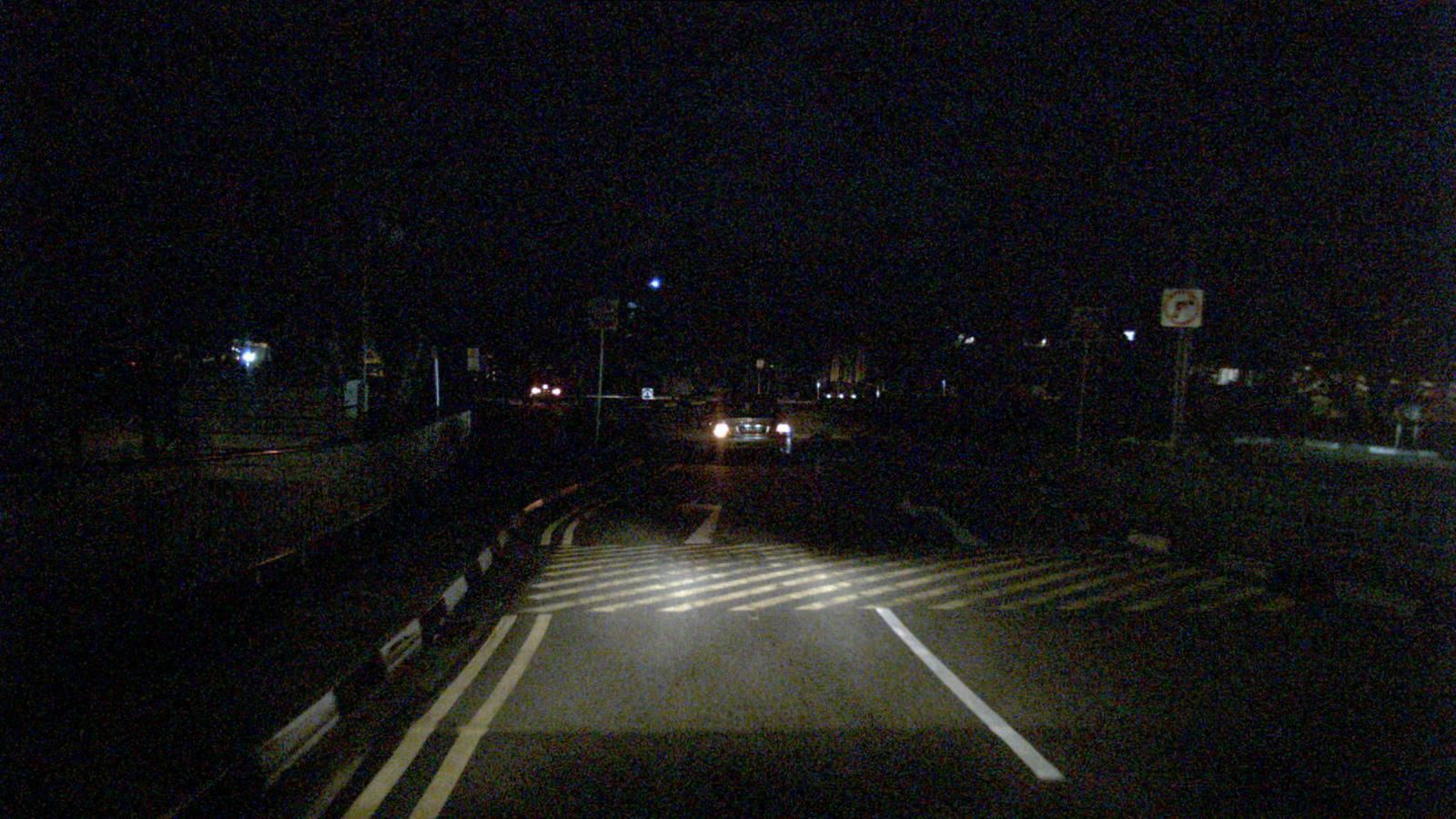} \\
		
		\vspace{-0.5mm}
		\rotatebox{90}{\hspace{0mm}}&
		\includegraphics[width=\turnheightnew]{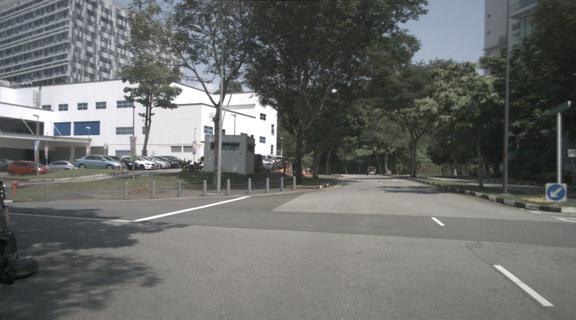} &
		\includegraphics[width=\turnheightnew]{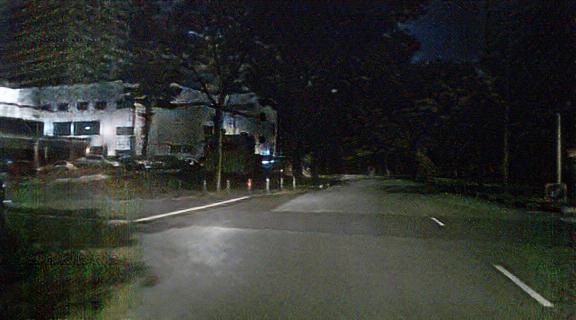} &
		\includegraphics[width=\turnheightnew]{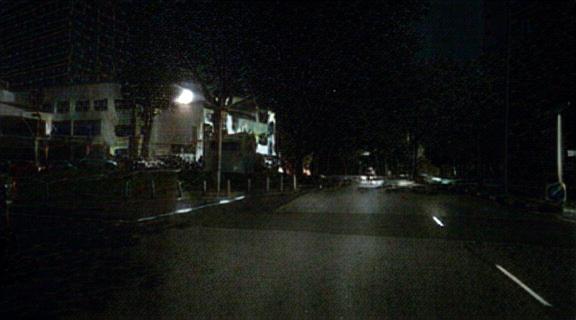} &
		\includegraphics[width=\turnheightnew]{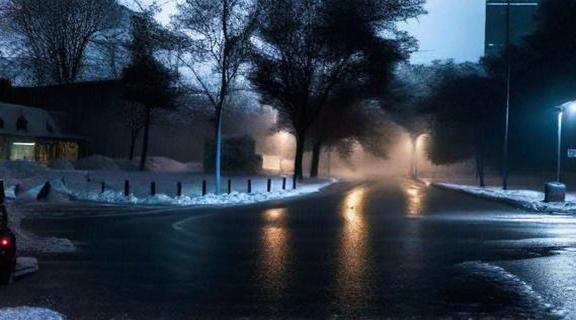} &
		\includegraphics[width=\turnheightnew]{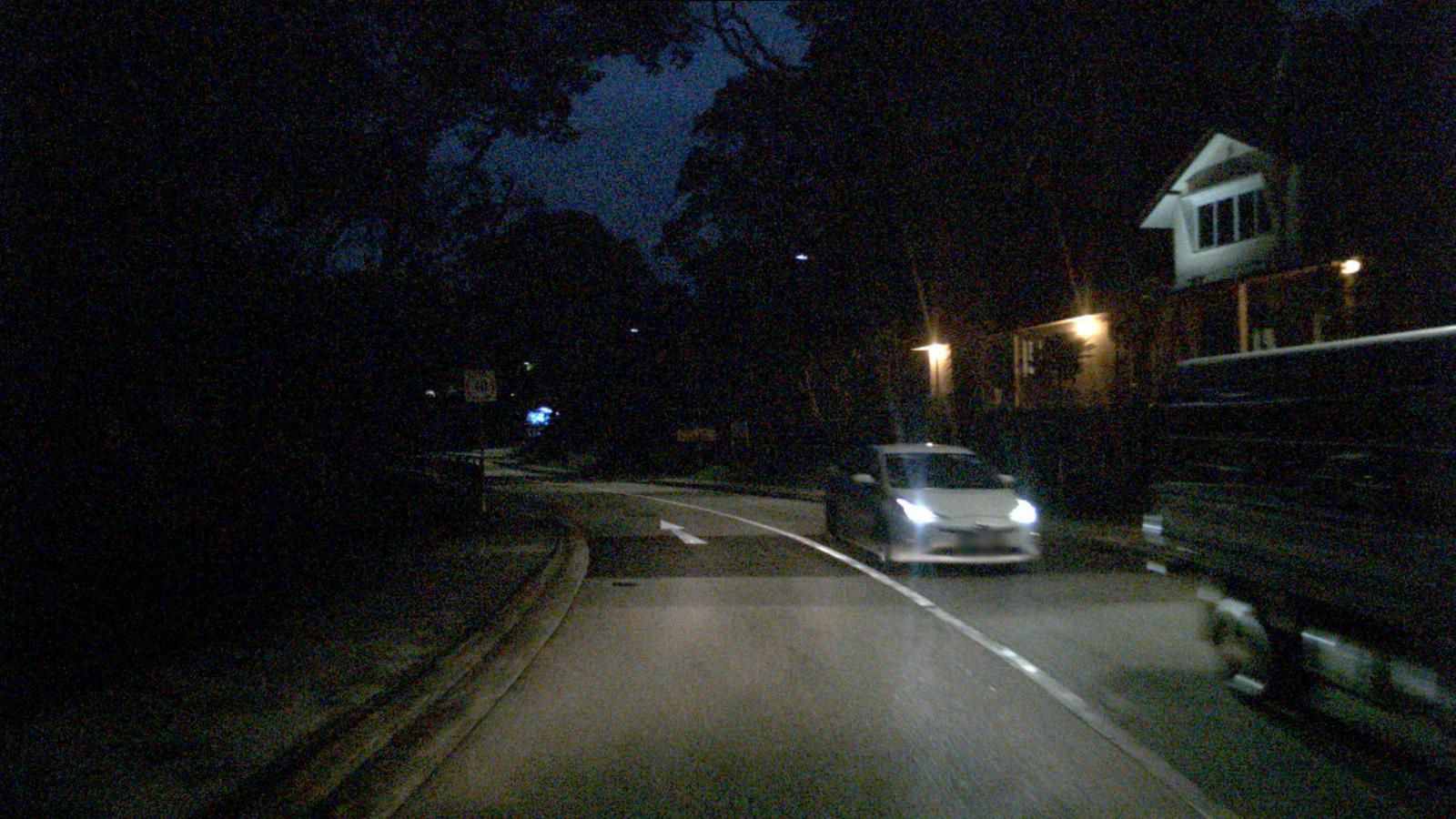} \\
		
		\vspace{-0.5mm}
		\rotatebox{90}{\hspace{0mm}}&
		\includegraphics[width=\turnheightnew]{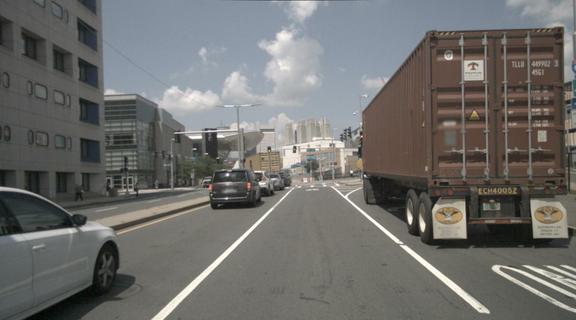} &
		\includegraphics[width=\turnheightnew]{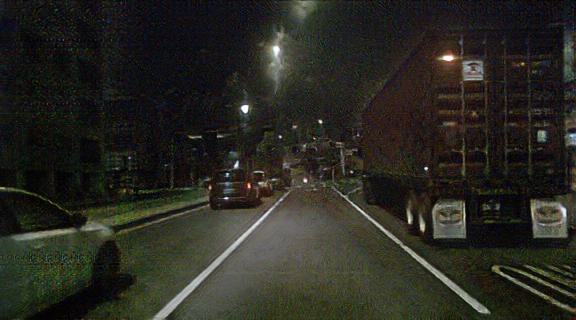} &
		\includegraphics[width=\turnheightnew]{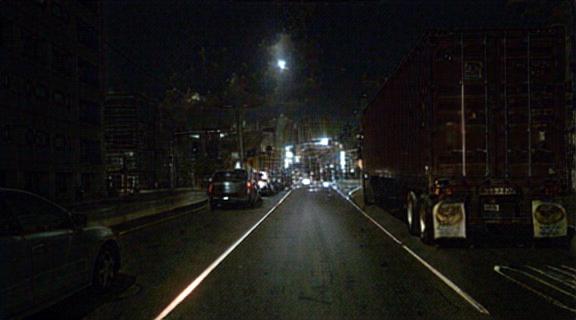} &
		\includegraphics[width=\turnheightnew]{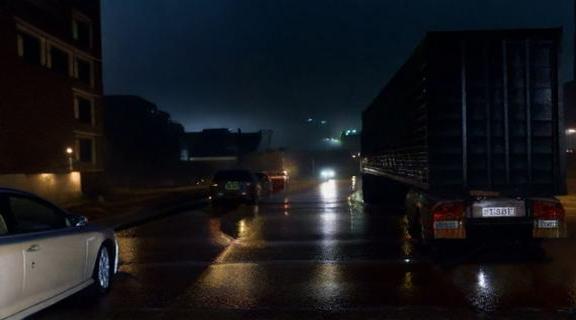} &
		\includegraphics[width=\turnheightnew]{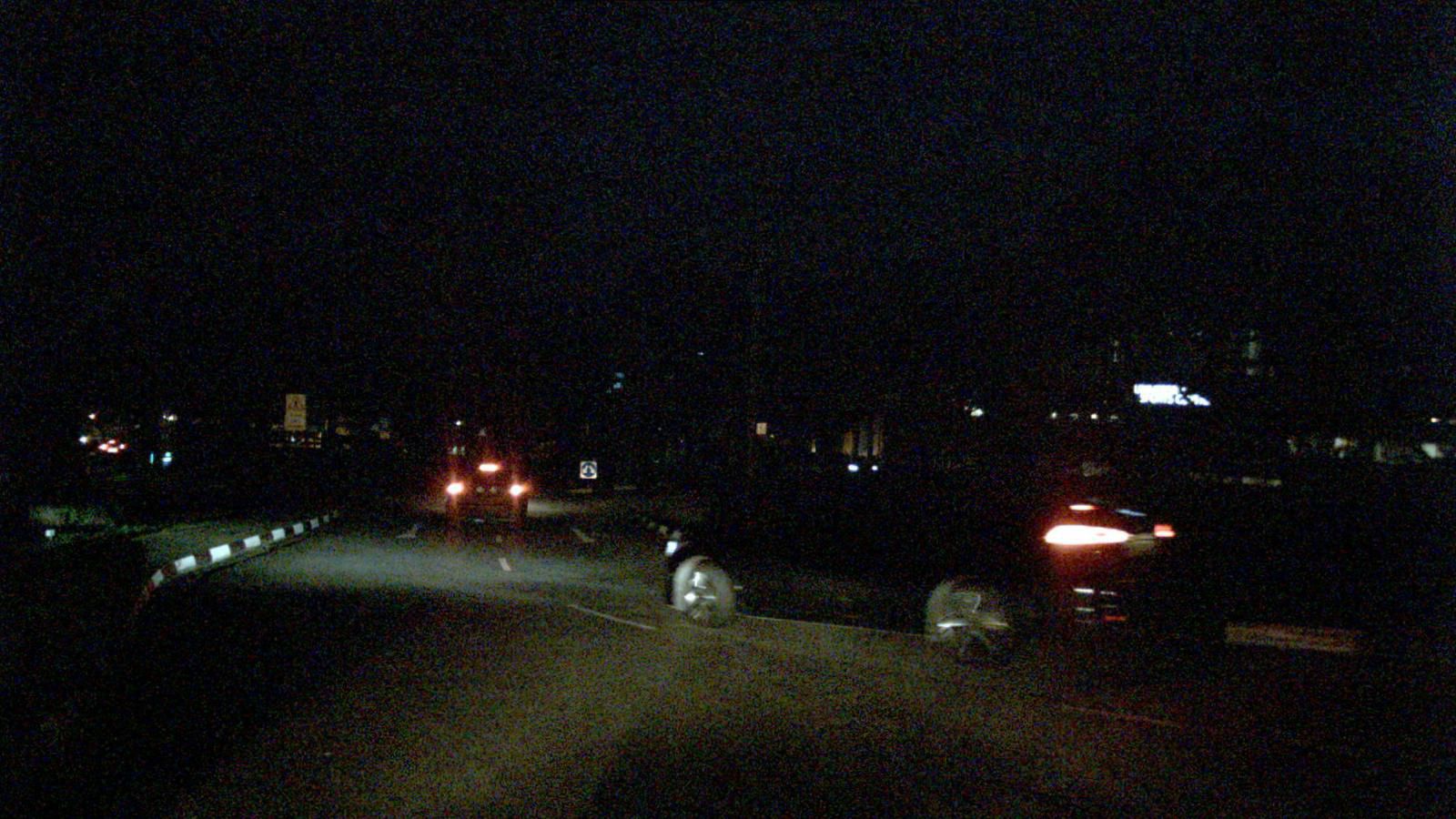} \\
		
		\vspace{-0.5mm}
		\rotatebox{90}{\hspace{0mm}}&
		\includegraphics[width=\turnheightnew]{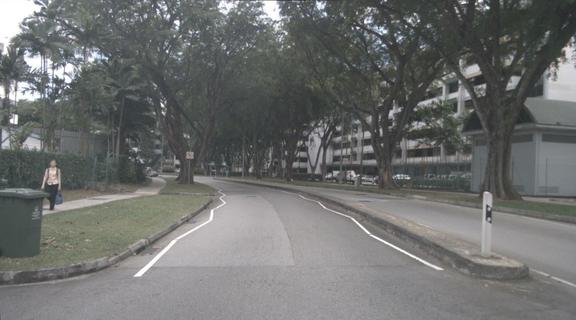} &
		\includegraphics[width=\turnheightnew]{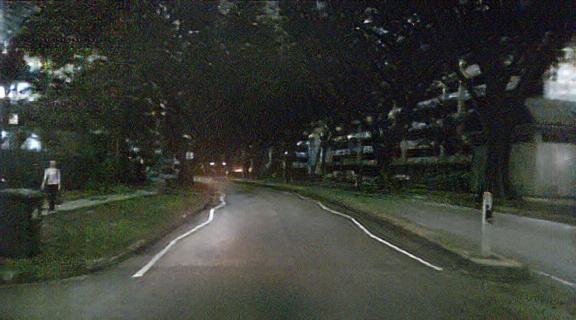} &
		\includegraphics[width=\turnheightnew]{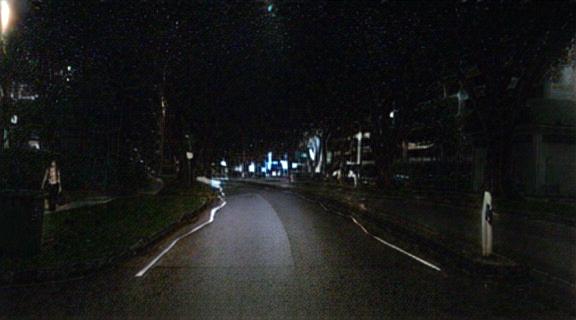} &
		\includegraphics[width=\turnheightnew]{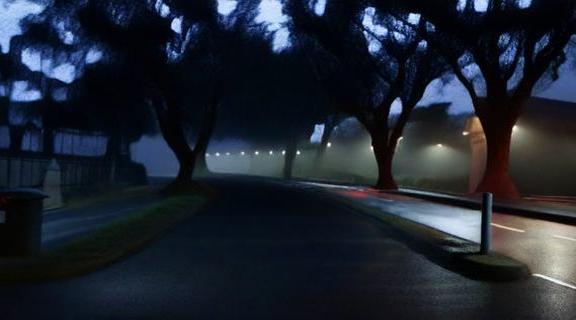} &
		\includegraphics[width=\turnheightnew]{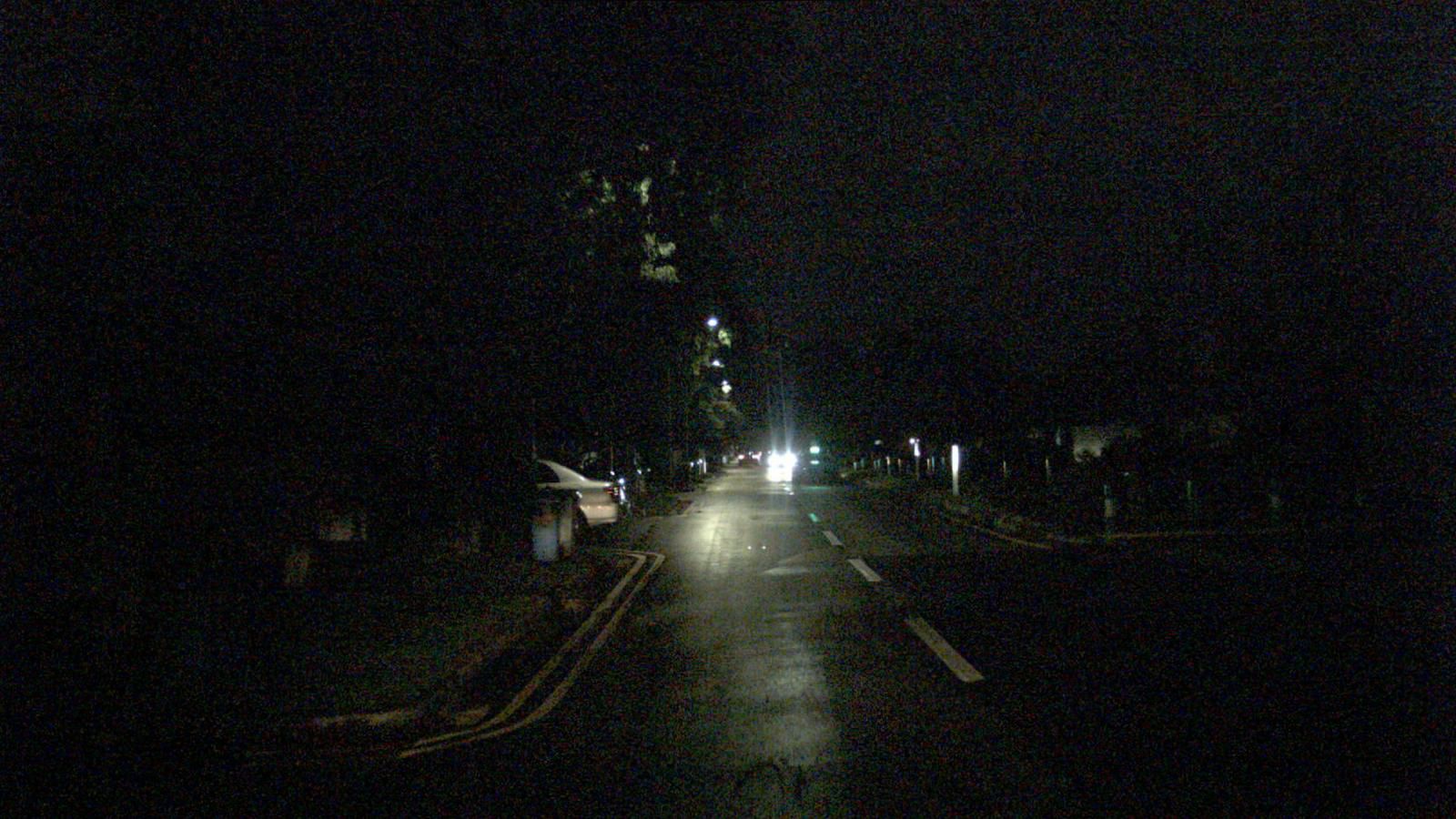} \\
		
		\vspace{0.5mm}
		\rotatebox{90}{\hspace{0mm}}&
		\includegraphics[width=\turnheightnew]{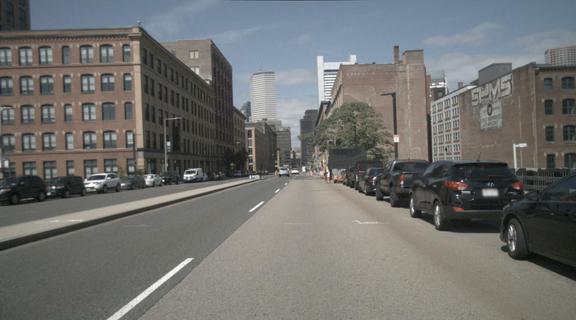} &
		\includegraphics[width=\turnheightnew]{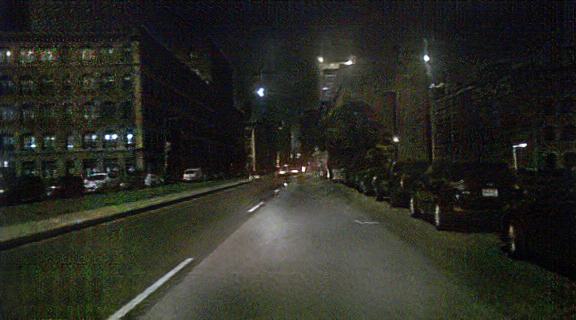} &
		\includegraphics[width=\turnheightnew]{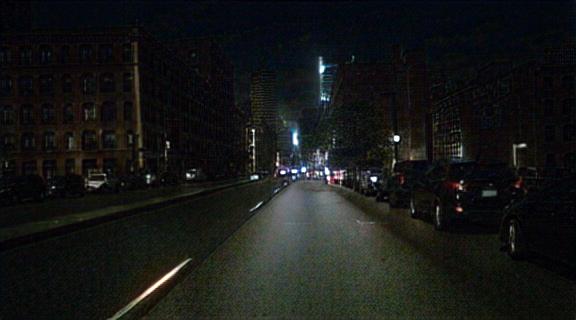} &
		\includegraphics[width=\turnheightnew]{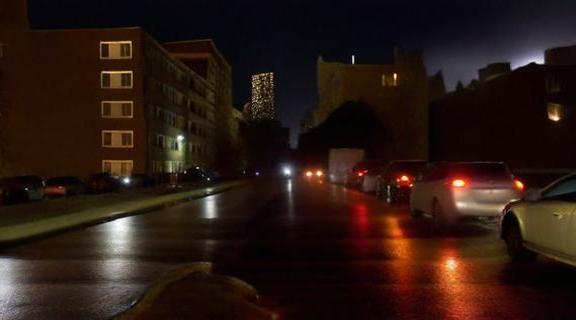} &
		\includegraphics[width=\turnheightnew]{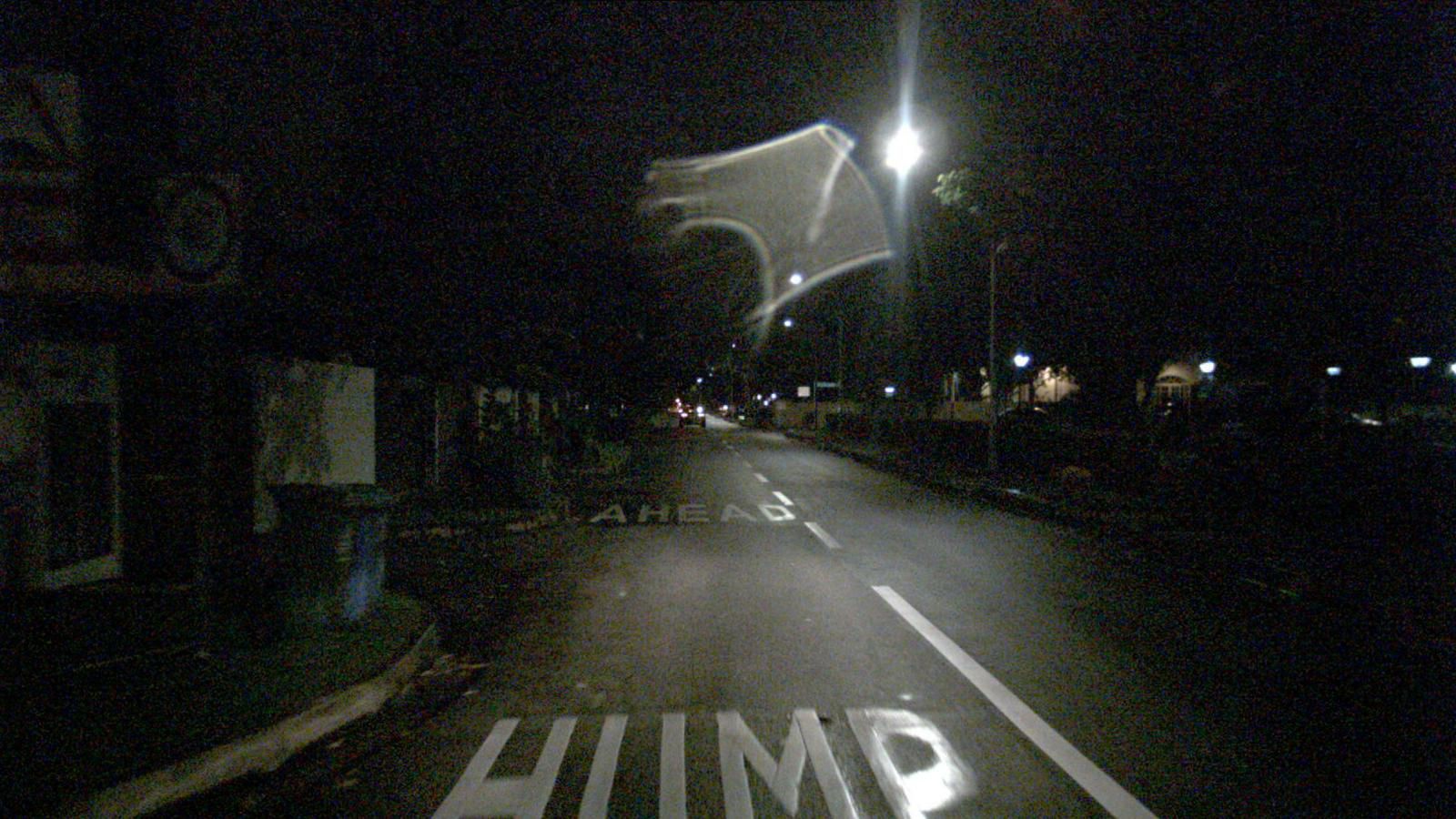} \\
		
		\vspace{-0.5mm}
		{\rotatebox{90}{\hspace{2mm}}} &
		\includegraphics[width=\turnheightnew,keepaspectratio]{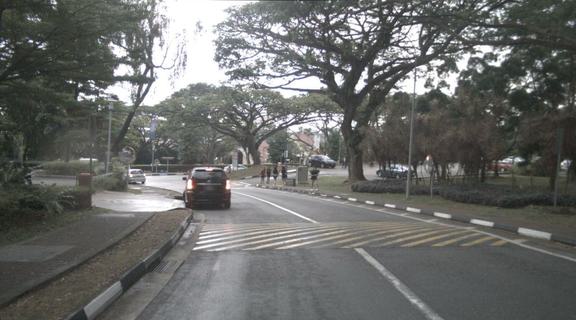} &
		\includegraphics[width=\turnheightnew,keepaspectratio]{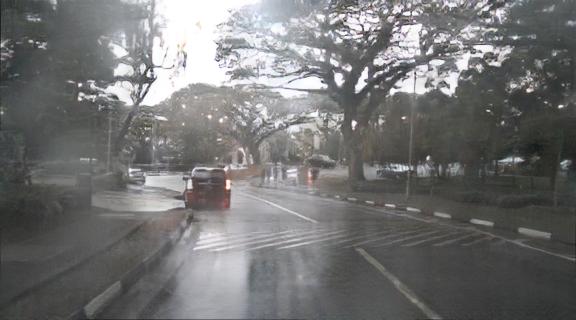} &
		\includegraphics[width=\turnheightnew,keepaspectratio]{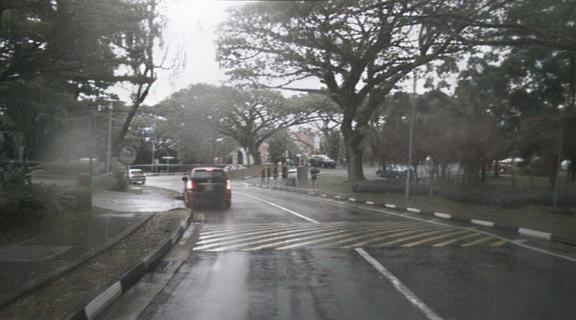} &
		\includegraphics[width=\turnheightnew,keepaspectratio]{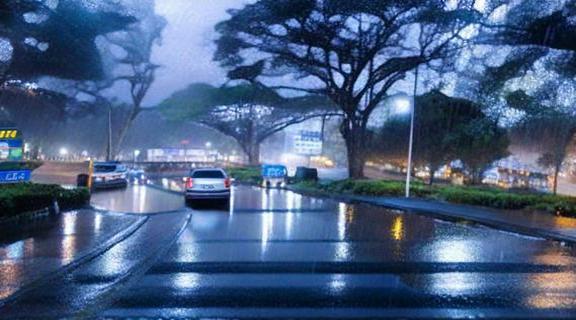} &
		\includegraphics[width=\turnheightnew,keepaspectratio]{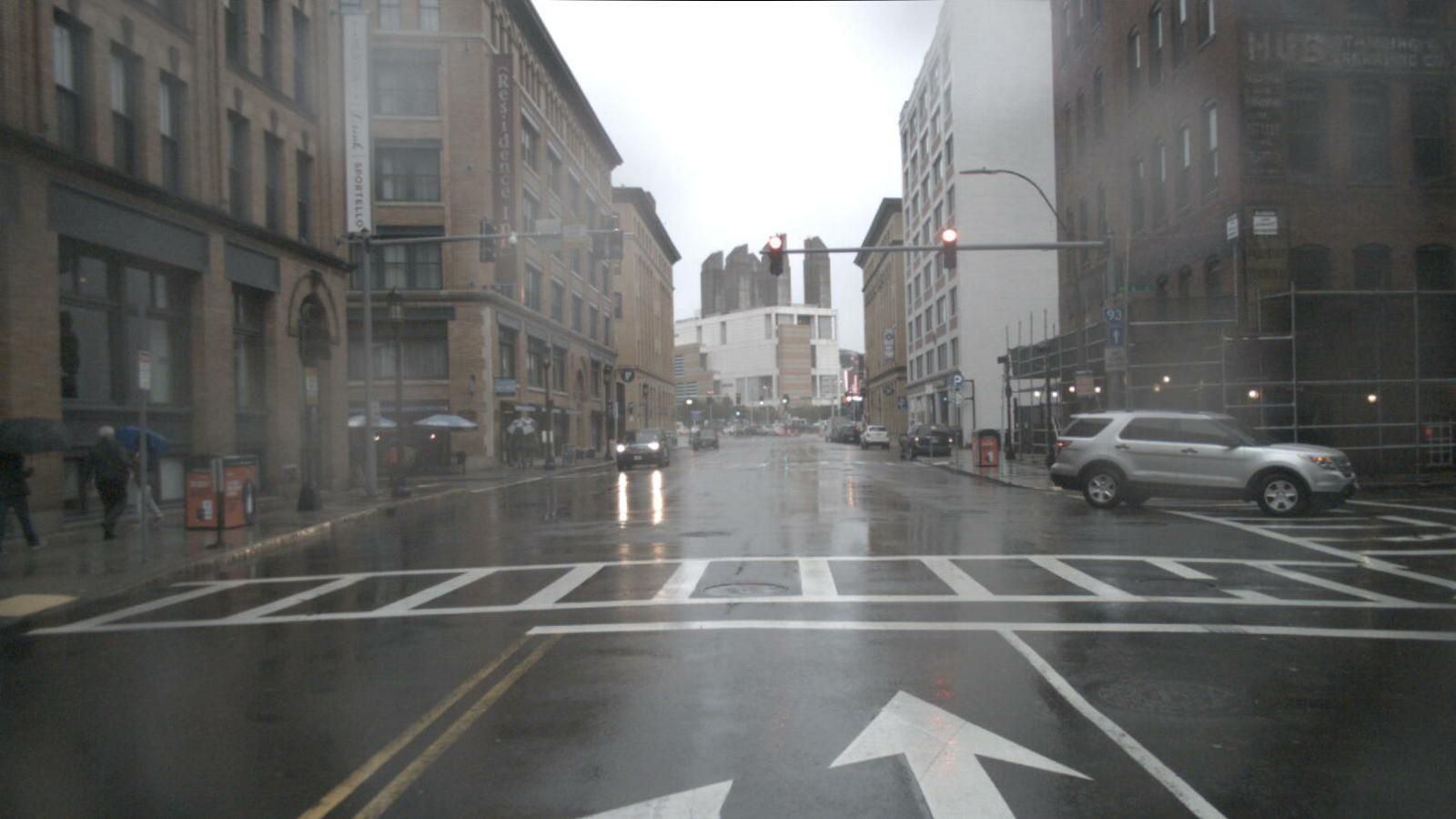} \\
		
		\vspace{-0.5mm}
		\rotatebox{90}{\hspace{0mm}}&
		\includegraphics[width=\turnheightnew]{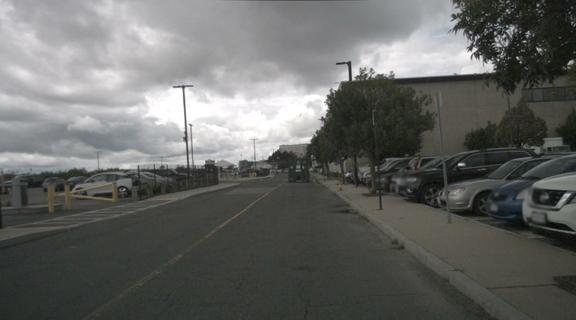} &
		\includegraphics[width=\turnheightnew]{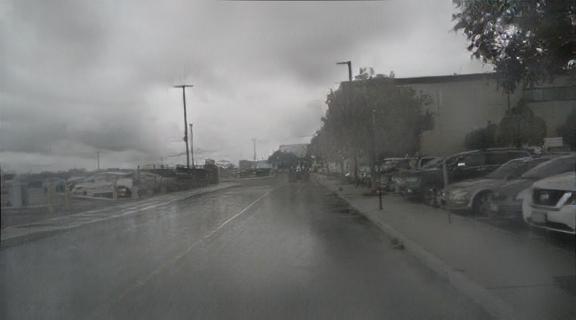} &
		\includegraphics[width=\turnheightnew]{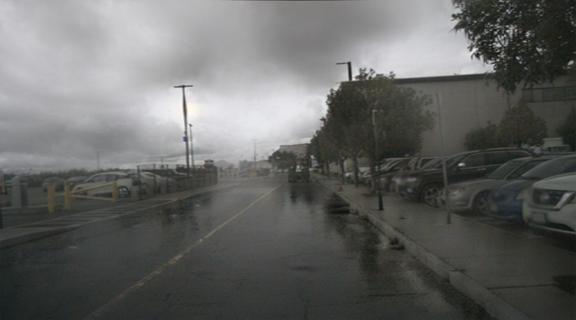} &
		\includegraphics[width=\turnheightnew]{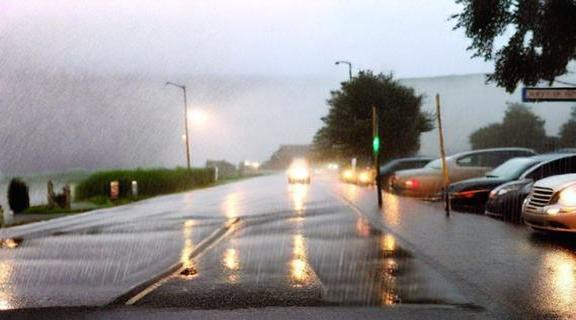} &
		\includegraphics[width=\turnheightnew]{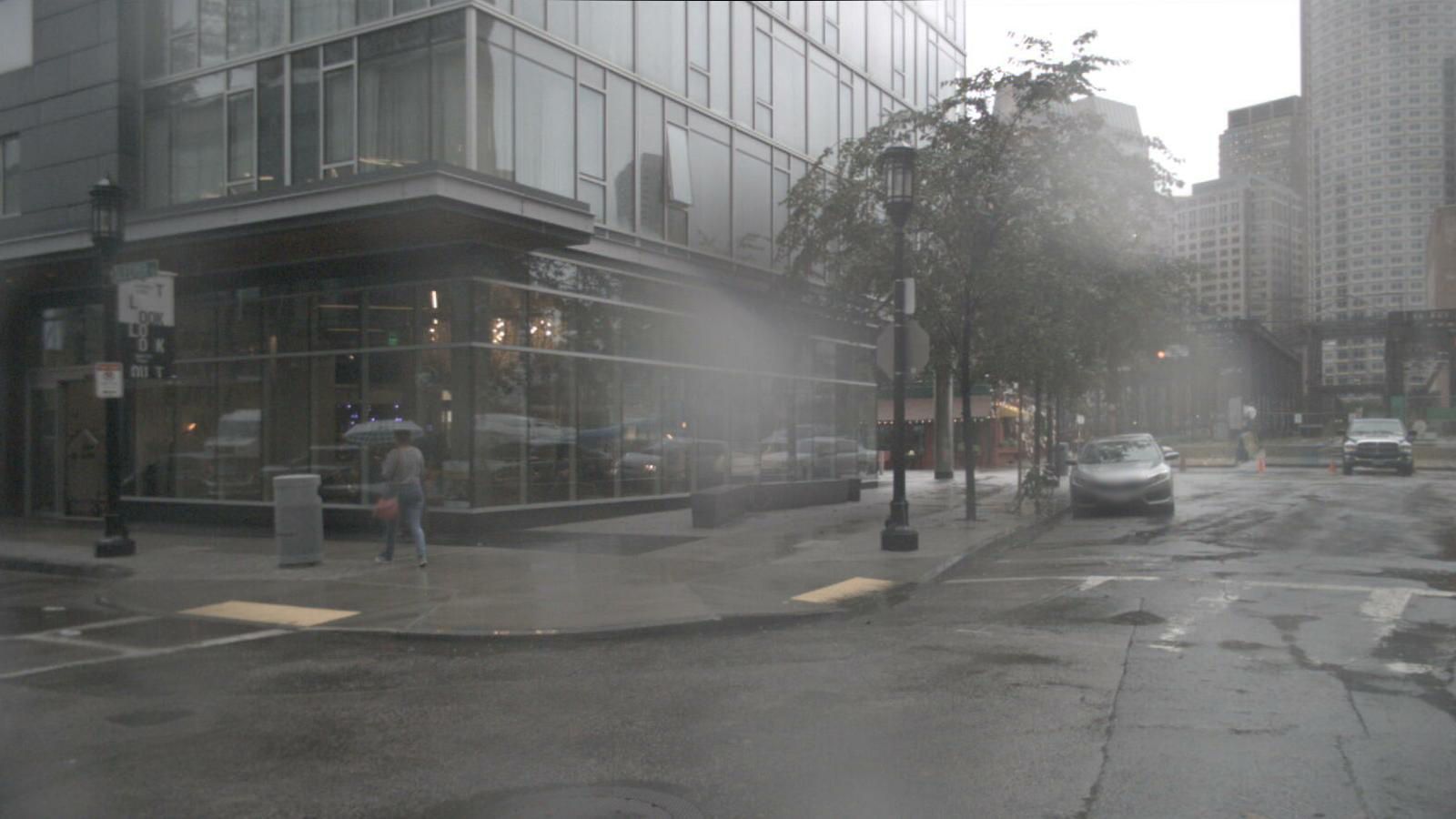} \\
		
		\vspace{-0.5mm}
		\rotatebox{90}{\hspace{0mm}}&
		\includegraphics[width=\turnheightnew]{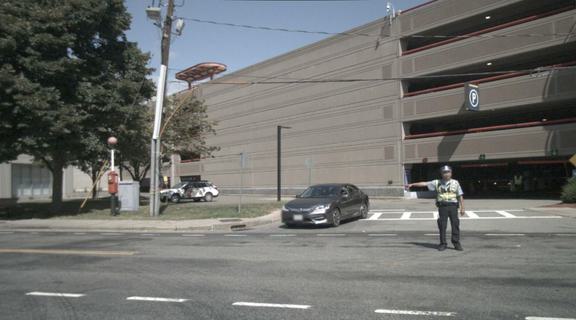} &
		\includegraphics[width=\turnheightnew]{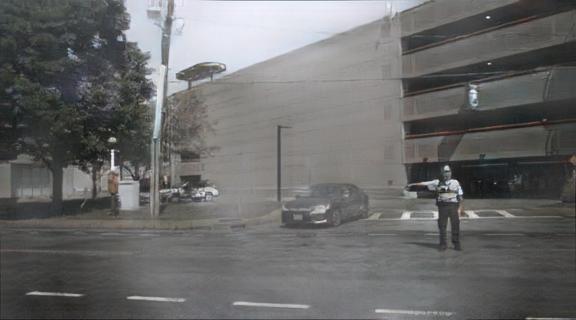} &
		\includegraphics[width=\turnheightnew]{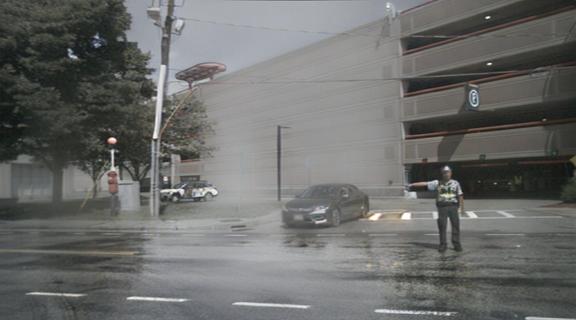} &
		\includegraphics[width=\turnheightnew]{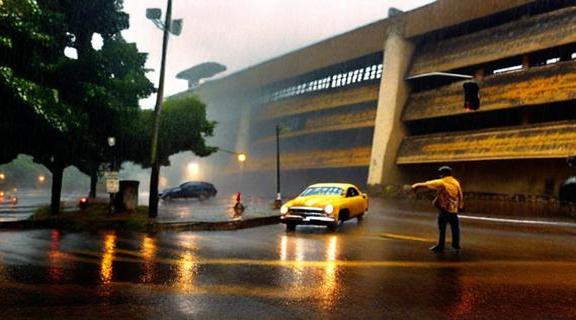} &
		\includegraphics[width=\turnheightnew]{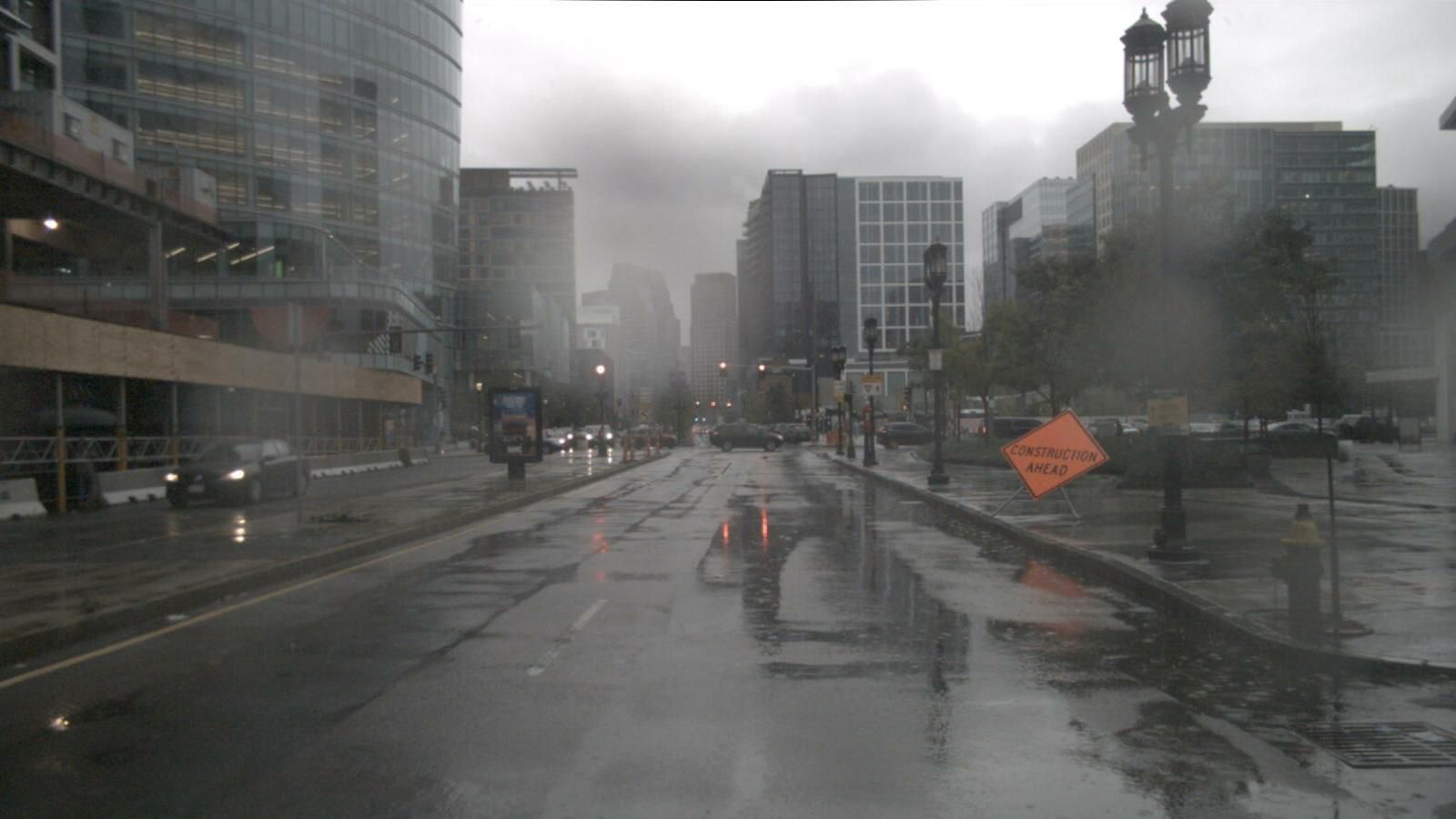} \\
		
		\vspace{-0.5mm}
		\rotatebox{90}{\hspace{0mm}}&
		\includegraphics[width=\turnheightnew]{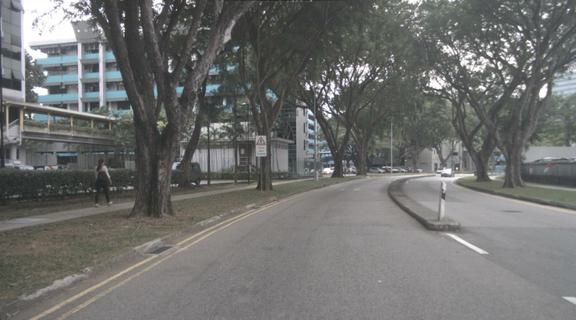} &
		\includegraphics[width=\turnheightnew]{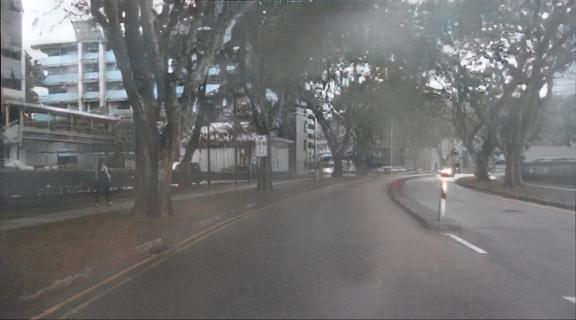} &
		\includegraphics[width=\turnheightnew]{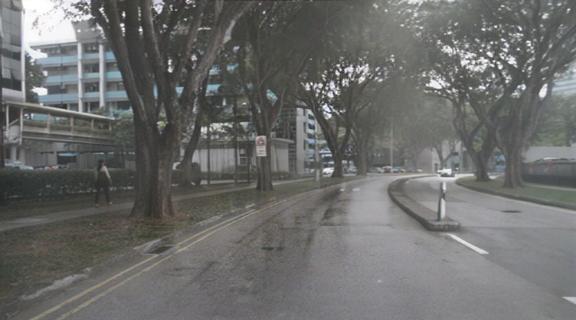} &
		\includegraphics[width=\turnheightnew]{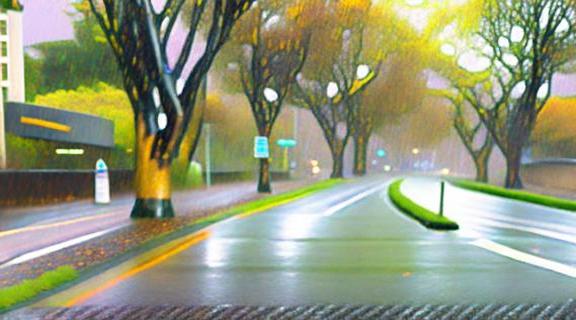} &
		\includegraphics[width=\turnheightnew]{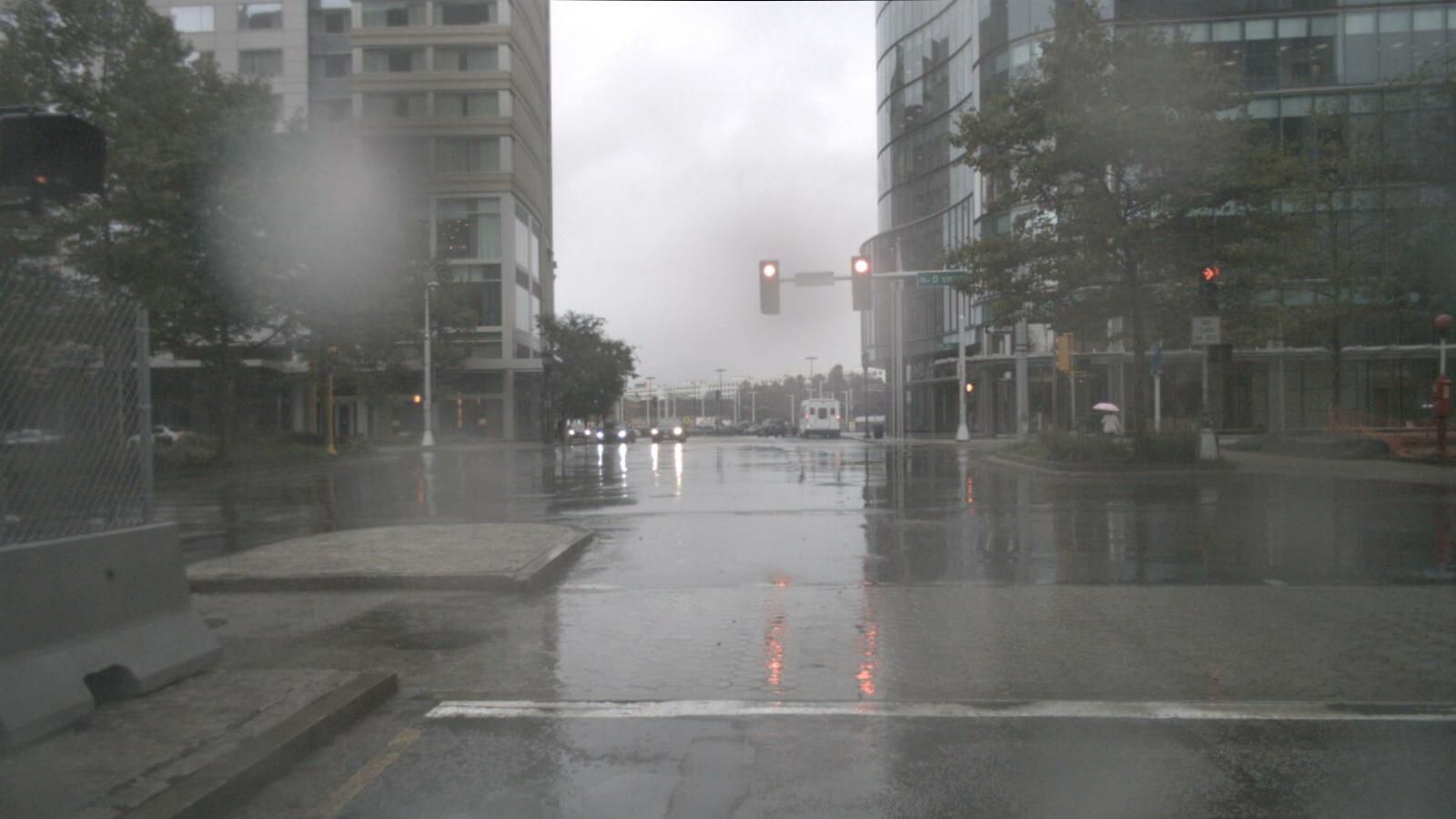} \\
		
		\vspace{-0.5mm}
		\rotatebox{90}{\hspace{0mm}}&
		\includegraphics[width=\turnheightnew]{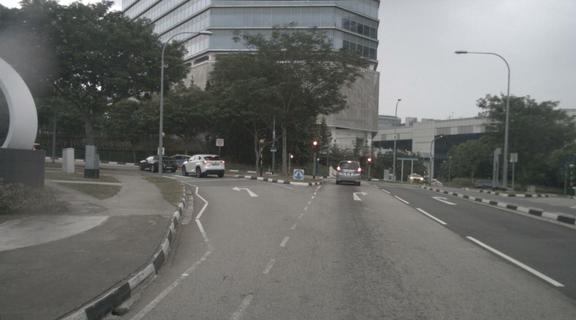} &
		\includegraphics[width=\turnheightnew]{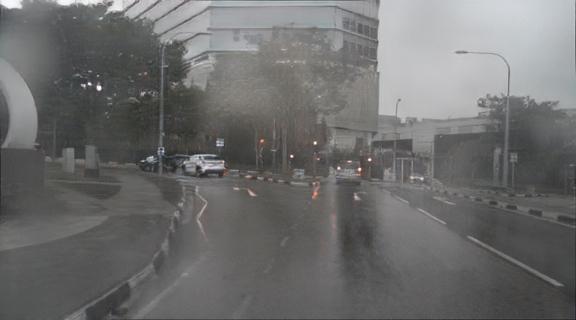} &
		\includegraphics[width=\turnheightnew]{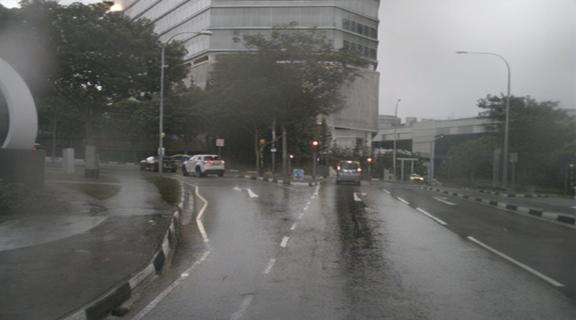} &
		\includegraphics[width=\turnheightnew]{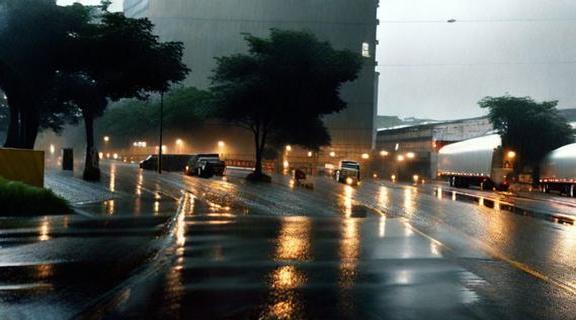} &
		\includegraphics[width=\turnheightnew]{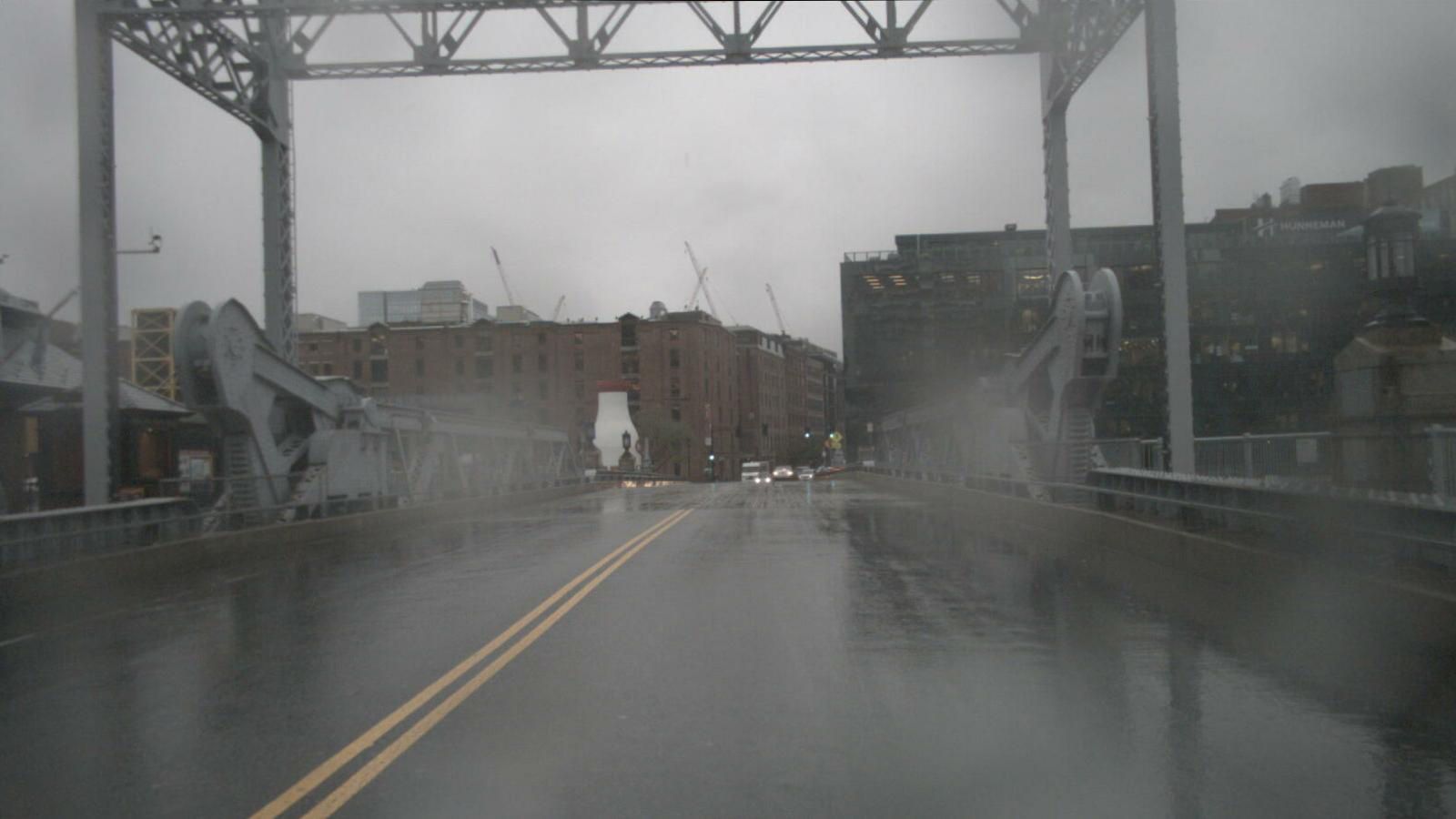} \\
		\multicolumn{1}{c}{} & 
		\multicolumn{1}{c}{Clear-day} & 
		\multicolumn{1}{c}{ForkGan} & 
		\multicolumn{1}{c}{CycleGan-Turbo} & 
		\multicolumn{1}{c}{T2I-Adapter} & 
		\multicolumn{1}{c}{nuScenes} \\
	\end{tabular}
	\caption{Data generation results for nuScenes} 
	\label{fig:6}
\end{figure*}

For the nuScenes, we show result from three different translators in the Fig~\ref{fig:6}. The first and last columns correspond to real samples and challenging scene samples from the nuScenes dataset, respectively. In the night scene, compared to the ForkGan translation methods, the images generated by CycleGAN-Turbo are closer to the real lighting conditions in nuScenes. GAN-based methods often introduce additional light sources and suffer from overexposure issues. Despite this, both methods generate more realistic images than the T2I-Adapter~\cite{mou2024t2i}. The T2I-Adapter introduces significant style discrepancies between translated and real images and creates inconsistencies between the content of translated images and their original counterparts, thereby adding challenging variations to the training process. In rainy scenes, the CycleGAN-Turbo-based method can generate more realistic scenes compared to the GAN-based method. For example, CycleGAN-Turbo can learn to add raindrop effects and simulate reflections on water surfaces, which is beneficial for training a more robust depth estimation model.

\begin{figure*}[htbp]
	\centering
	\newcommand{\turnheightnew}{0.35\columnwidth}
	\begin{tabular}{@{\hskip 1mm}c@{\hskip 1mm}c@{\hskip 1mm}c@{\hskip 1mm}c@{\hskip 1mm}c@{\hskip 1mm}c@{}}
		\vspace{-0.5mm}
		{\rotatebox{90}{\hspace{2mm}}} &
	   \includegraphics[width=\turnheightnew]{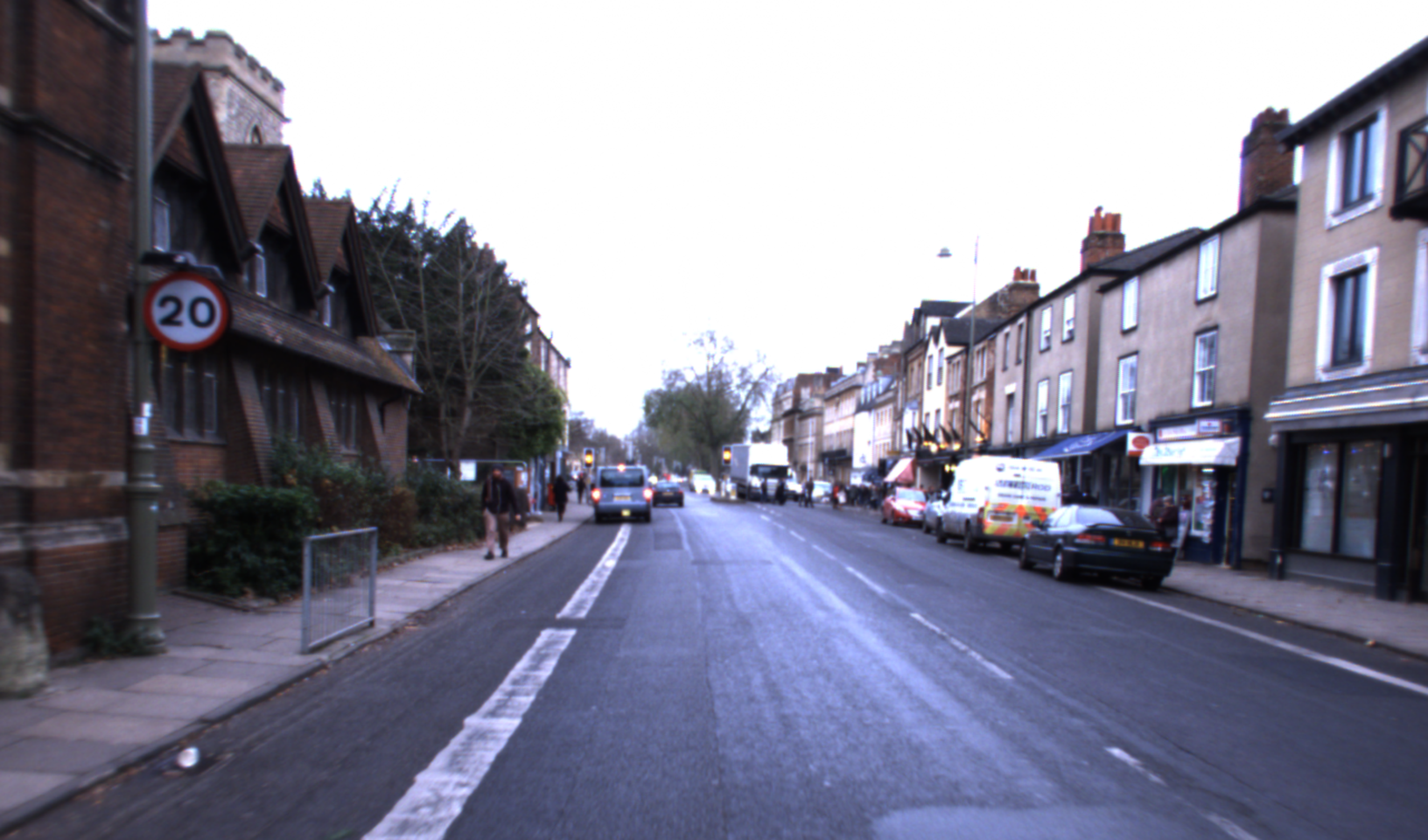} &
	   \includegraphics[width=\turnheightnew]{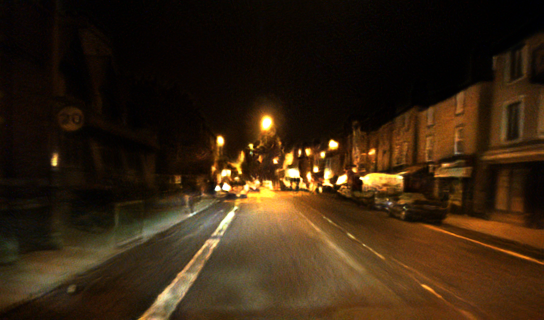} &
	   \includegraphics[width=\turnheightnew]{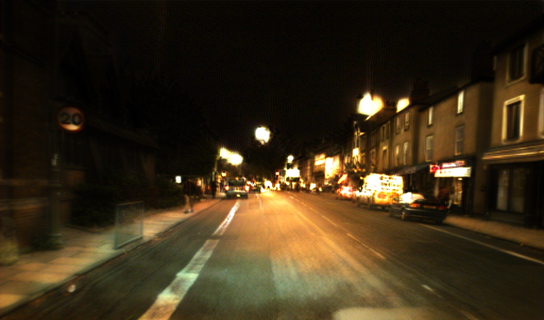} &
	   \includegraphics[width=\turnheightnew]{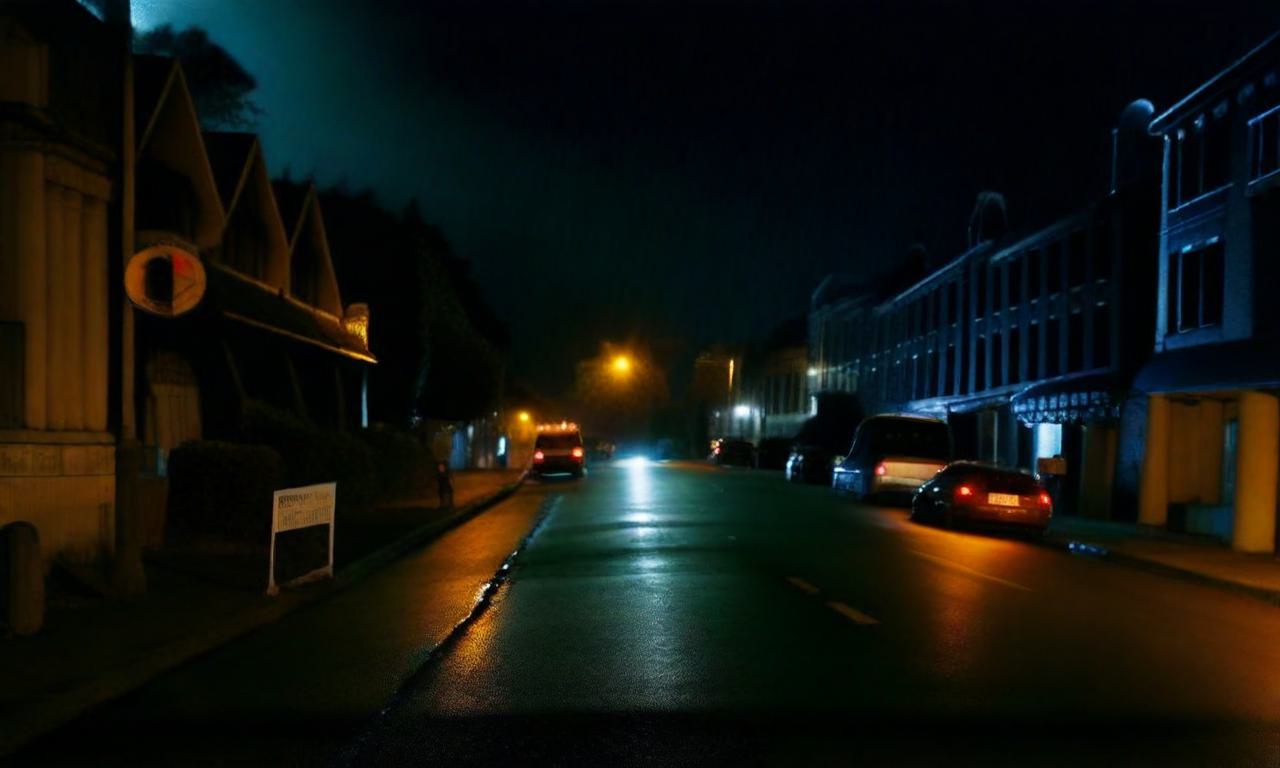} &
	   \includegraphics[width=\turnheightnew]{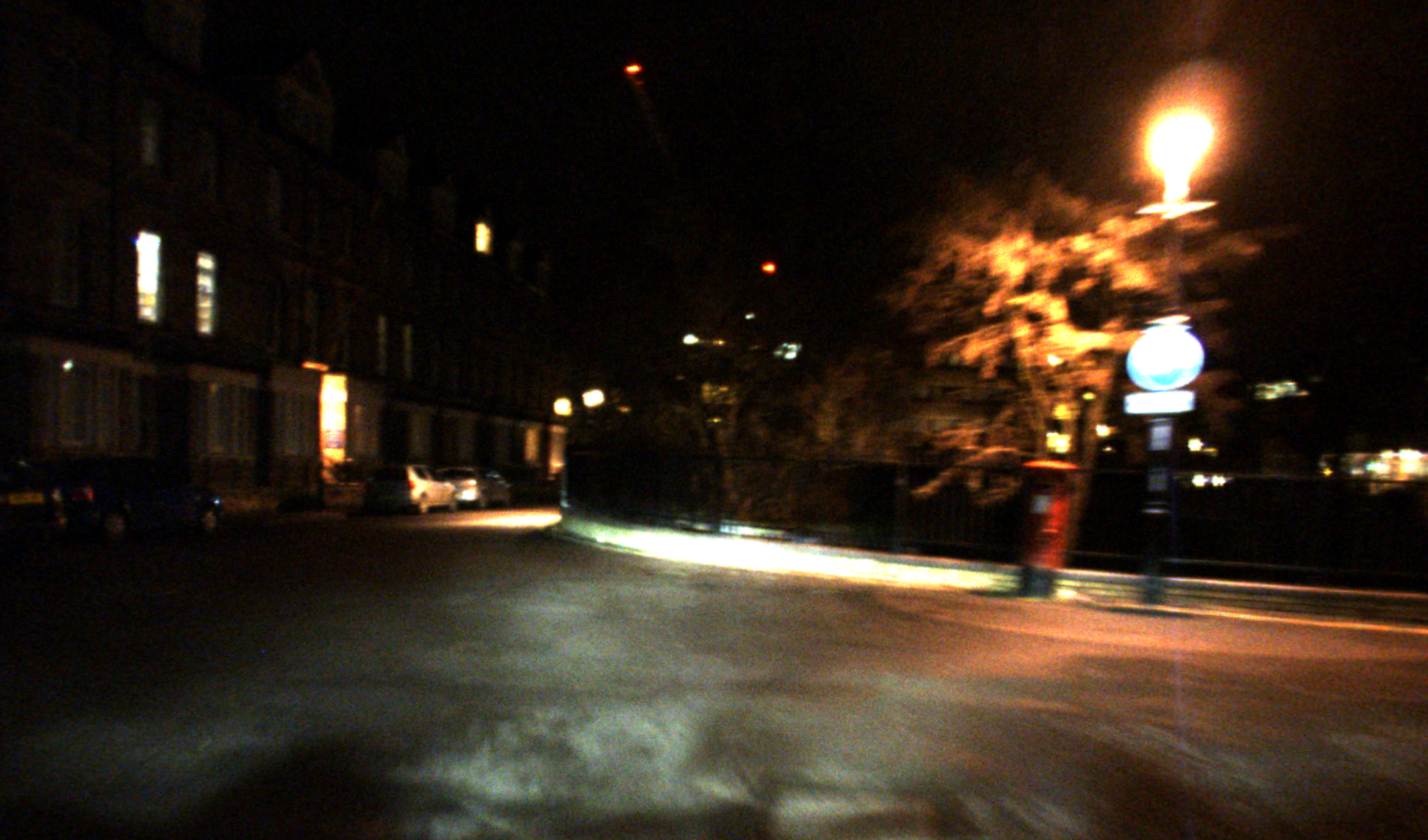} \\
		
		\vspace{-0.5mm}
		\rotatebox{90}{\hspace{0mm}}&
	   \includegraphics[width=\turnheightnew]{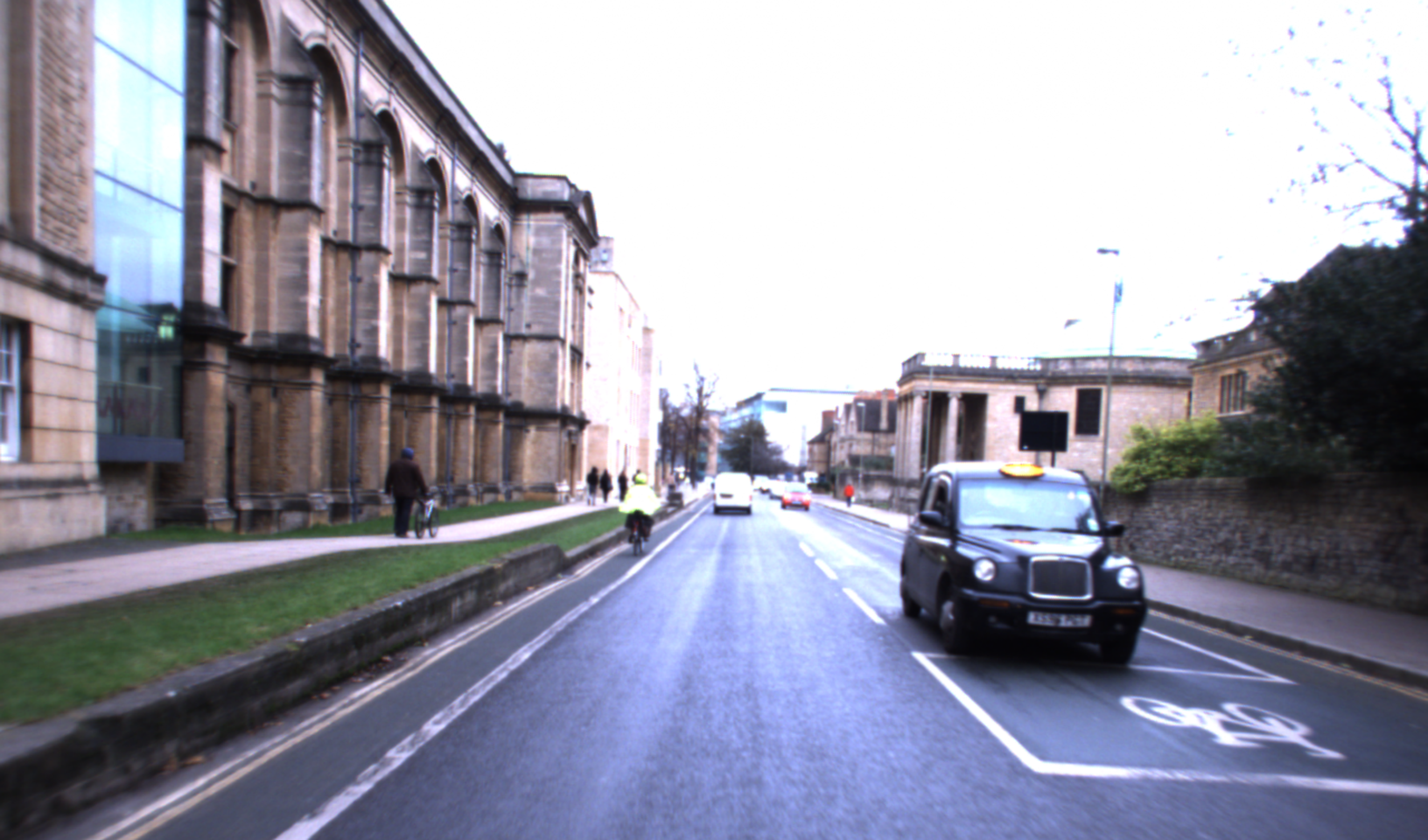} &
	   \includegraphics[width=\turnheightnew]{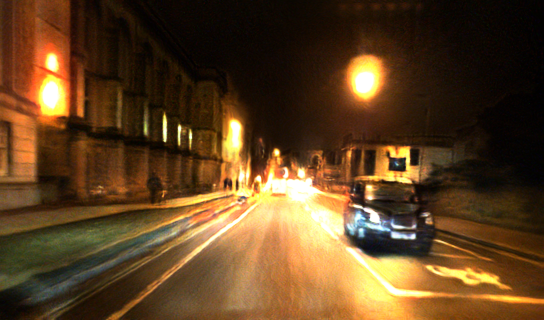} &
	   \includegraphics[width=\turnheightnew]{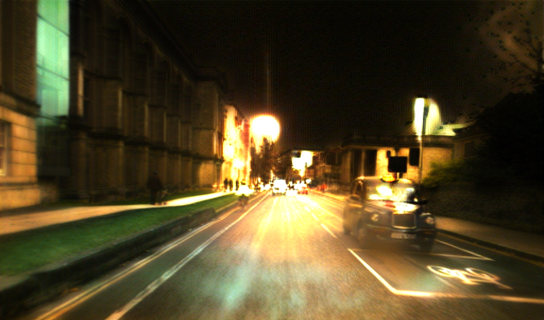} &
	   \includegraphics[width=\turnheightnew]{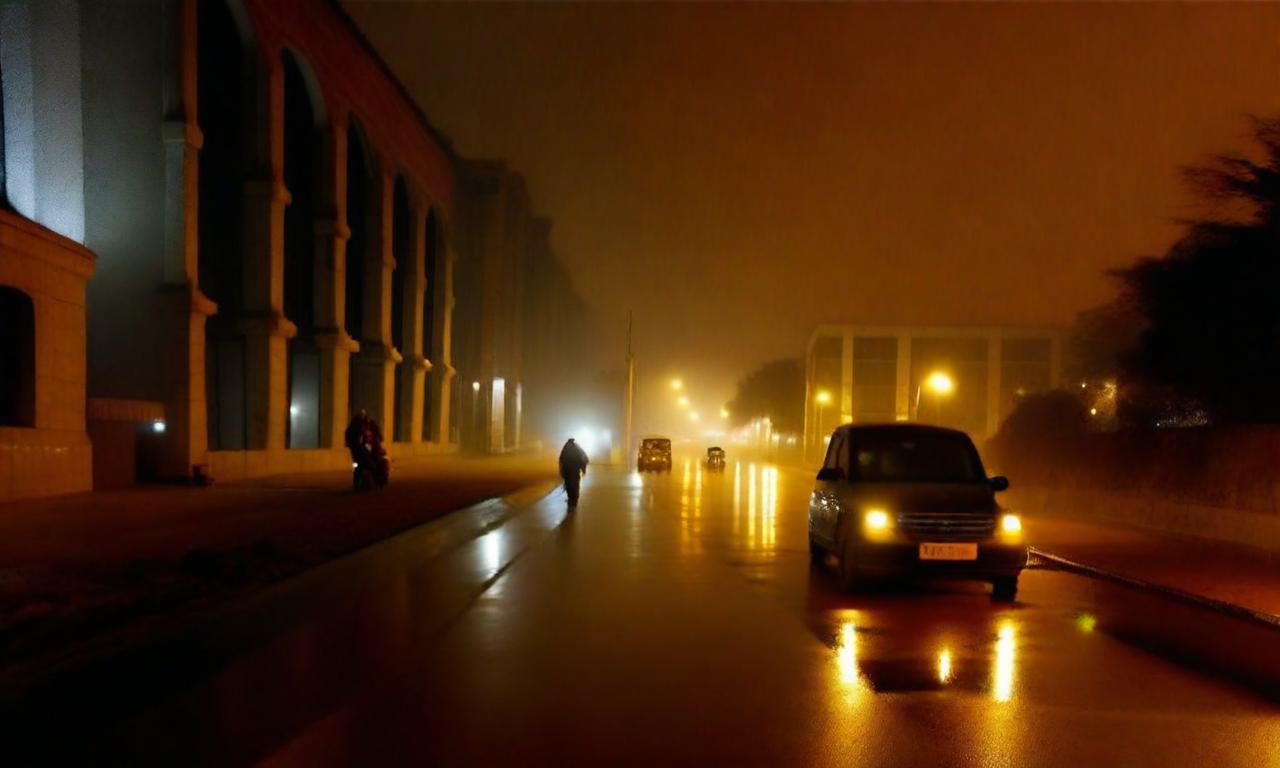} &
	   \includegraphics[width=\turnheightnew]{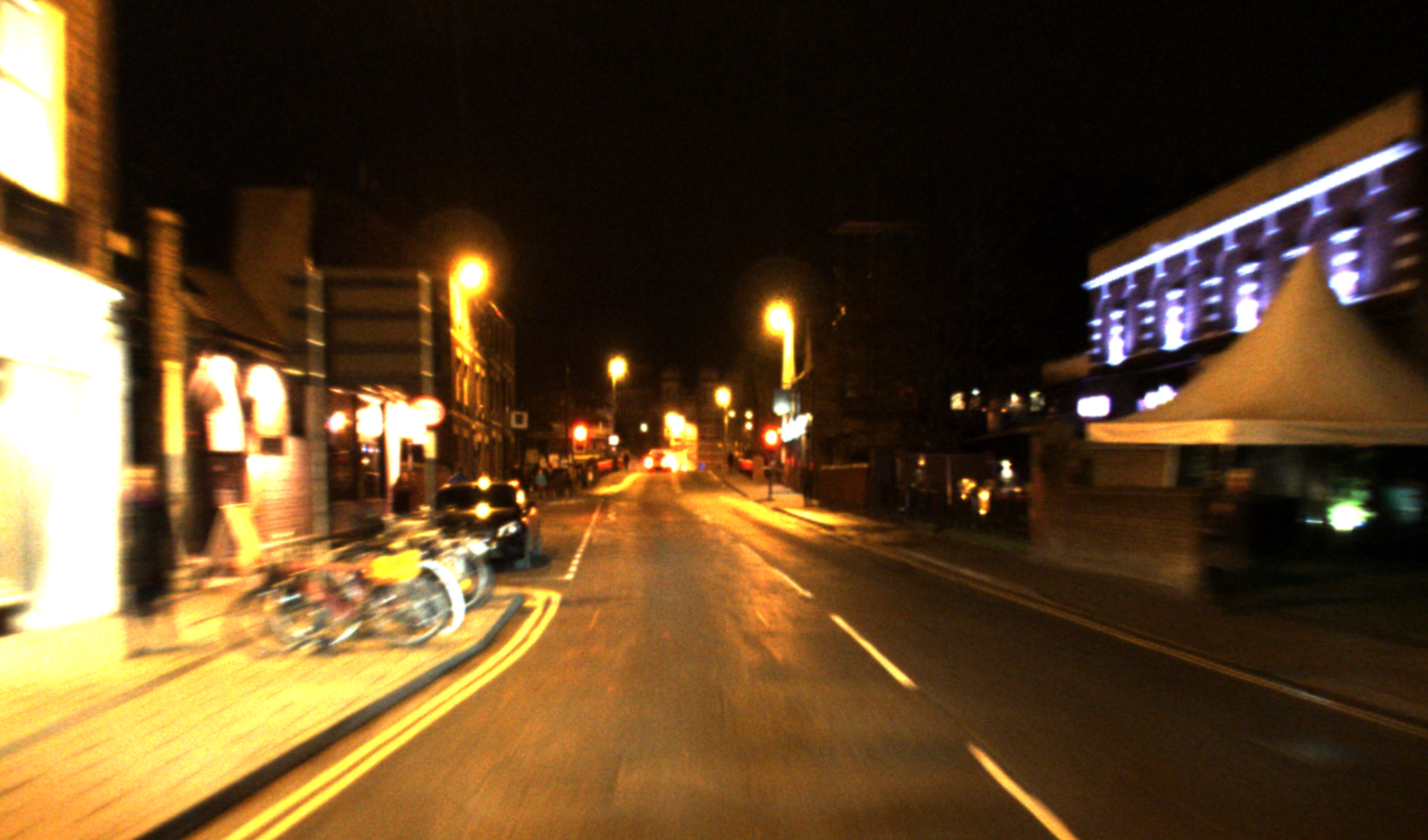} \\
		
		\vspace{-0.5mm}
		\rotatebox{90}{\hspace{0mm}}&
	   \includegraphics[width=\turnheightnew]{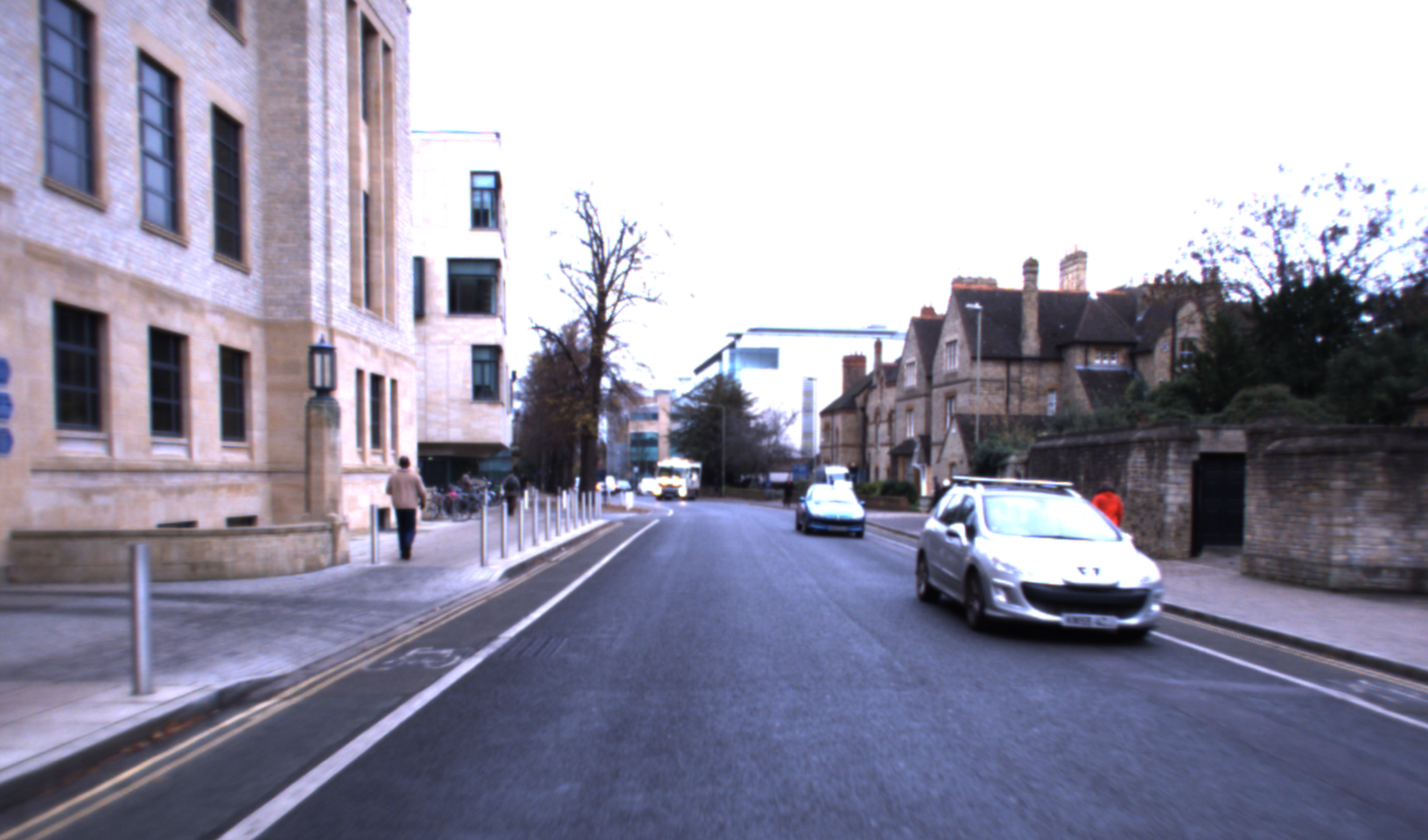} &
	   \includegraphics[width=\turnheightnew]{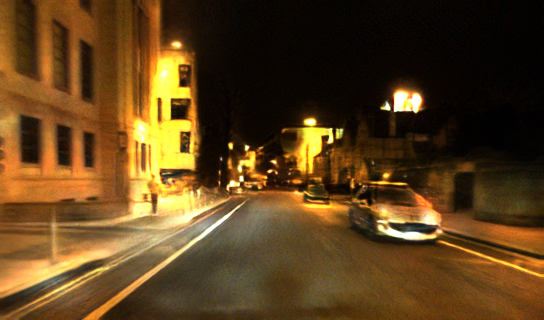} &
	   \includegraphics[width=\turnheightnew]{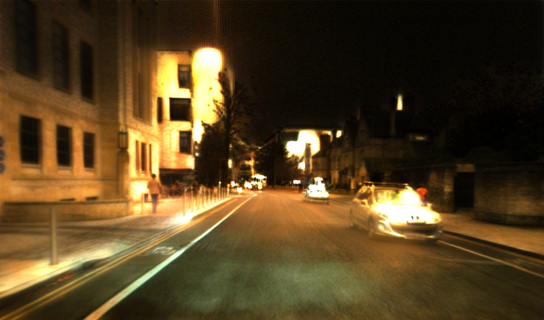} &
	   \includegraphics[width=\turnheightnew]{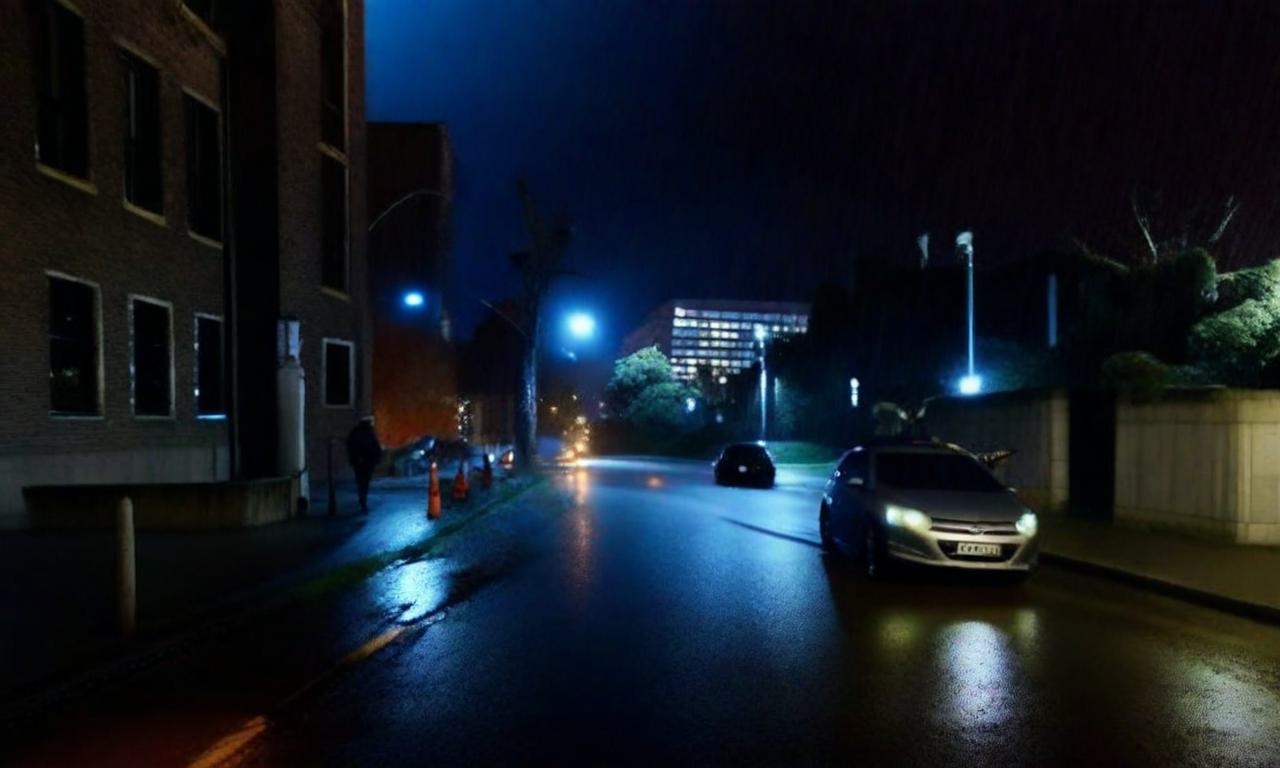} &
	   \includegraphics[width=\turnheightnew]{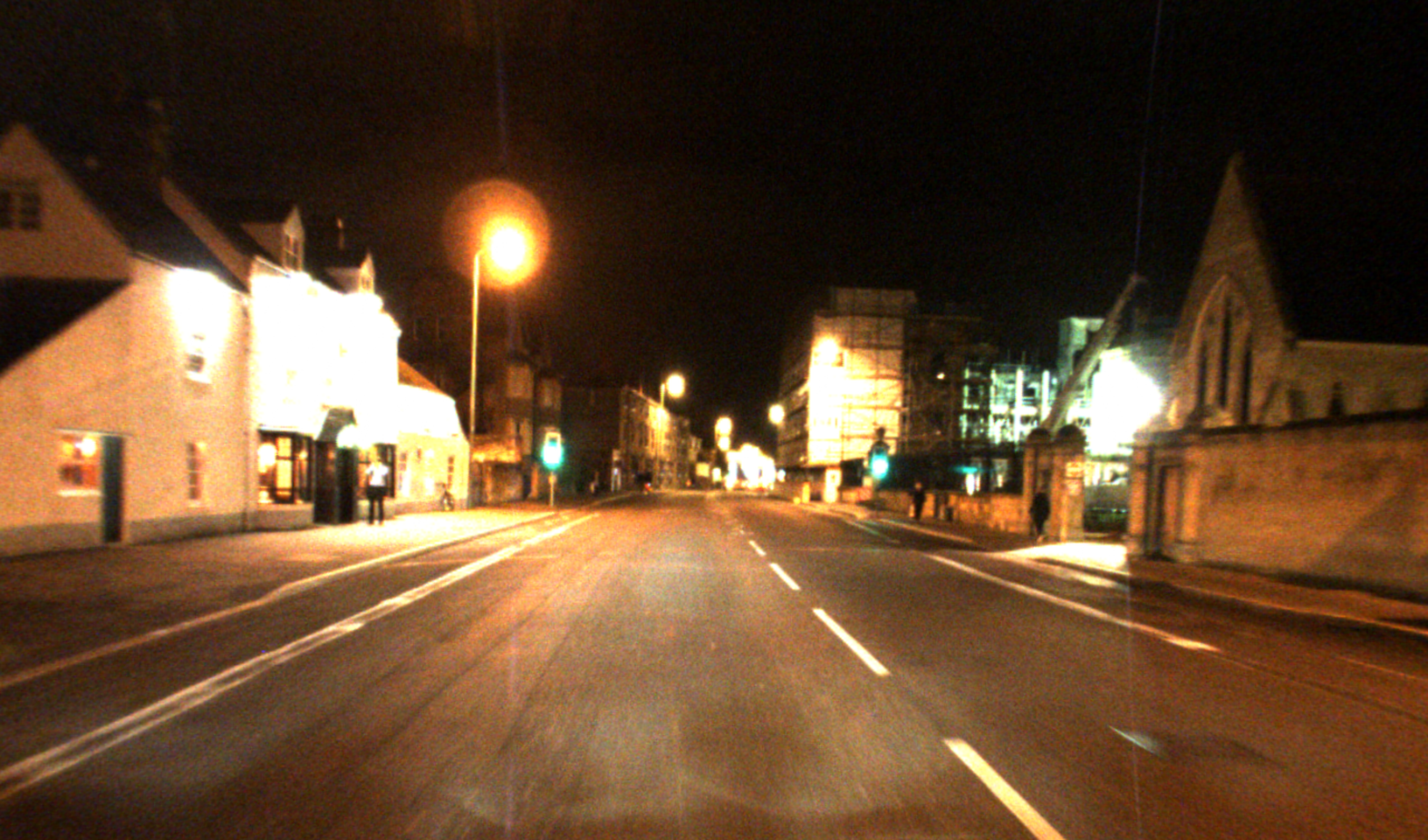} \\
		
		\vspace{-0.5mm}
		\rotatebox{90}{\hspace{0mm}}&
	   \includegraphics[width=\turnheightnew]{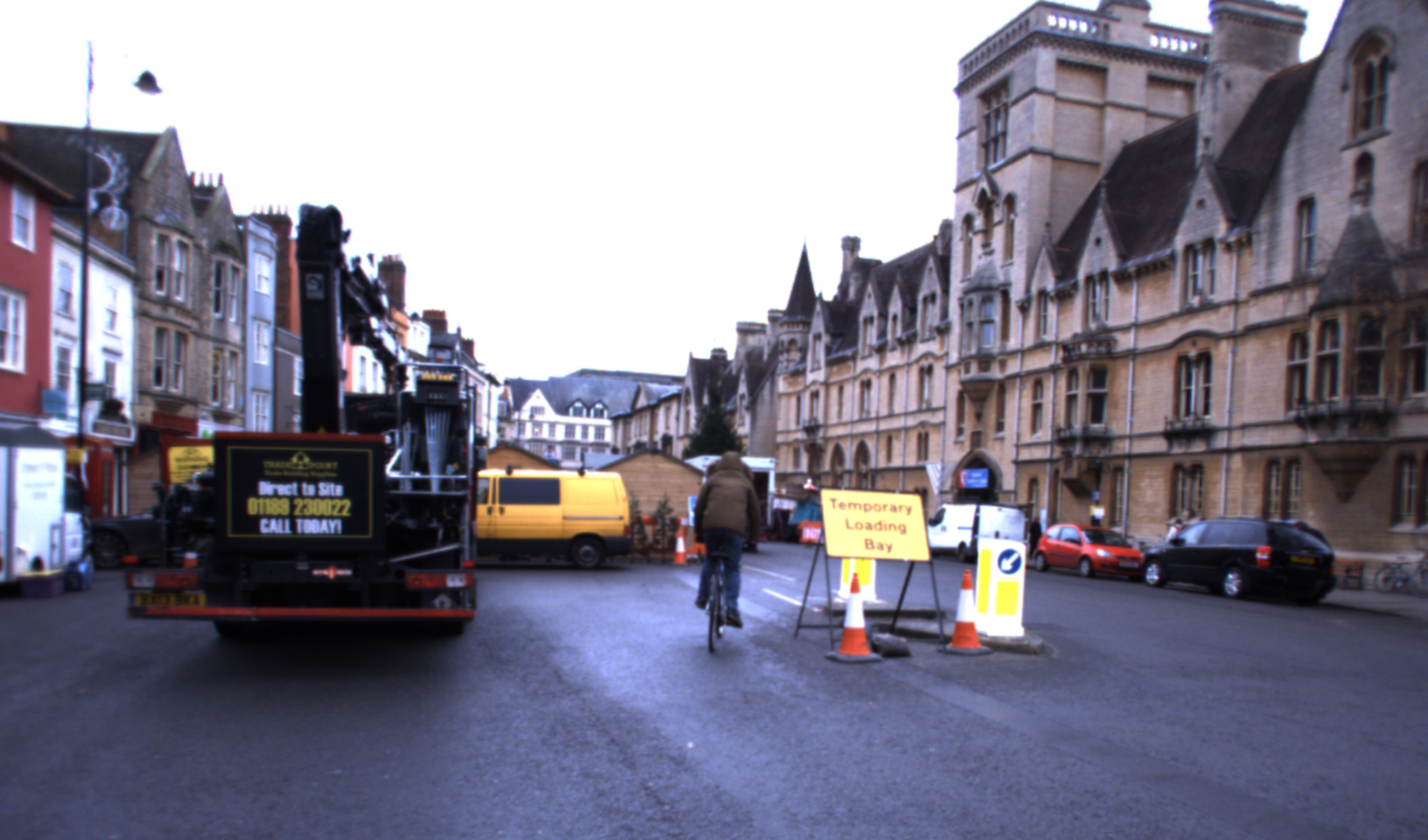} &
	   \includegraphics[width=\turnheightnew]{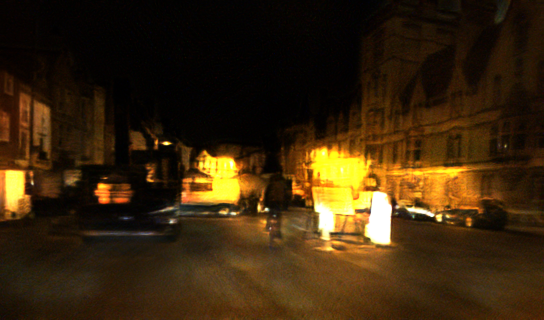} &
	   \includegraphics[width=\turnheightnew]{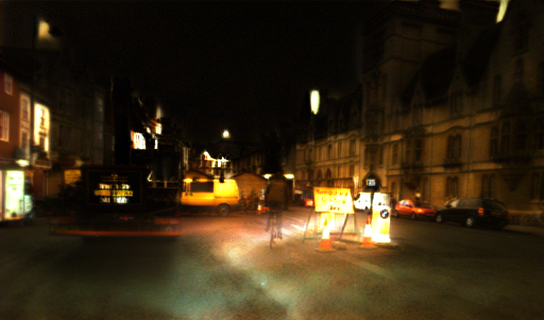} &
	   \includegraphics[width=\turnheightnew]{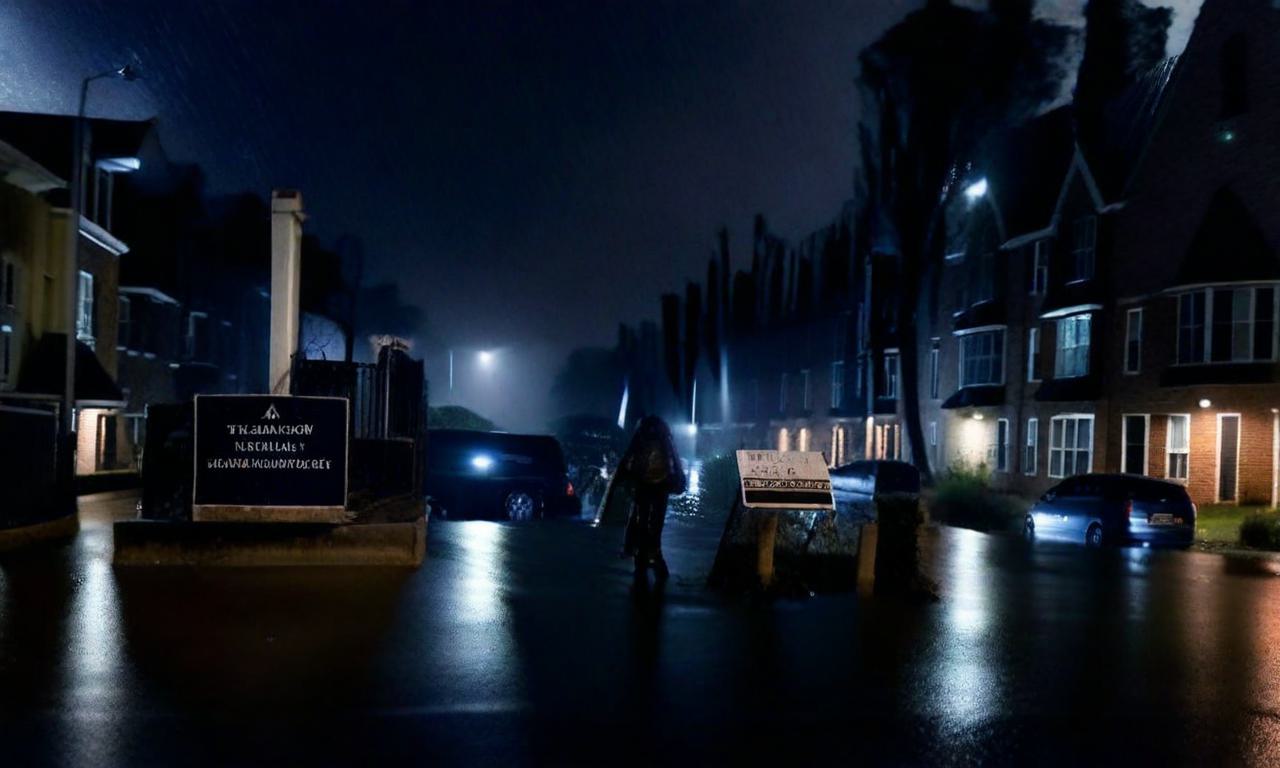} &
	   \includegraphics[width=\turnheightnew]{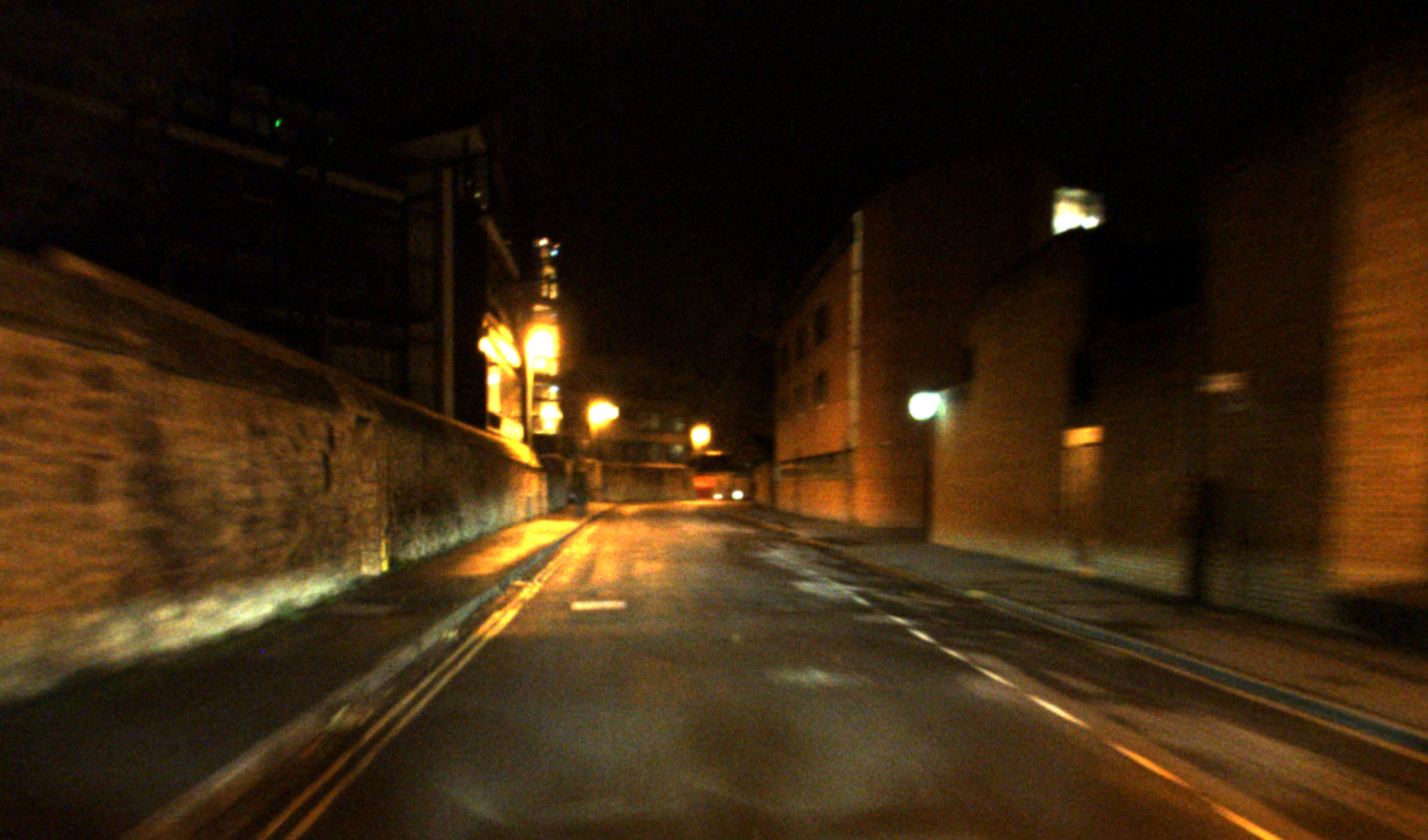} \\
		
		\vspace{-0.5mm}
		\rotatebox{90}{\hspace{0mm}}&
	   \includegraphics[width=\turnheightnew]{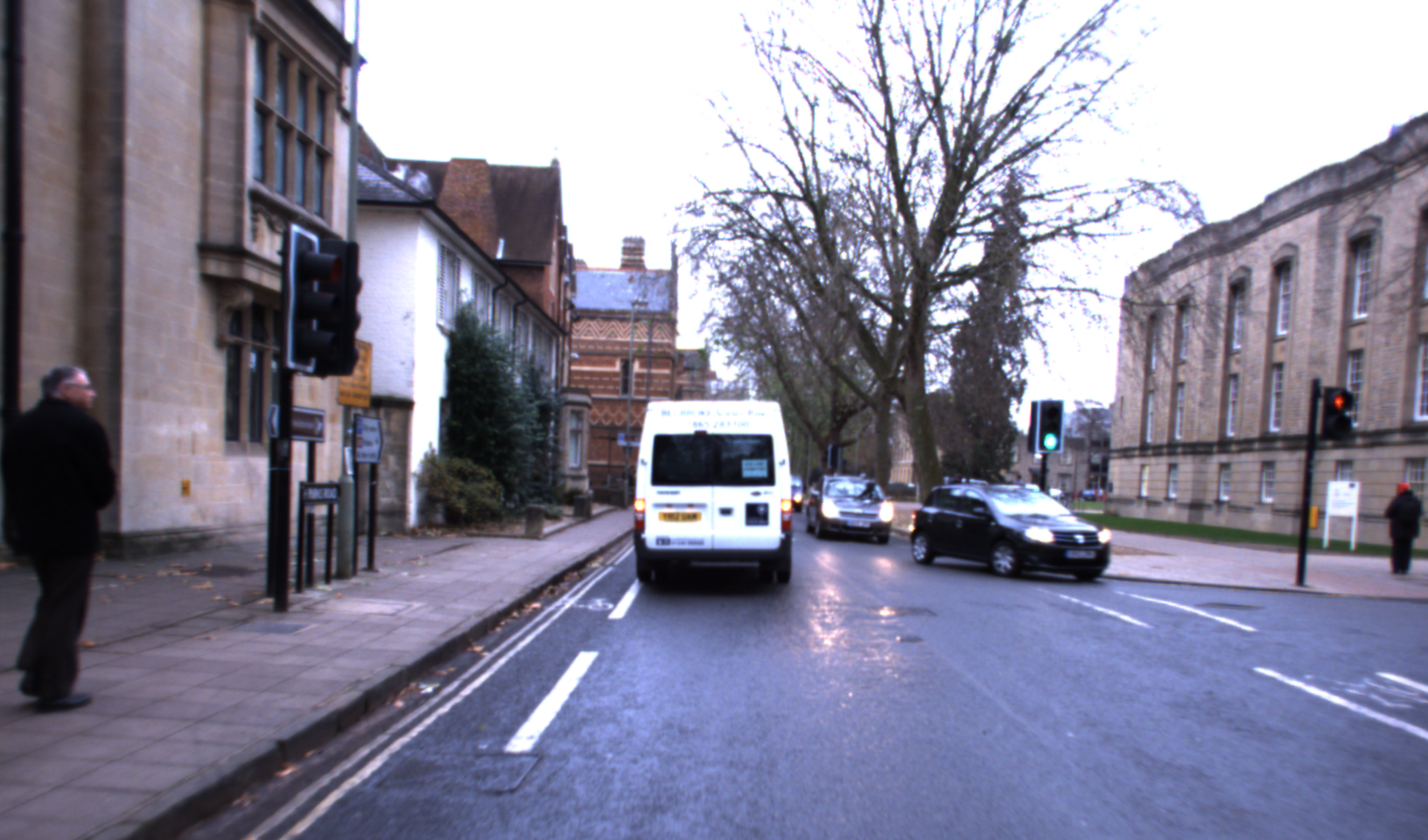} &
	   \includegraphics[width=\turnheightnew]{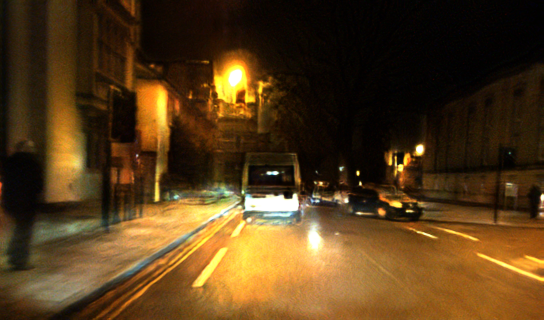} &
	   \includegraphics[width=\turnheightnew]{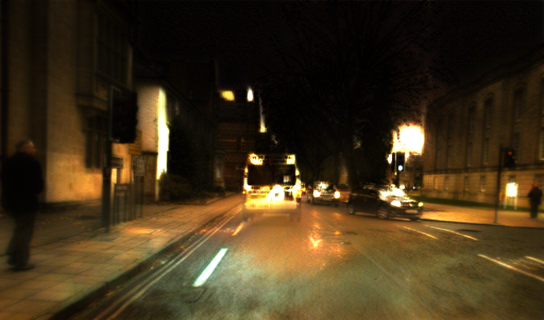} &
	   \includegraphics[width=\turnheightnew]{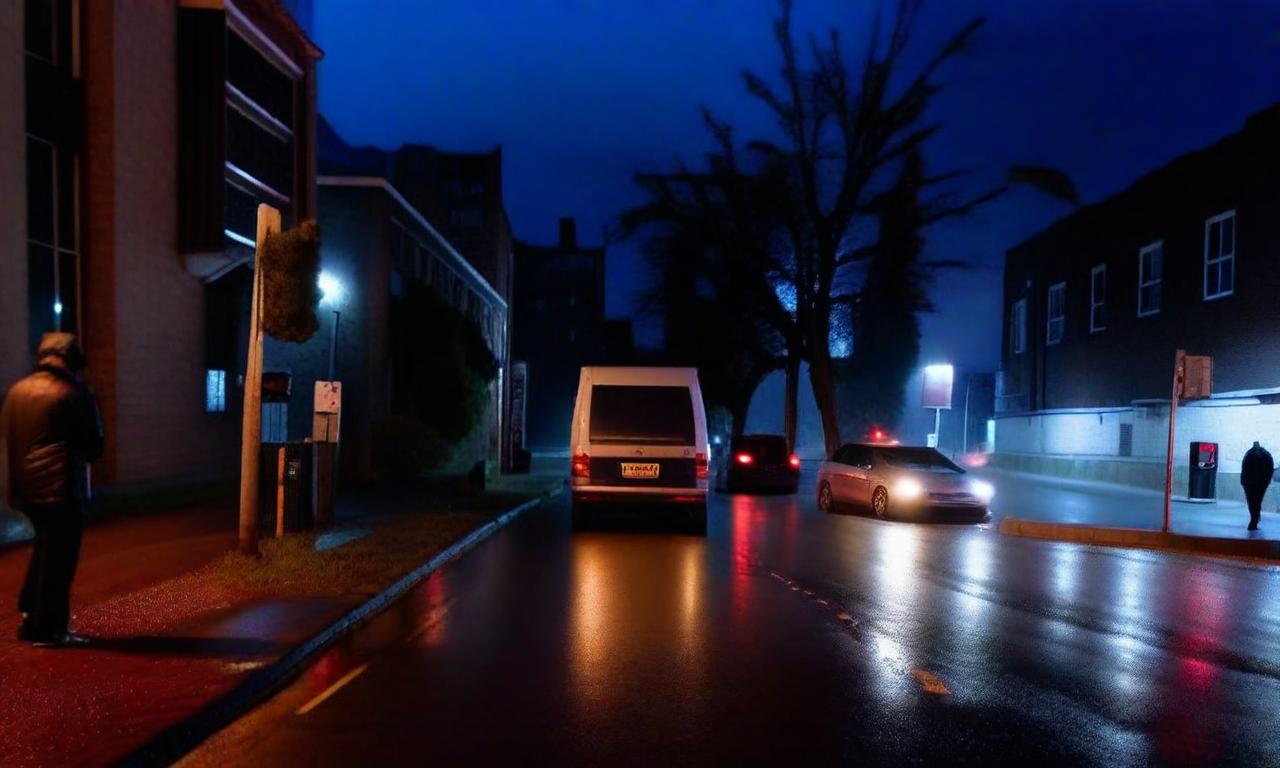} &
	   \includegraphics[width=\turnheightnew]{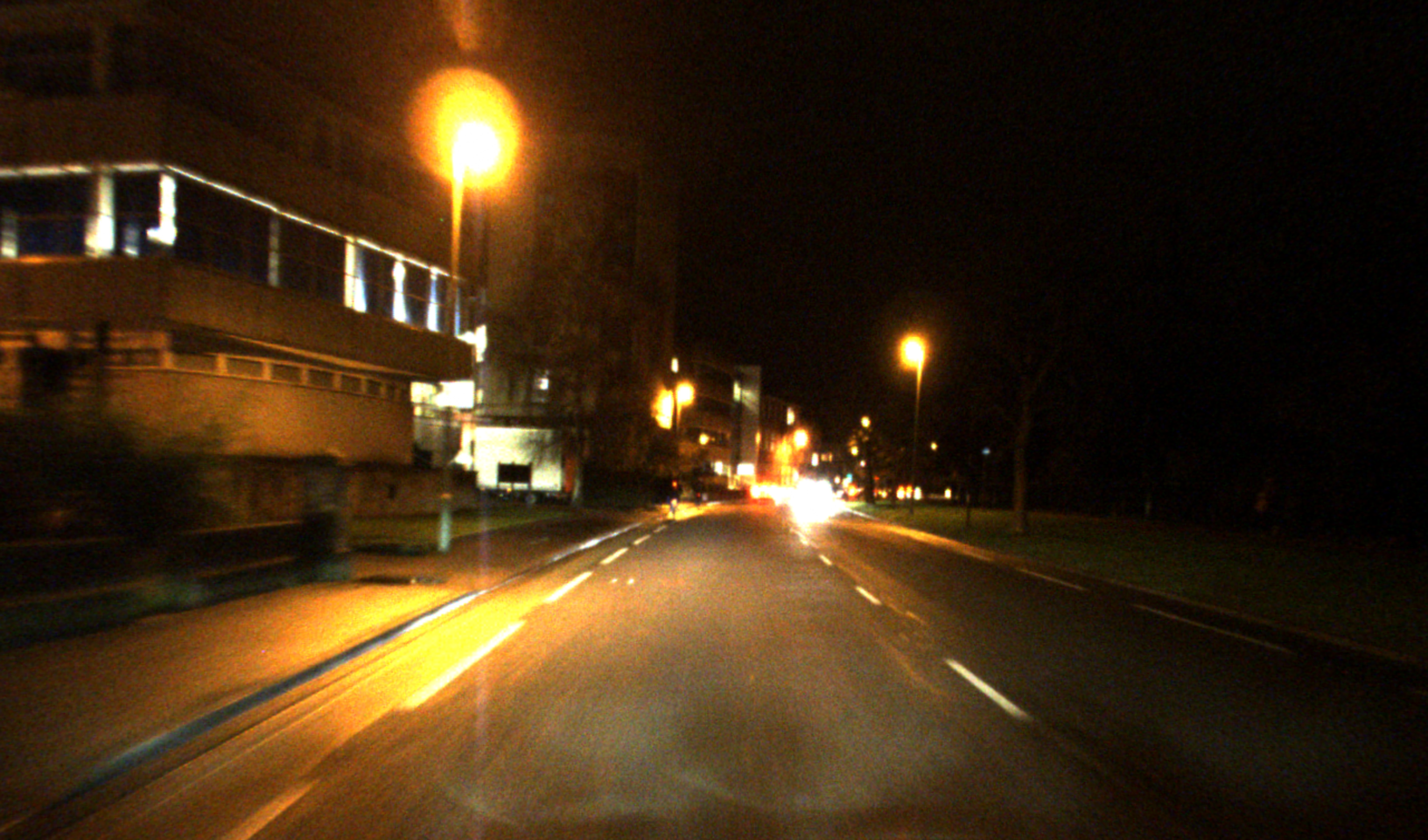} \\
		
		\multicolumn{1}{c}{} & 
		\multicolumn{1}{c}{day} & 
		\multicolumn{1}{c}{ForkGan} & 
		\multicolumn{1}{c}{CycleGAN-Turbo} & 
		\multicolumn{1}{c}{T2I-Adapter} & 
		\multicolumn{1}{c}{RobotCar} \\
	\end{tabular}
	\caption{Data generation results for RobotCar} 
	\label{fig:7}
\end{figure*}

Fig.~\ref{fig:7} presents the results of three different translators for RobotCar. The first and last columns correspond to the daytime and nighttime samples, respectively. Unlike the nuScenes dataset, which has a more uniform lighting distribution, nighttime samples in RobotCar exhibit significant lighting variability, requiring more data to capture this distribution during CycleGAN-Turbo training. The translation results from ForkGAN and CycleGAN-Turbo are shown in the second and third columns of Fig.~\ref{fig:7}, respectively. These methods produce more realistic translations compared to T2I-Adapter. Additionally, CycleGAN-Turbo requires significantly fewer training samples than ForkGAN, enabling better generalization to scenarios with limited real samples.

\section{Additional Qualitative Results on nuScenes and RobotCar}

\begin{figure*}[htbp]
	\centering
	\newcommand{\turnheightnew}{0.35\columnwidth}
	\begin{tabular}{@{\hskip 1mm}c@{\hskip 1mm}c@{\hskip 1mm}c@{\hskip 1mm}c@{\hskip 1mm}c@{\hskip 1mm}c@{}}
		\vspace{-0.5mm}
		{\rotatebox{90}{\hspace{2mm}}} &
		\includegraphics[width=\turnheightnew,keepaspectratio]{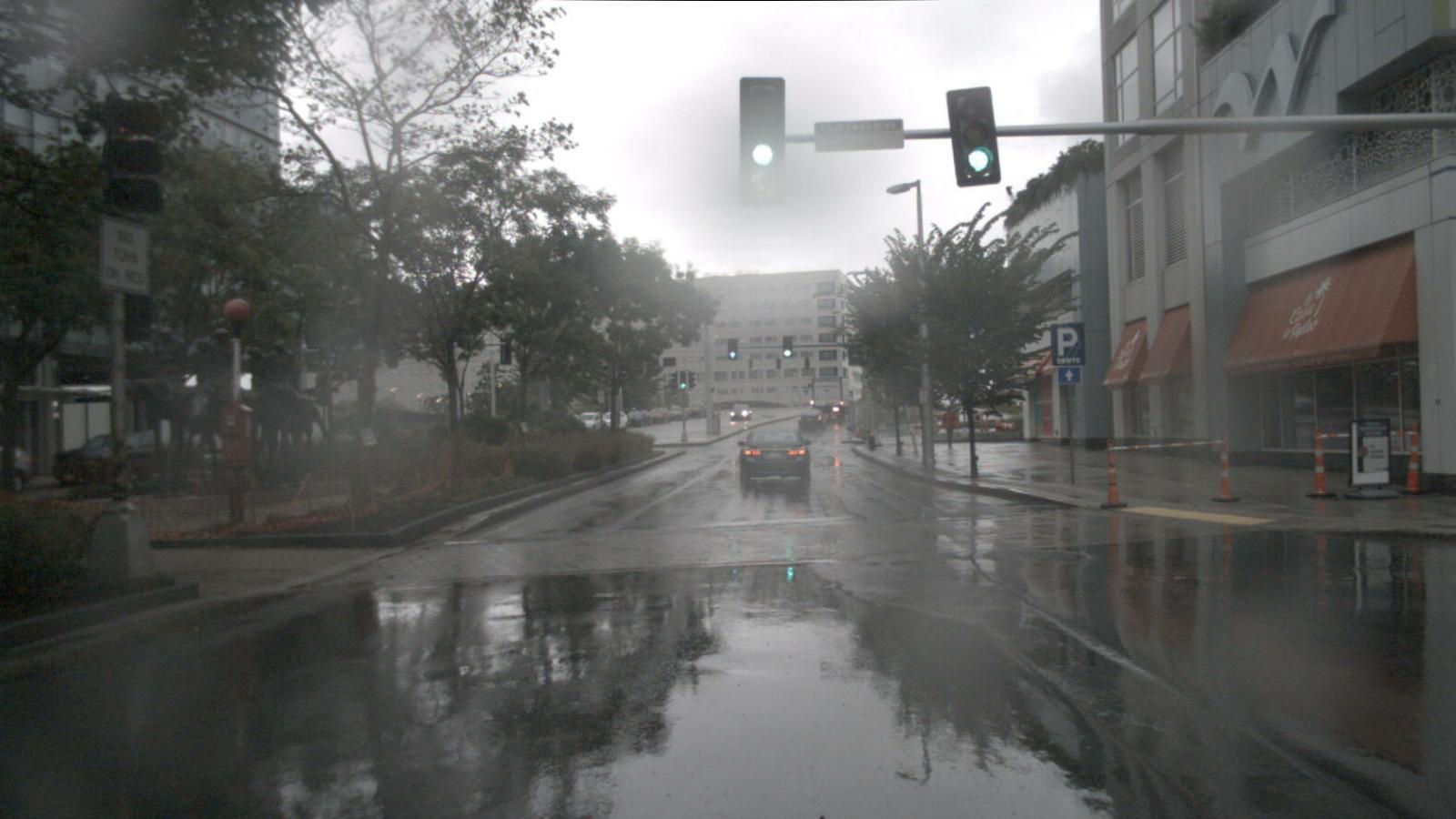} &
		\includegraphics[width=\turnheightnew,keepaspectratio]{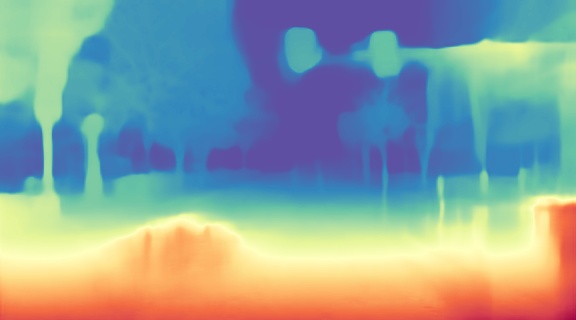} &
		\includegraphics[width=\turnheightnew,keepaspectratio]{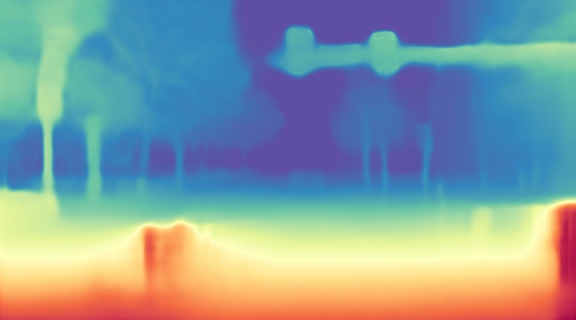} &
		\includegraphics[width=\turnheightnew,keepaspectratio]{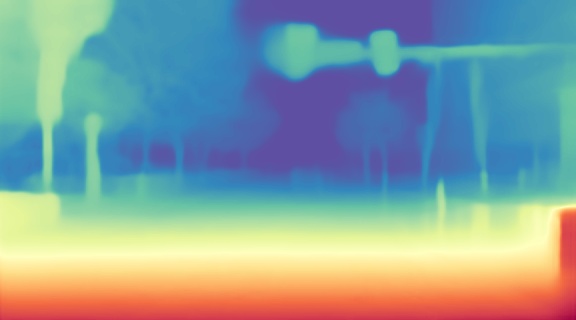} &
		\includegraphics[width=\turnheightnew,keepaspectratio]{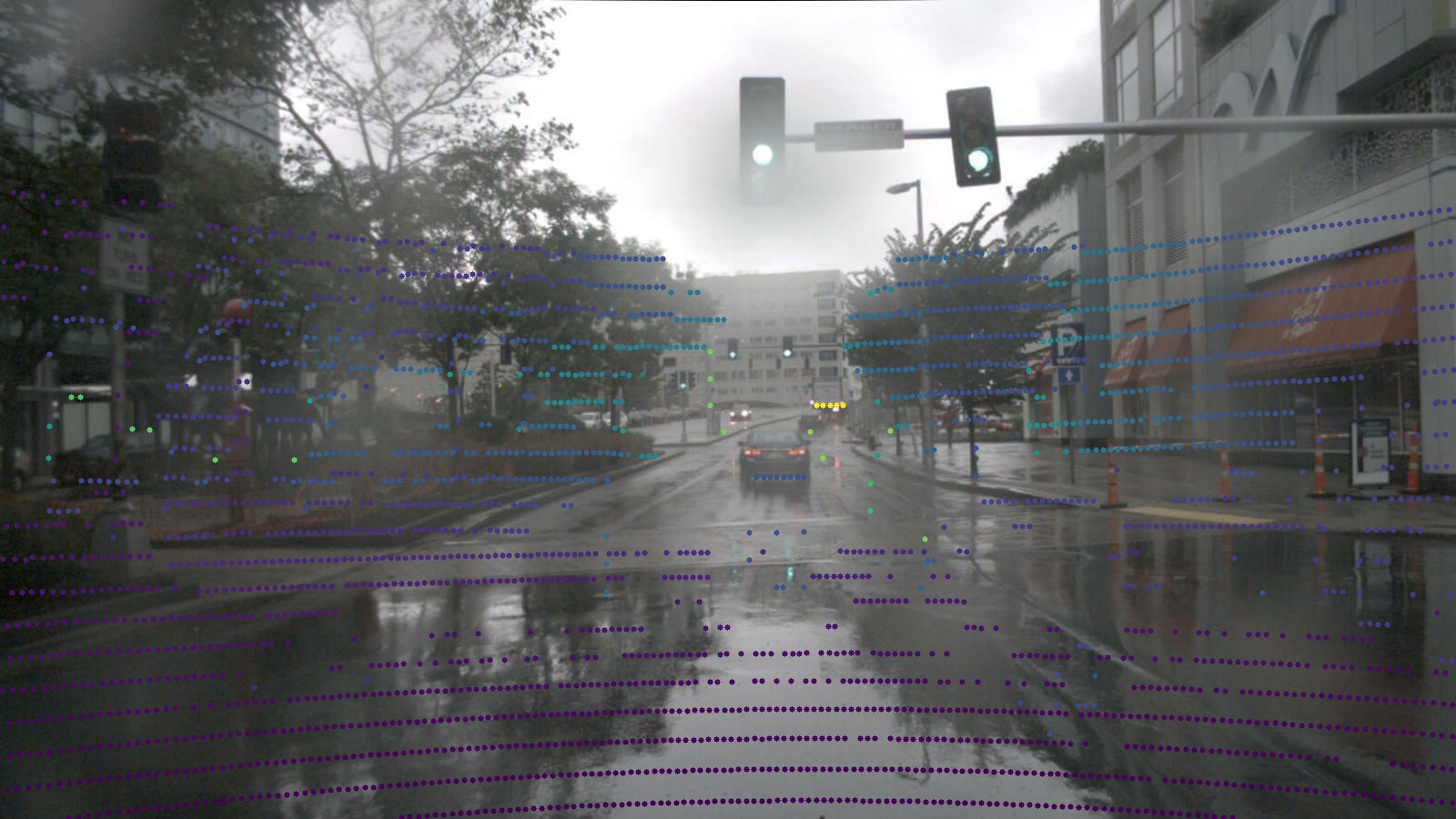} \\
		
		\vspace{-0.5mm}
		{\rotatebox{90}{\hspace{2mm}}} &
		\includegraphics[width=\turnheightnew,keepaspectratio]{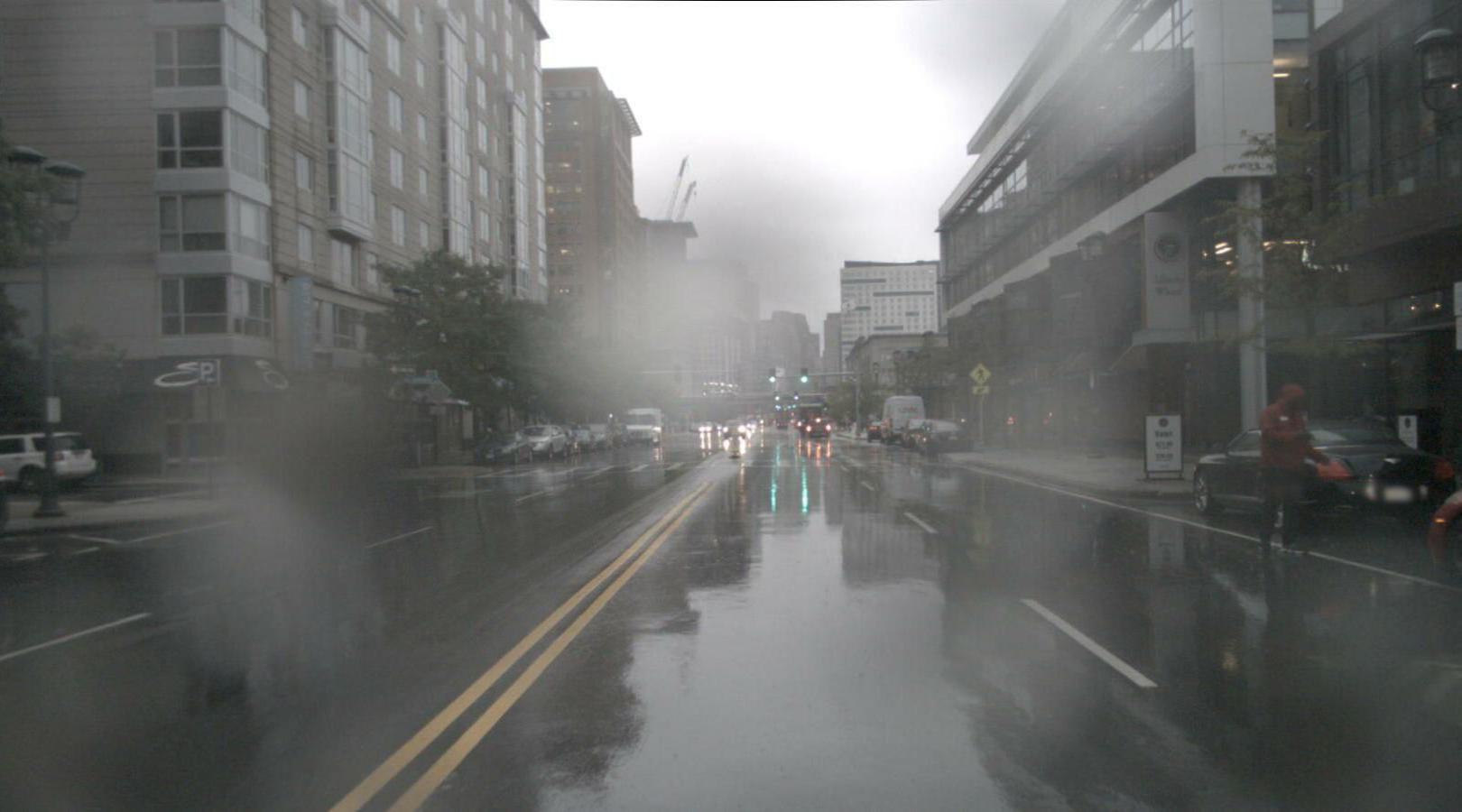} &
		\includegraphics[width=\turnheightnew,keepaspectratio]{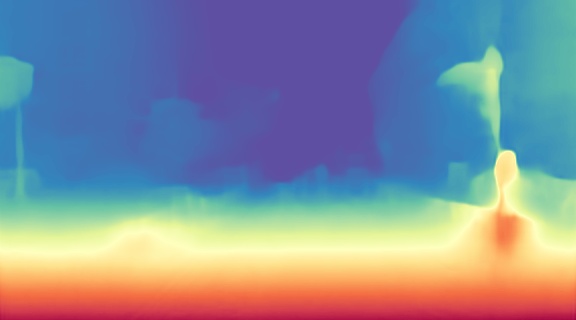} &
		\includegraphics[width=\turnheightnew,keepaspectratio]{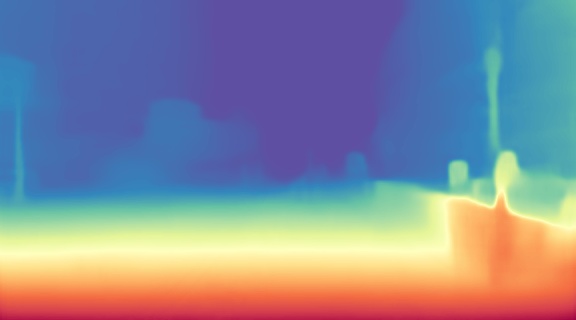} &
		\includegraphics[width=\turnheightnew,keepaspectratio]{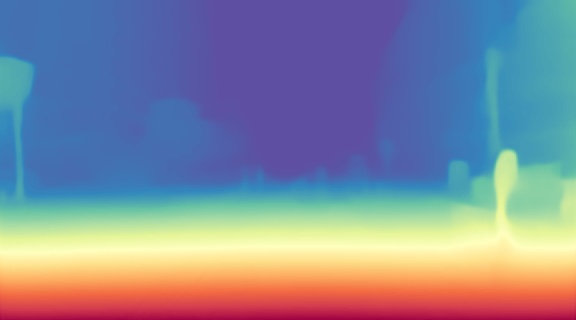} &
		\includegraphics[width=\turnheightnew,keepaspectratio]{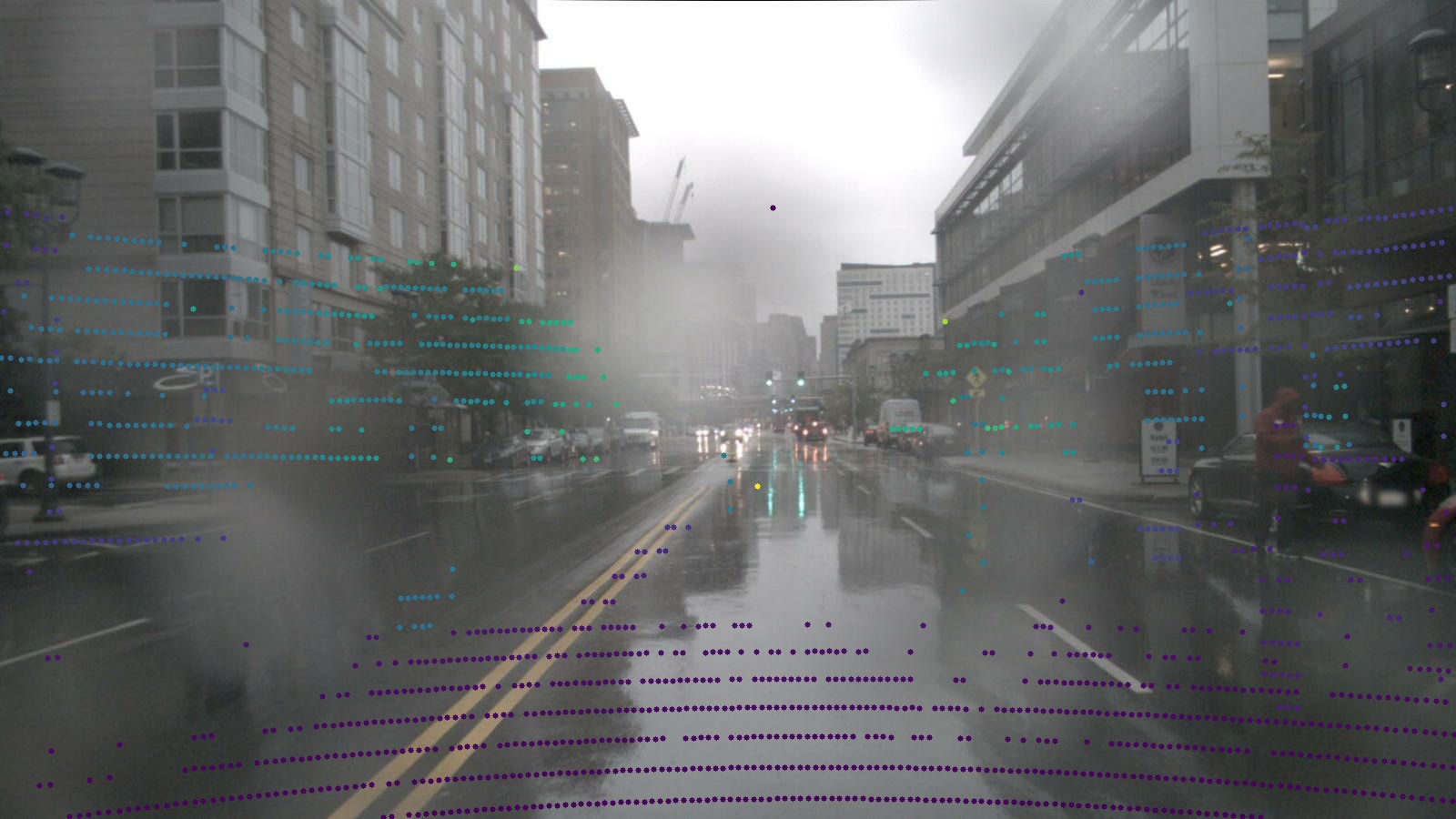} \\
		
		\vspace{-0.5mm}
		{\rotatebox{90}{\hspace{2mm}}} &
		\includegraphics[width=\turnheightnew,keepaspectratio]{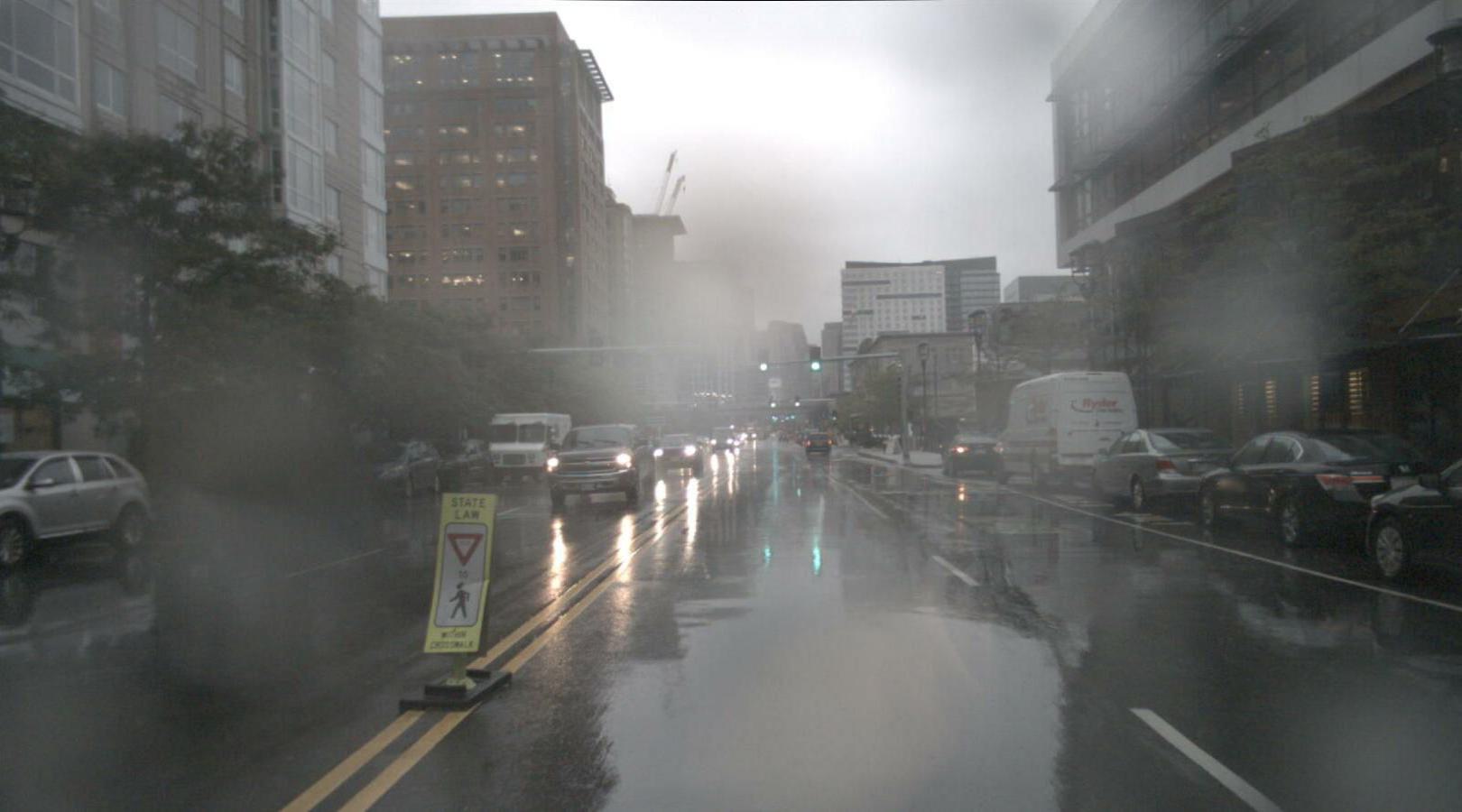} &
		\includegraphics[width=\turnheightnew,keepaspectratio]{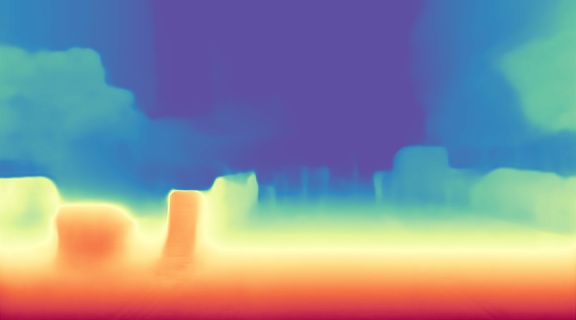} &
		\includegraphics[width=\turnheightnew,keepaspectratio]{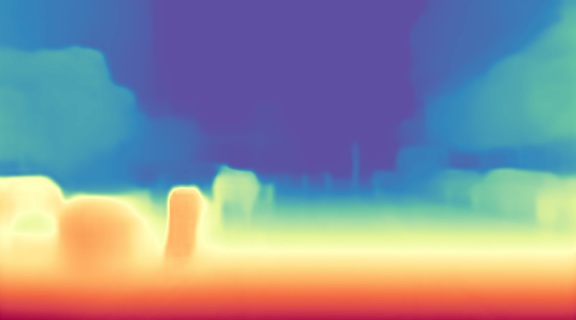} &
		\includegraphics[width=\turnheightnew,keepaspectratio]{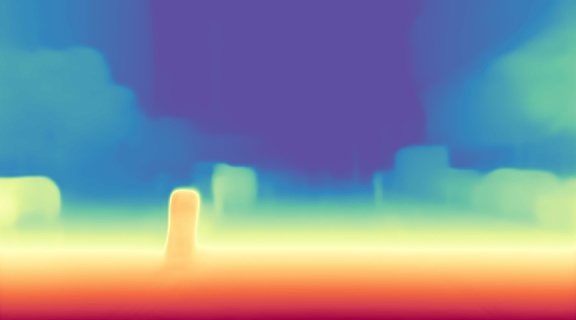} &
		\includegraphics[width=\turnheightnew,keepaspectratio]{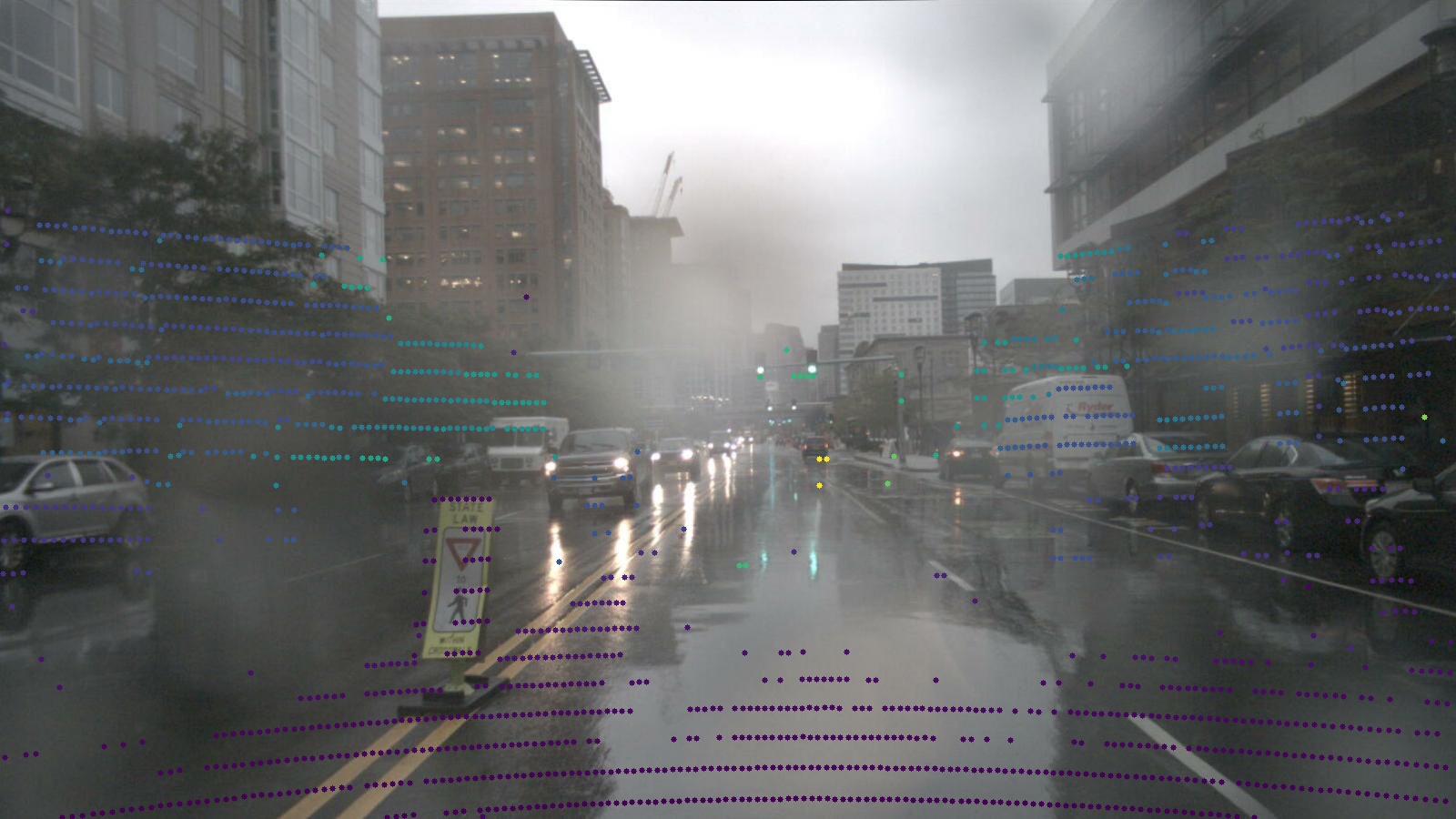} \\
		
		\vspace{-0.5mm}
		{\rotatebox{90}{\hspace{2mm}}} &
		\includegraphics[width=\turnheightnew,keepaspectratio]{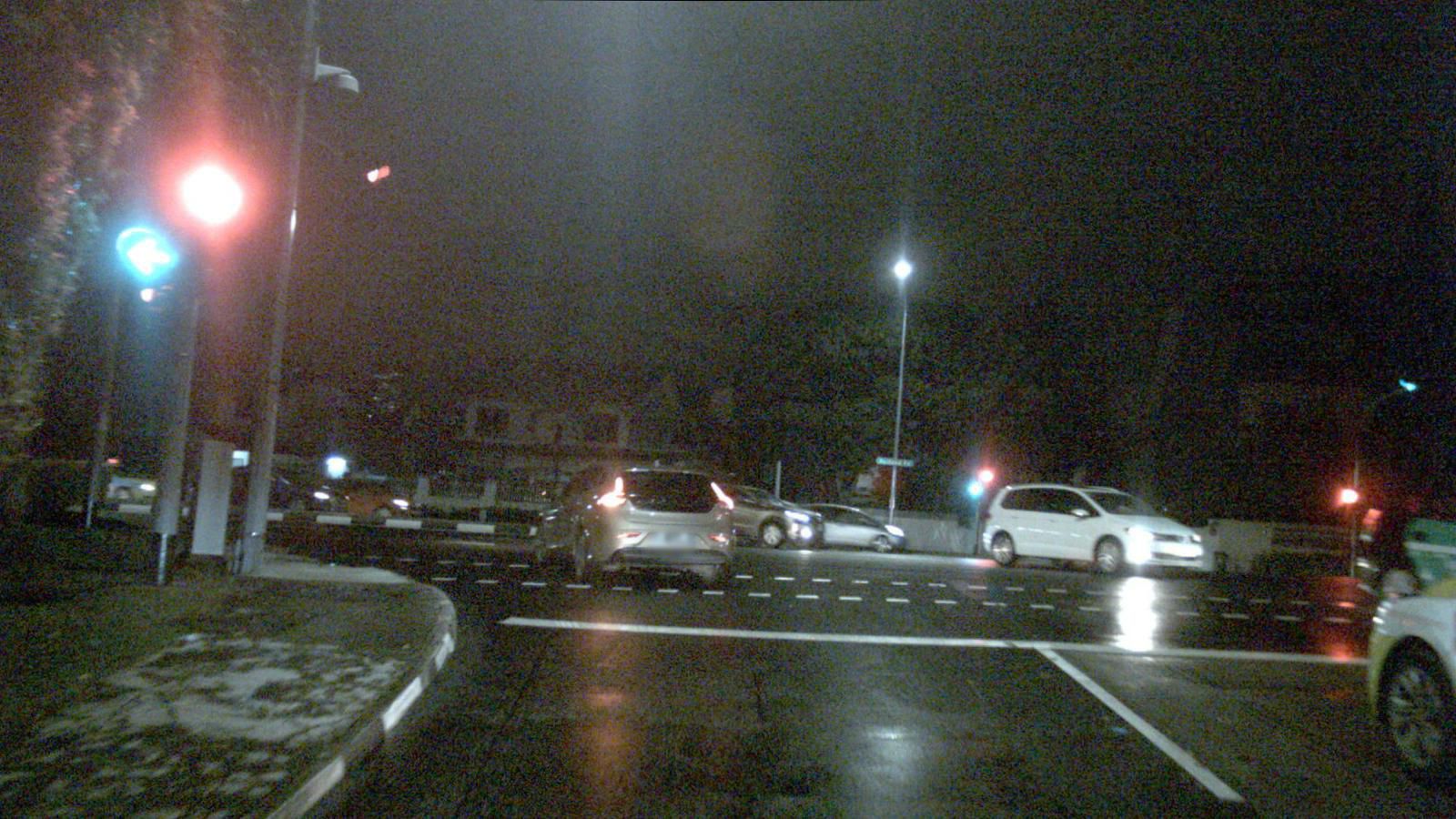} &
		\includegraphics[width=\turnheightnew,keepaspectratio]{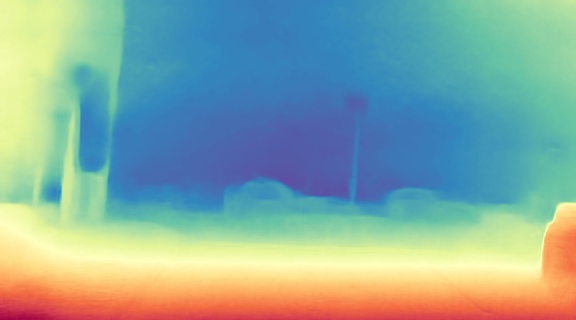} &
		\includegraphics[width=\turnheightnew,keepaspectratio]{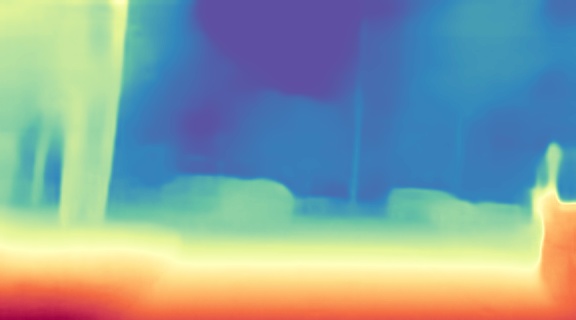} &
		\includegraphics[width=\turnheightnew,keepaspectratio]{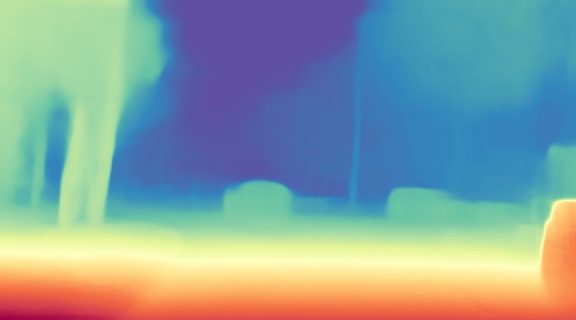} &
		\includegraphics[width=\turnheightnew,keepaspectratio]{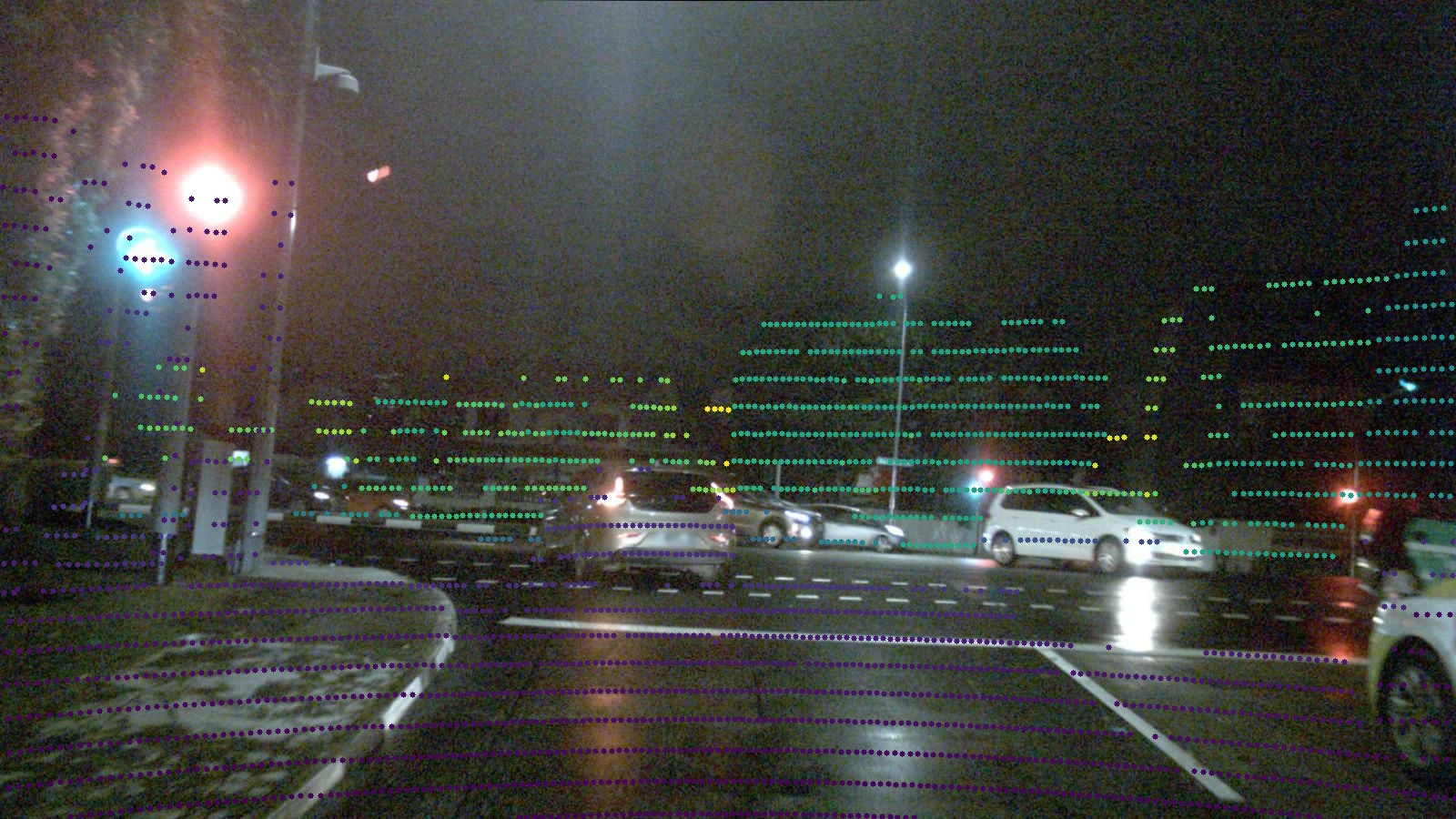} \\
		
		\vspace{-0.5mm}
		{\rotatebox{90}{\hspace{2mm}}} &
		\includegraphics[width=\turnheightnew,keepaspectratio]{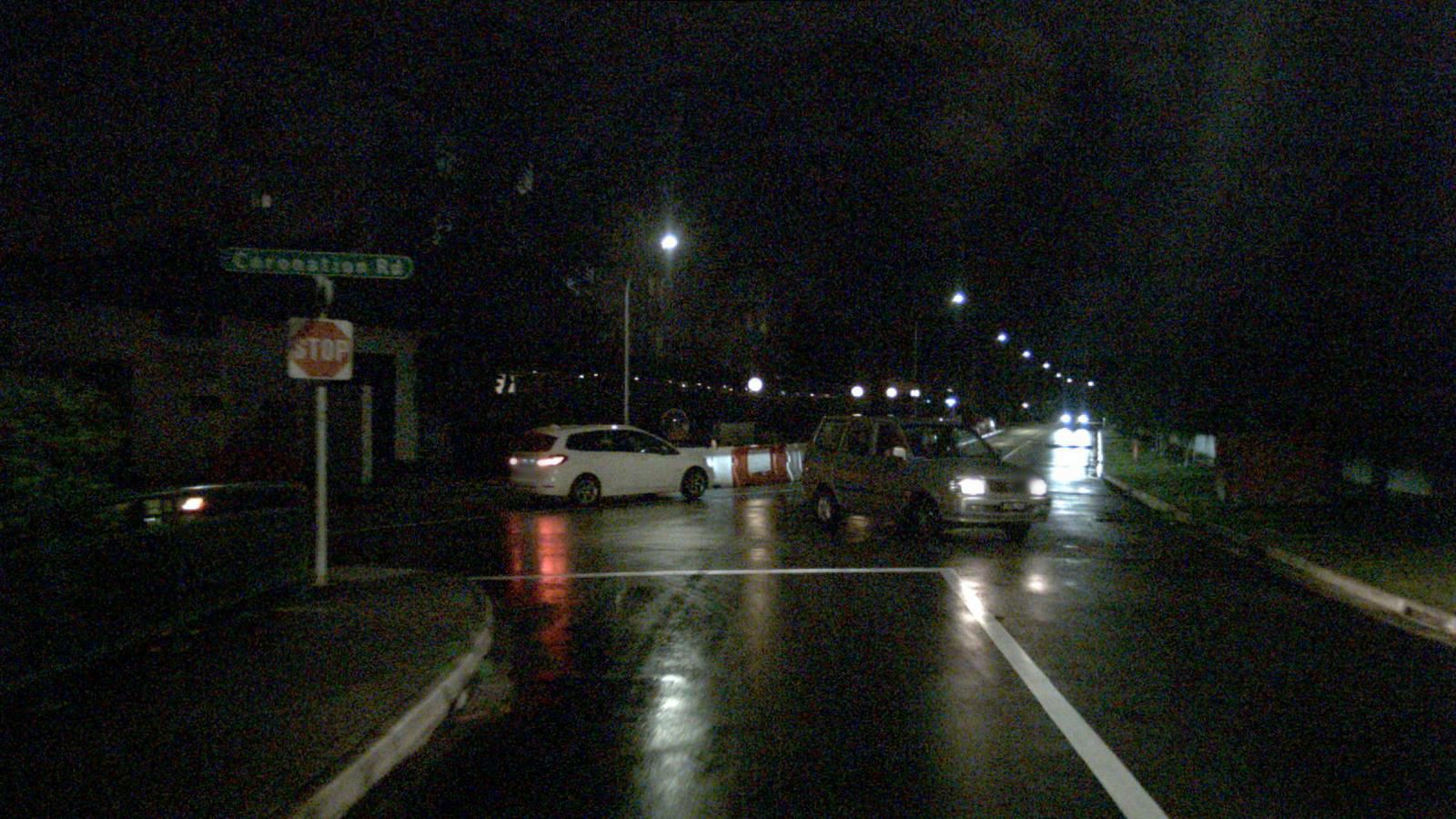} &
		\includegraphics[width=\turnheightnew,keepaspectratio]{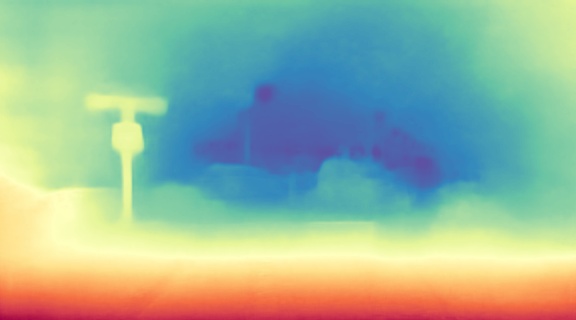} &
		\includegraphics[width=\turnheightnew,keepaspectratio]{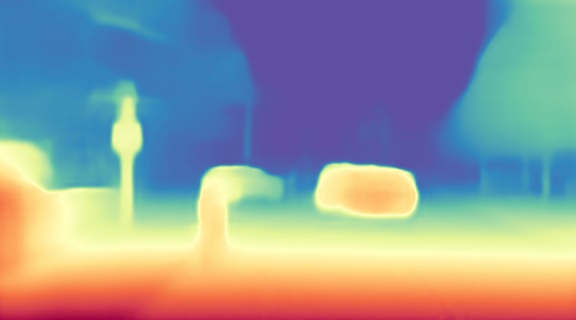} &
		\includegraphics[width=\turnheightnew,keepaspectratio]{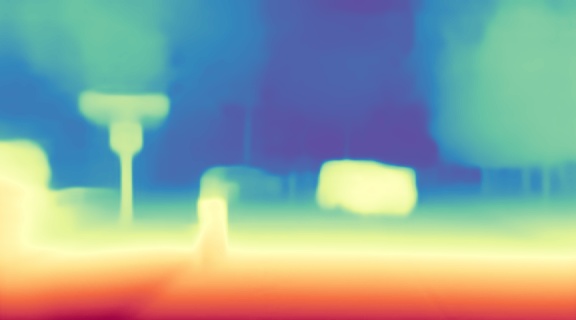} &
		\includegraphics[width=\turnheightnew,keepaspectratio]{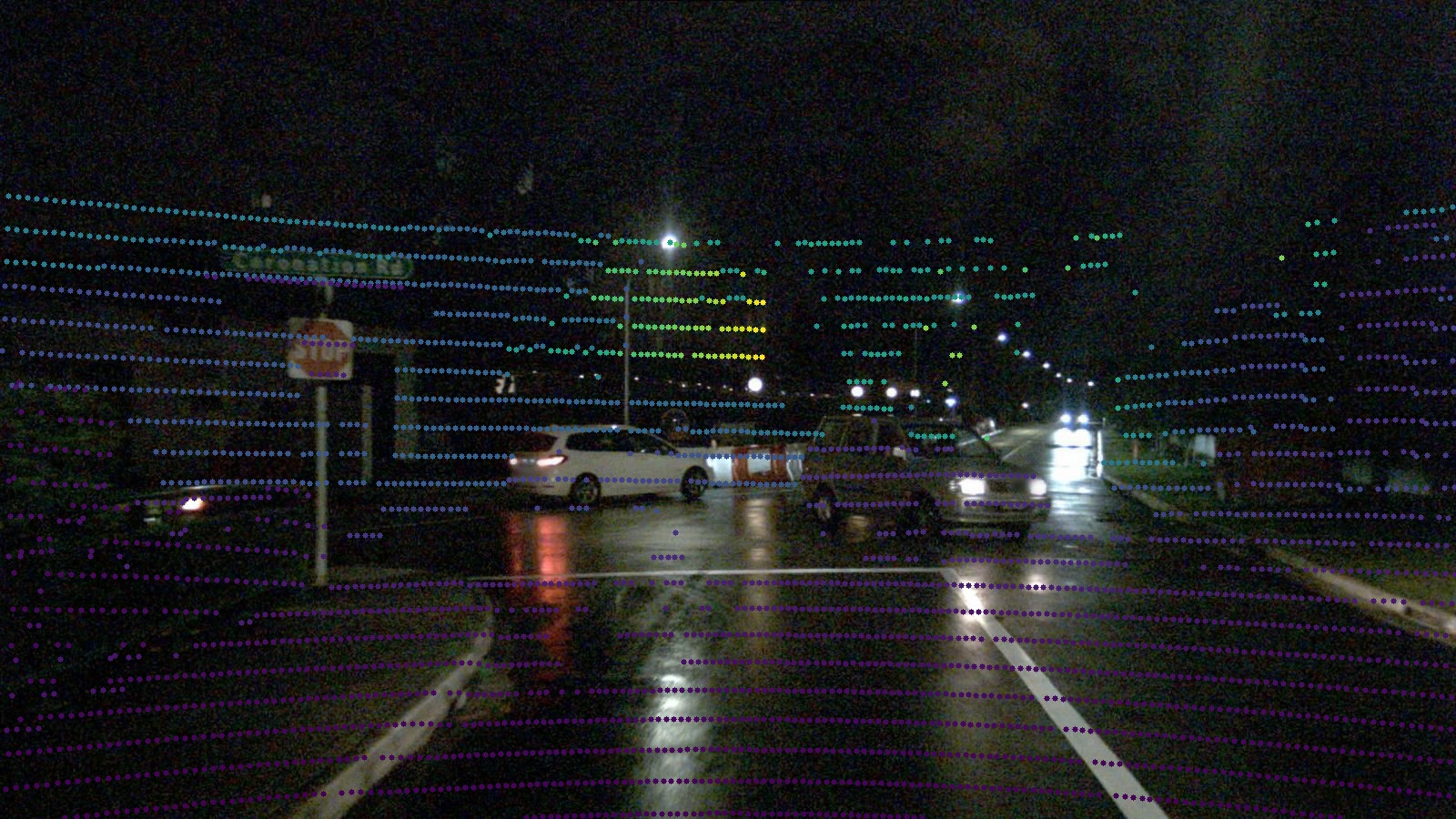} \\
		
		\vspace{-0.5mm}
		{\rotatebox{90}{\hspace{2mm}}} &
		\includegraphics[width=\turnheightnew,keepaspectratio]{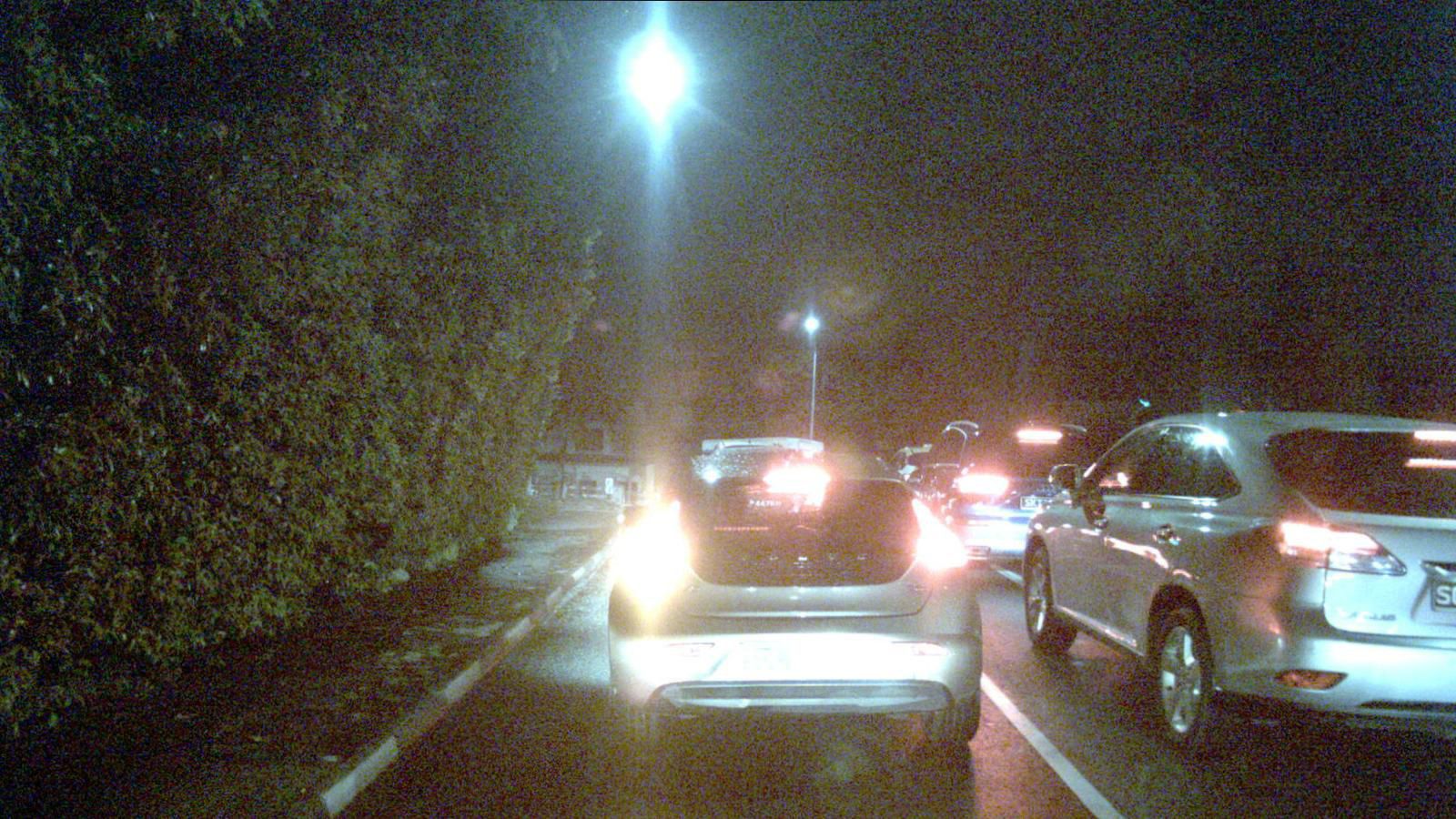} &
		\includegraphics[width=\turnheightnew,keepaspectratio]{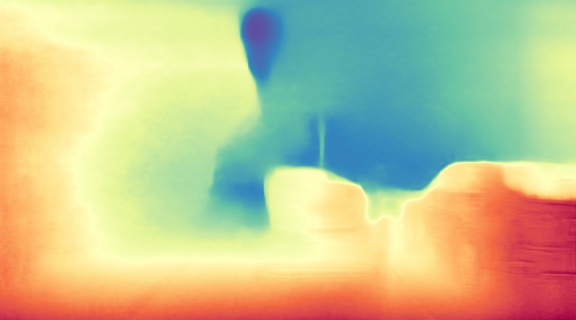} &
		\includegraphics[width=\turnheightnew,keepaspectratio]{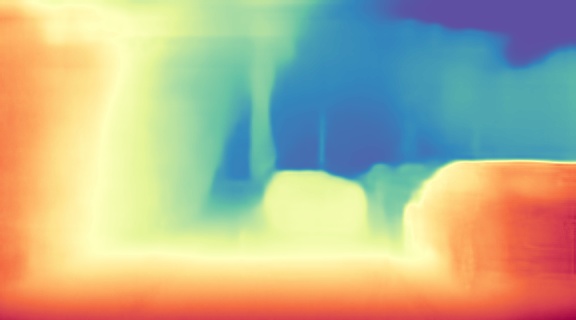} &
		\includegraphics[width=\turnheightnew,keepaspectratio]{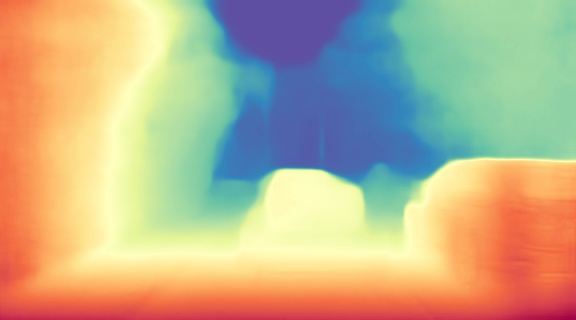} &
		\includegraphics[width=\turnheightnew,keepaspectratio]{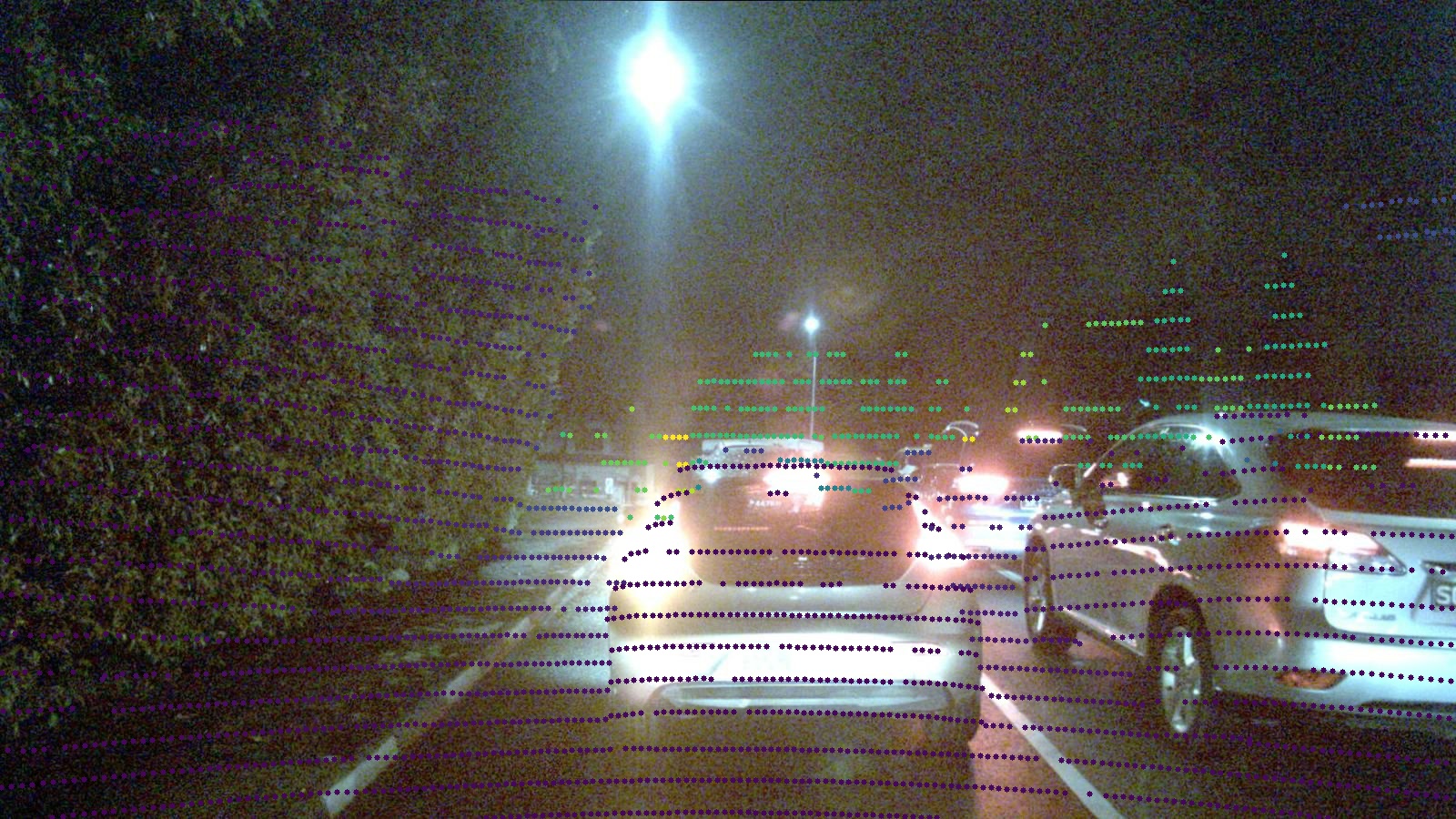} \\
		
		\vspace{-0.5mm}
		{\rotatebox{90}{\hspace{2mm}}} &
		\includegraphics[width=\turnheightnew,keepaspectratio]{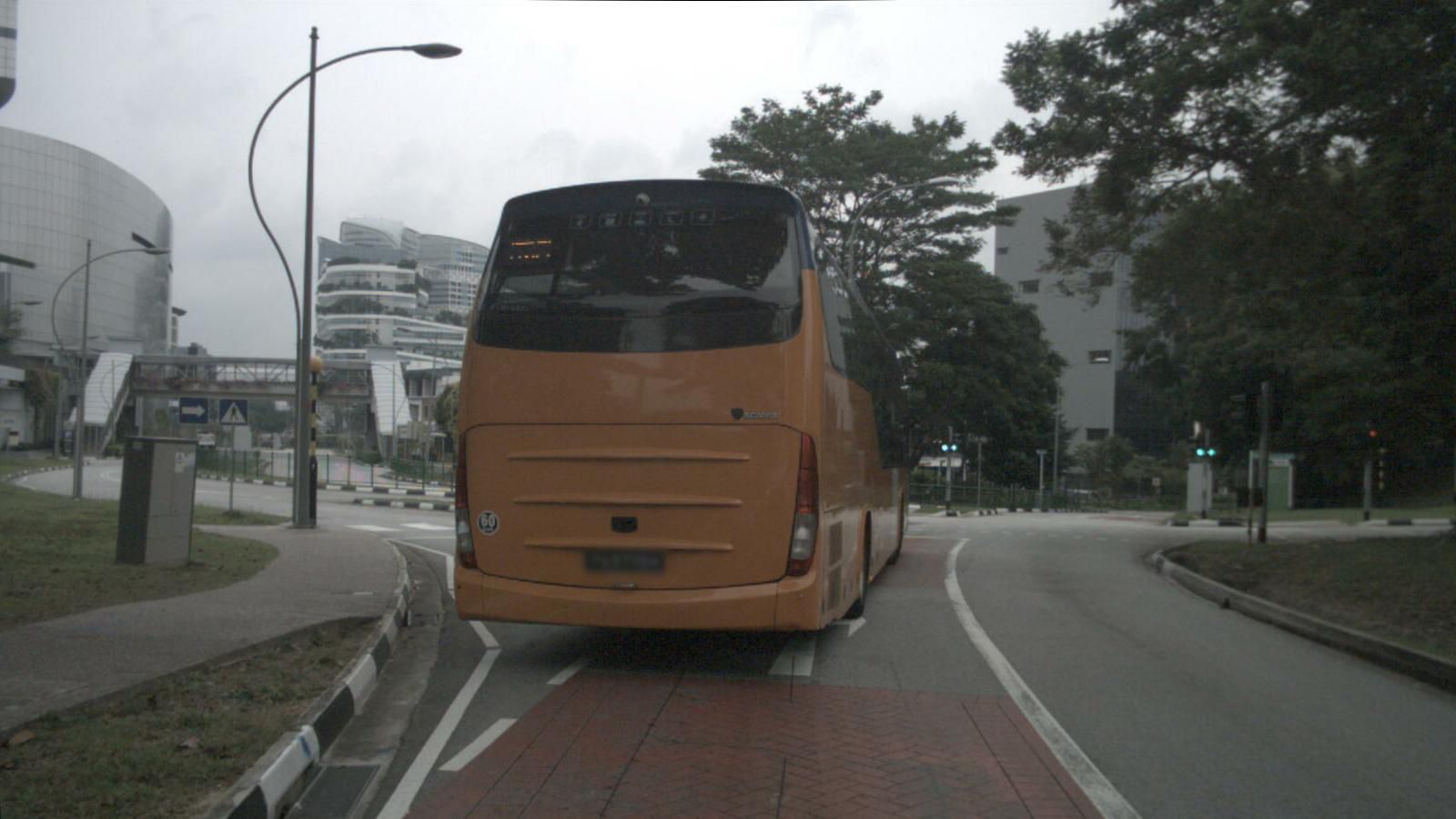} &
		\includegraphics[width=\turnheightnew,keepaspectratio]{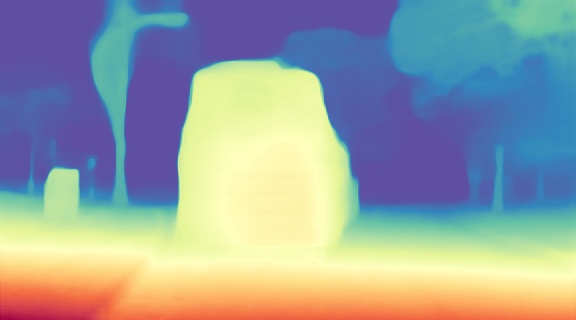} &
		\includegraphics[width=\turnheightnew,keepaspectratio]{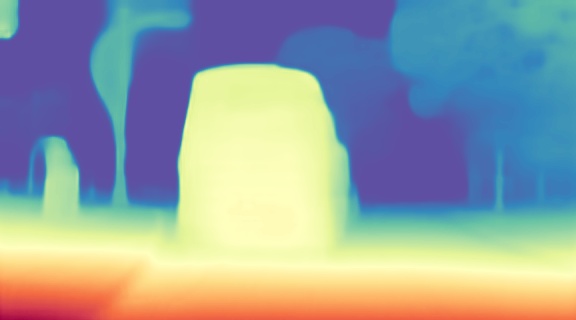} &
		\includegraphics[width=\turnheightnew,keepaspectratio]{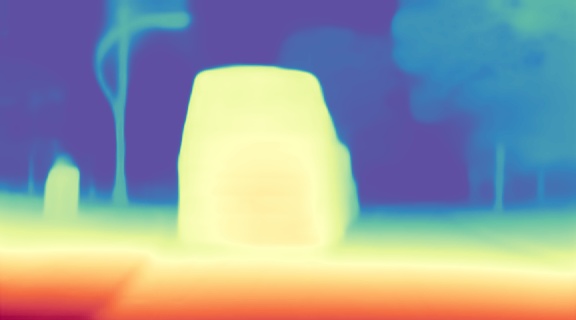} &
		\includegraphics[width=\turnheightnew,keepaspectratio]{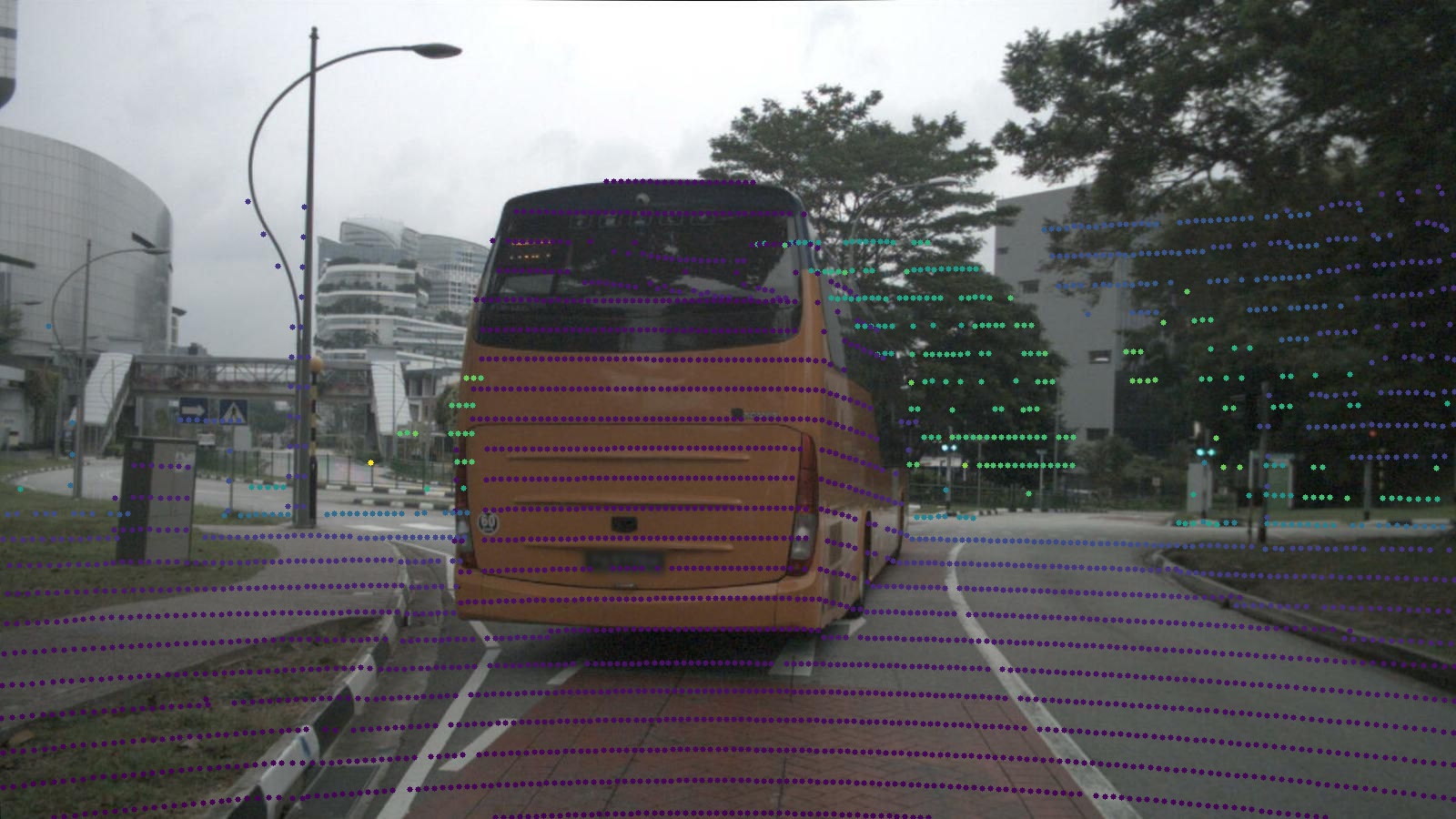} \\
		
		\vspace{-0.5mm}
		{\rotatebox{90}{\hspace{2mm}}} &
		\includegraphics[width=\turnheightnew,keepaspectratio]{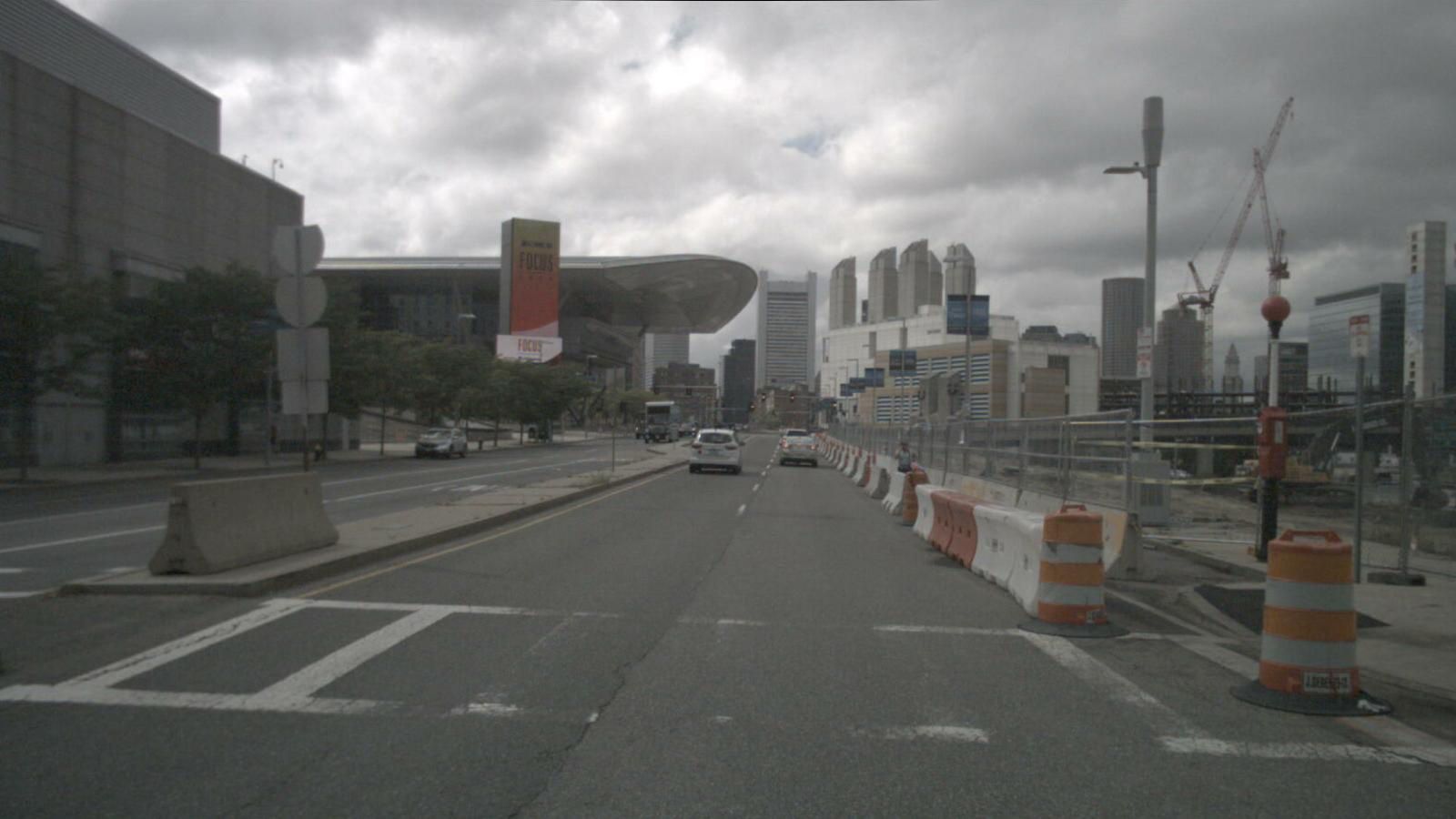} &
		\includegraphics[width=\turnheightnew,keepaspectratio]{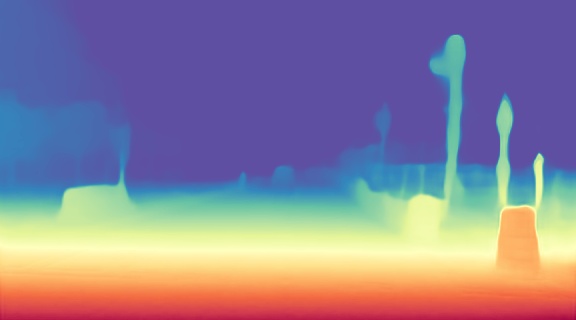} &
		\includegraphics[width=\turnheightnew,keepaspectratio]{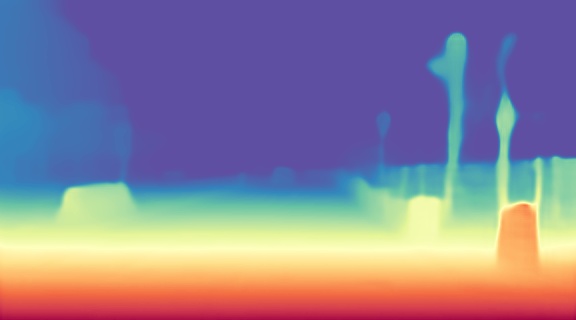} &
		\includegraphics[width=\turnheightnew,keepaspectratio]{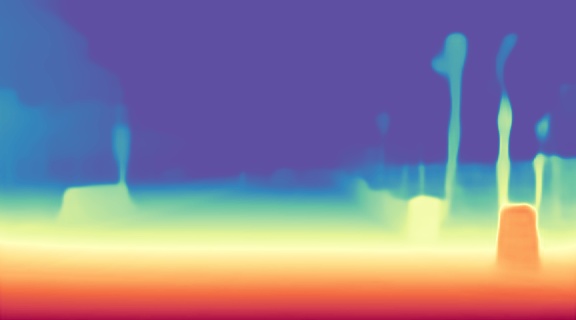} &
		\includegraphics[width=\turnheightnew,keepaspectratio]{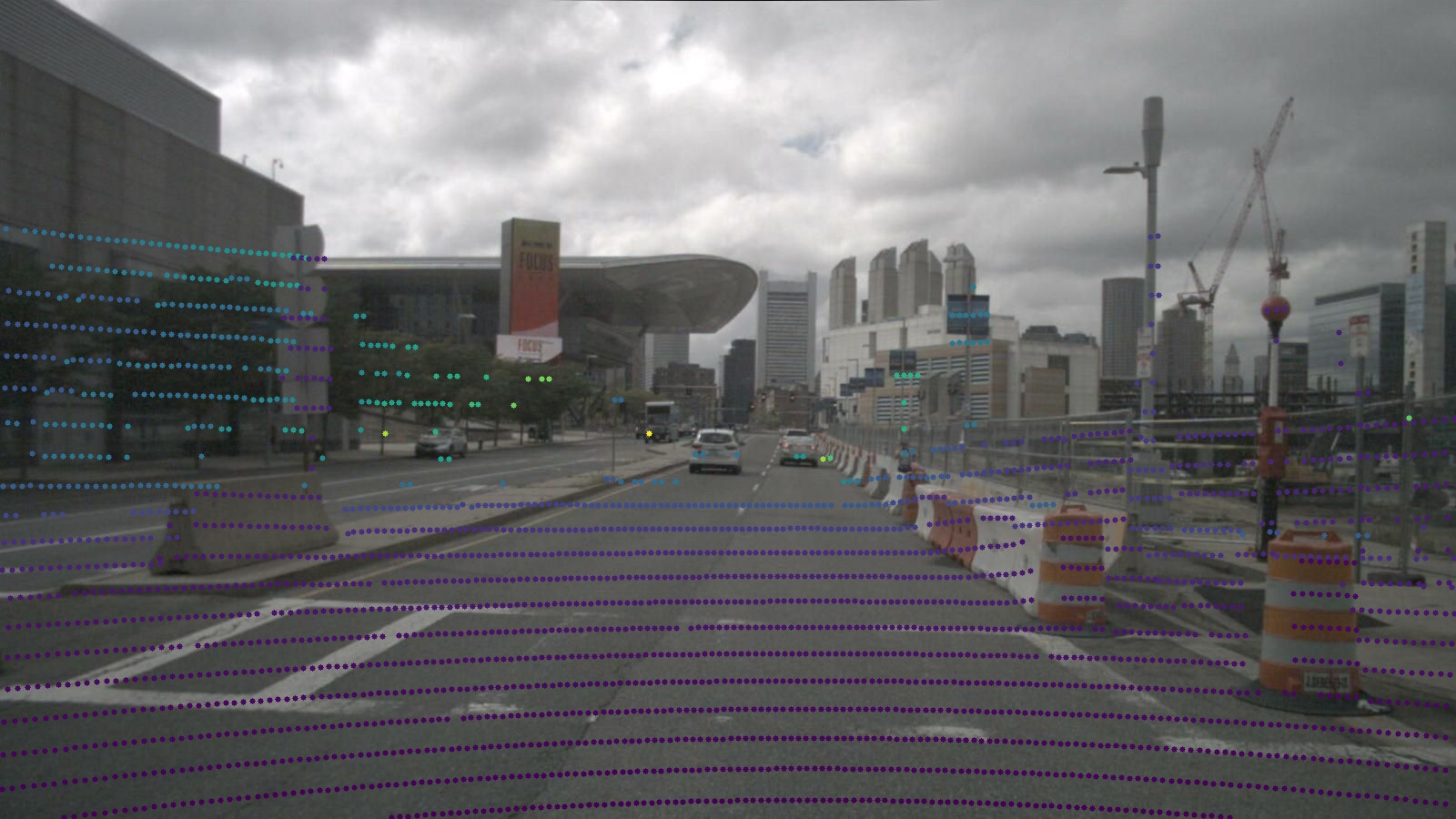} \\
		
		\vspace{-0.5mm}
		{\rotatebox{90}{\hspace{2mm}}} &
		\includegraphics[width=\turnheightnew,keepaspectratio]{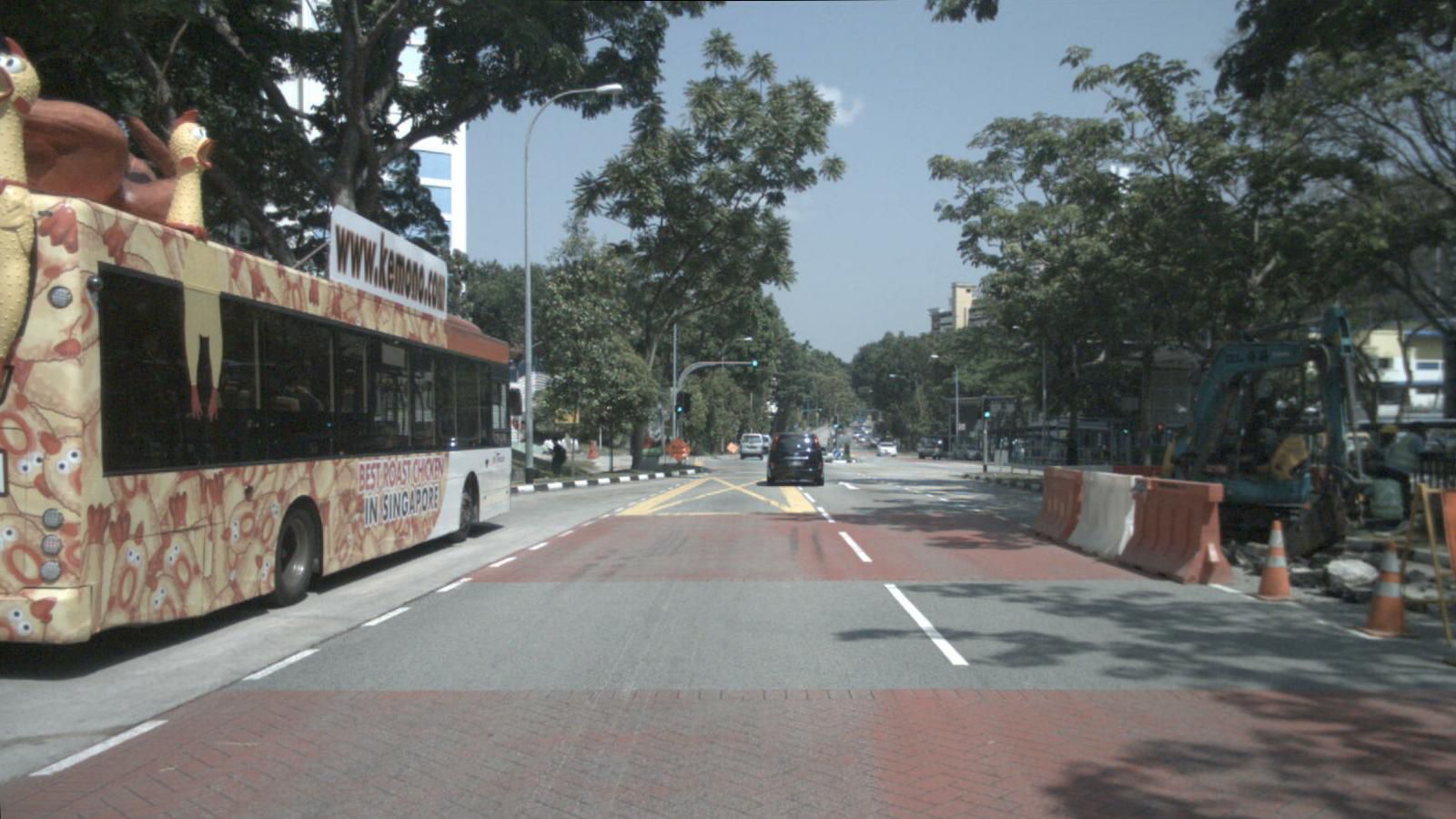} &
		\includegraphics[width=\turnheightnew,keepaspectratio]{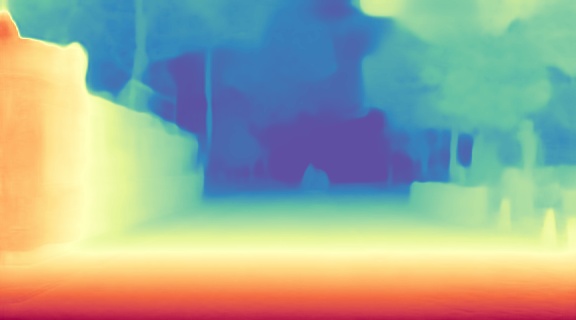} &
		\includegraphics[width=\turnheightnew,keepaspectratio]{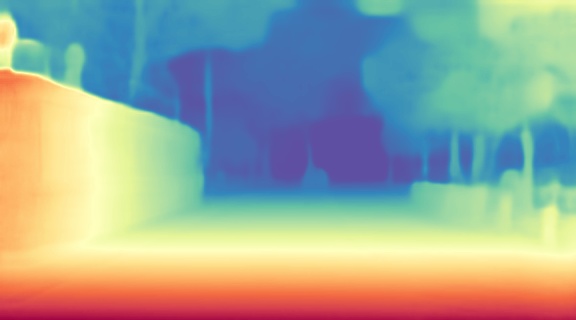} &
		\includegraphics[width=\turnheightnew,keepaspectratio]{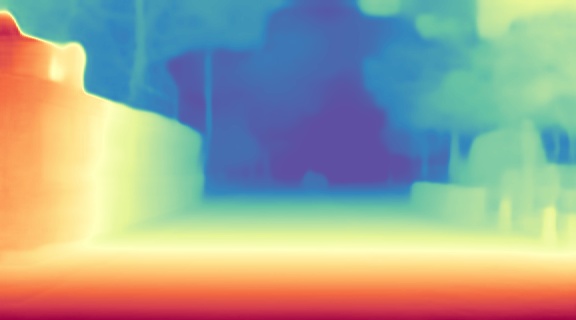} &
		\includegraphics[width=\turnheightnew,keepaspectratio]{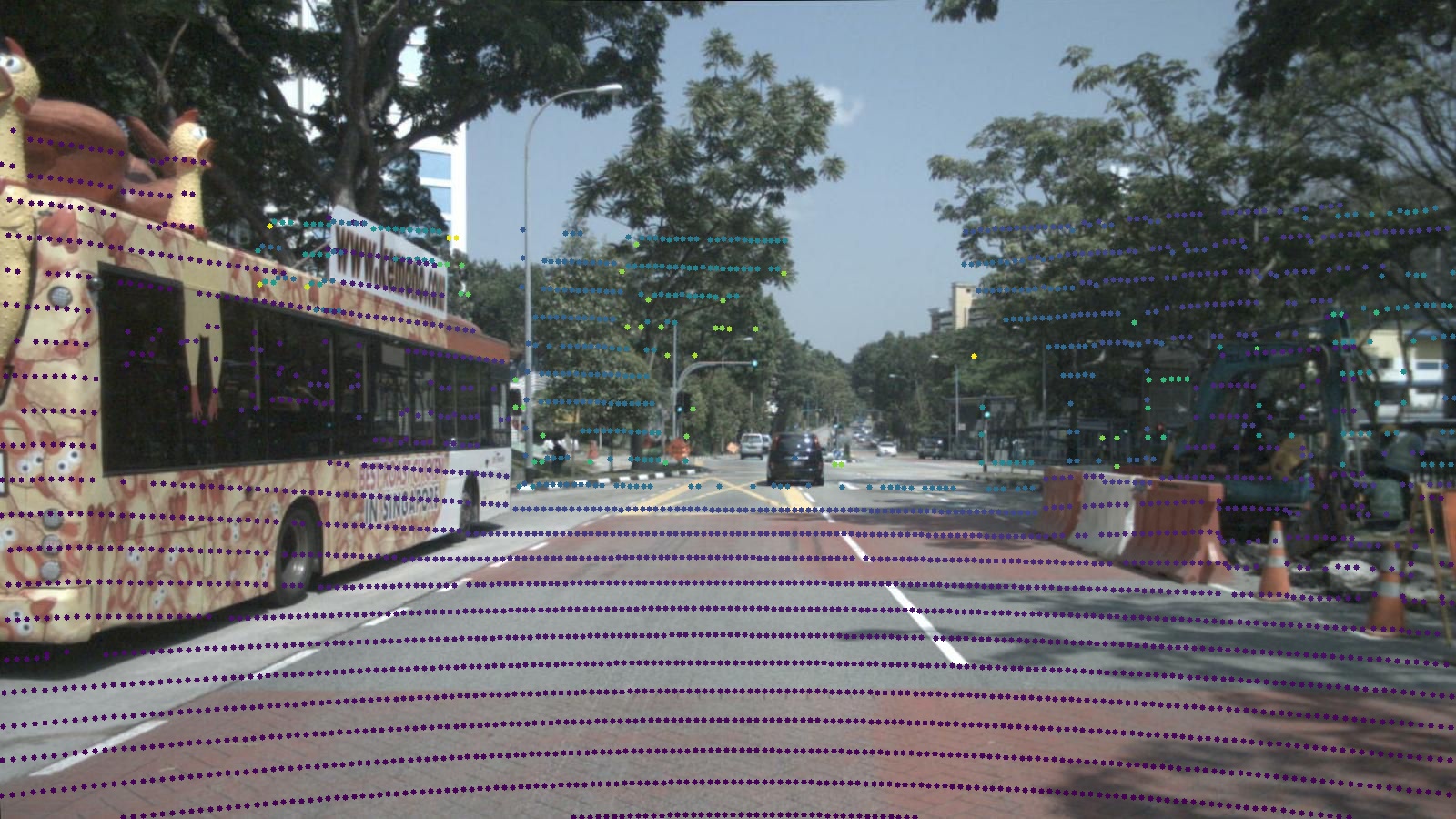} \\

		\multicolumn{1}{c}{} & 
		\multicolumn{1}{c}{Image} & 
		\multicolumn{1}{c}{Monodepth2} & 
		\multicolumn{1}{c}{md4all-DD} & 
		\multicolumn{1}{c}{ACDepth} & 
		\multicolumn{1}{c}{GT} \\
	\end{tabular}
	\caption{Qualitative results on nuScenes~\protect\cite{caesar2020nuscenes}} 
	\label{fig:8}
\end{figure*}

\begin{figure*}[htbp]
	\centering
	\newcommand{\turnheightnew}{0.35\columnwidth}
	\begin{tabular}{@{\hskip 1mm}c@{\hskip 1mm}c@{\hskip 1mm}c@{\hskip 1mm}c@{\hskip 1mm}c@{\hskip 1mm}c@{}}
    
		\vspace{-0.5mm}
		{\rotatebox{90}{\hspace{2mm}}} &
		\includegraphics[width=\turnheightnew,keepaspectratio]{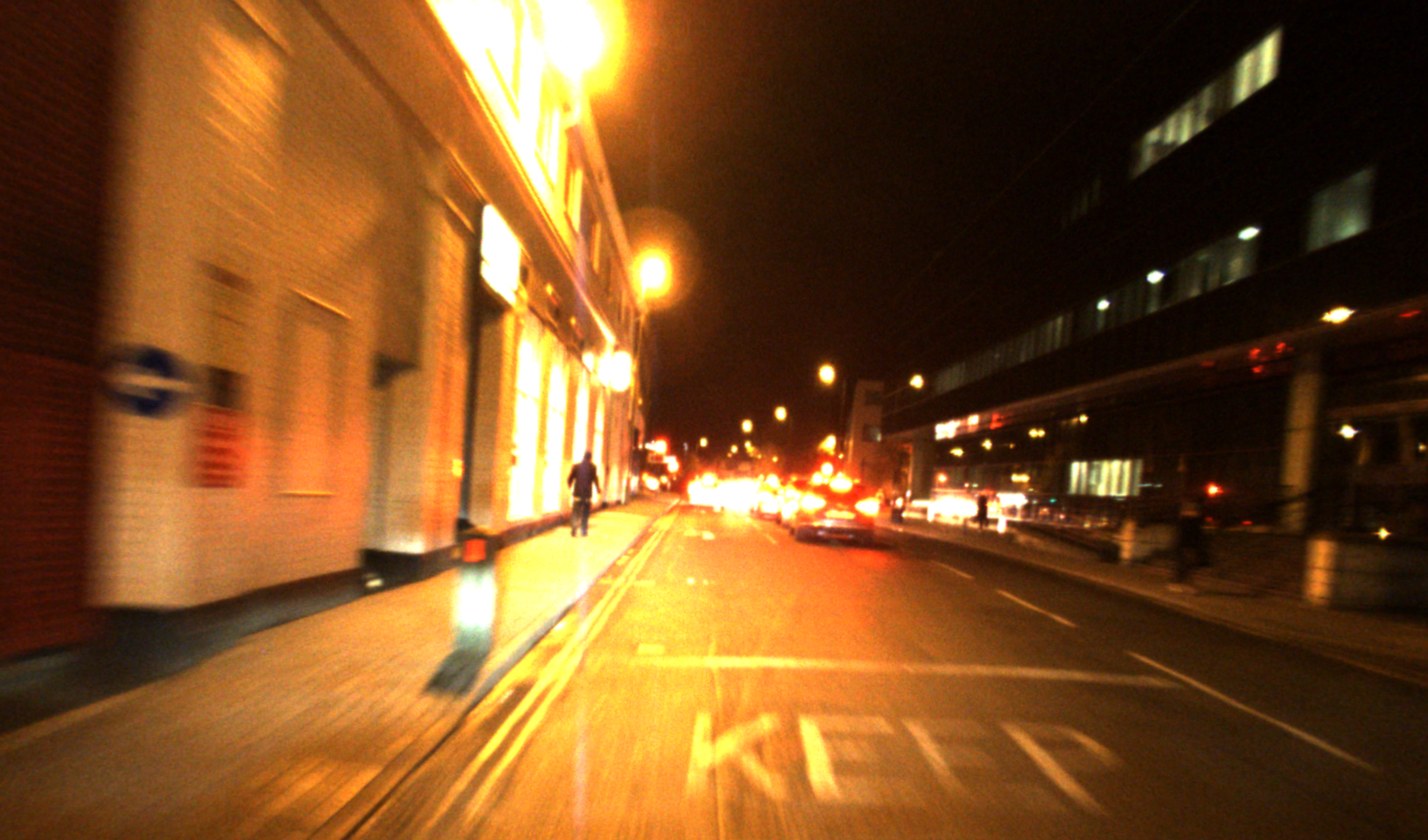} &
		\includegraphics[width=\turnheightnew,keepaspectratio]{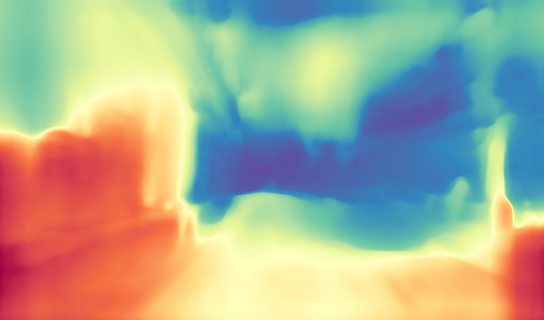} &
		\includegraphics[width=\turnheightnew,keepaspectratio]{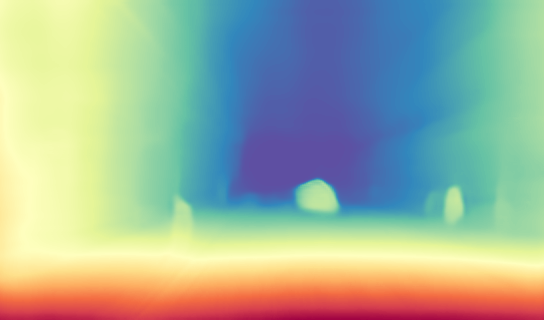} &
		\includegraphics[width=\turnheightnew,keepaspectratio]{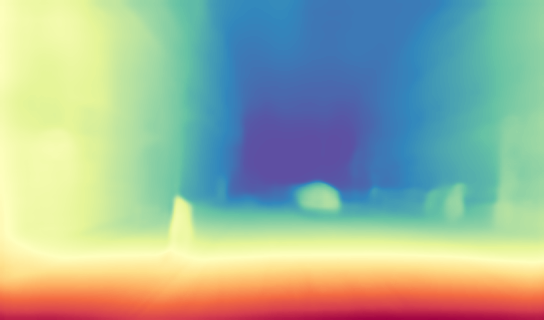} &
		\includegraphics[width=\turnheightnew,keepaspectratio]{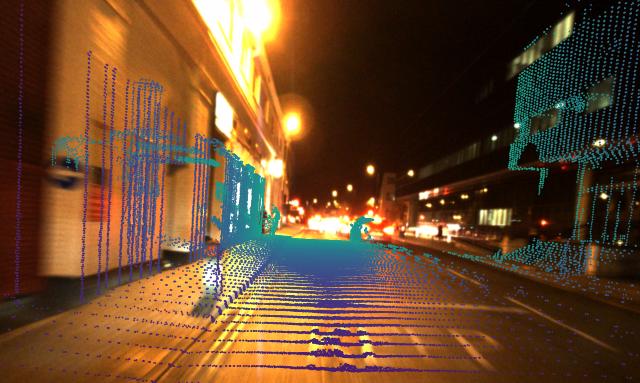} \\
		
		\vspace{-0.5mm}
		{\rotatebox{90}{\hspace{2mm}}} &
		\includegraphics[width=\turnheightnew,keepaspectratio]{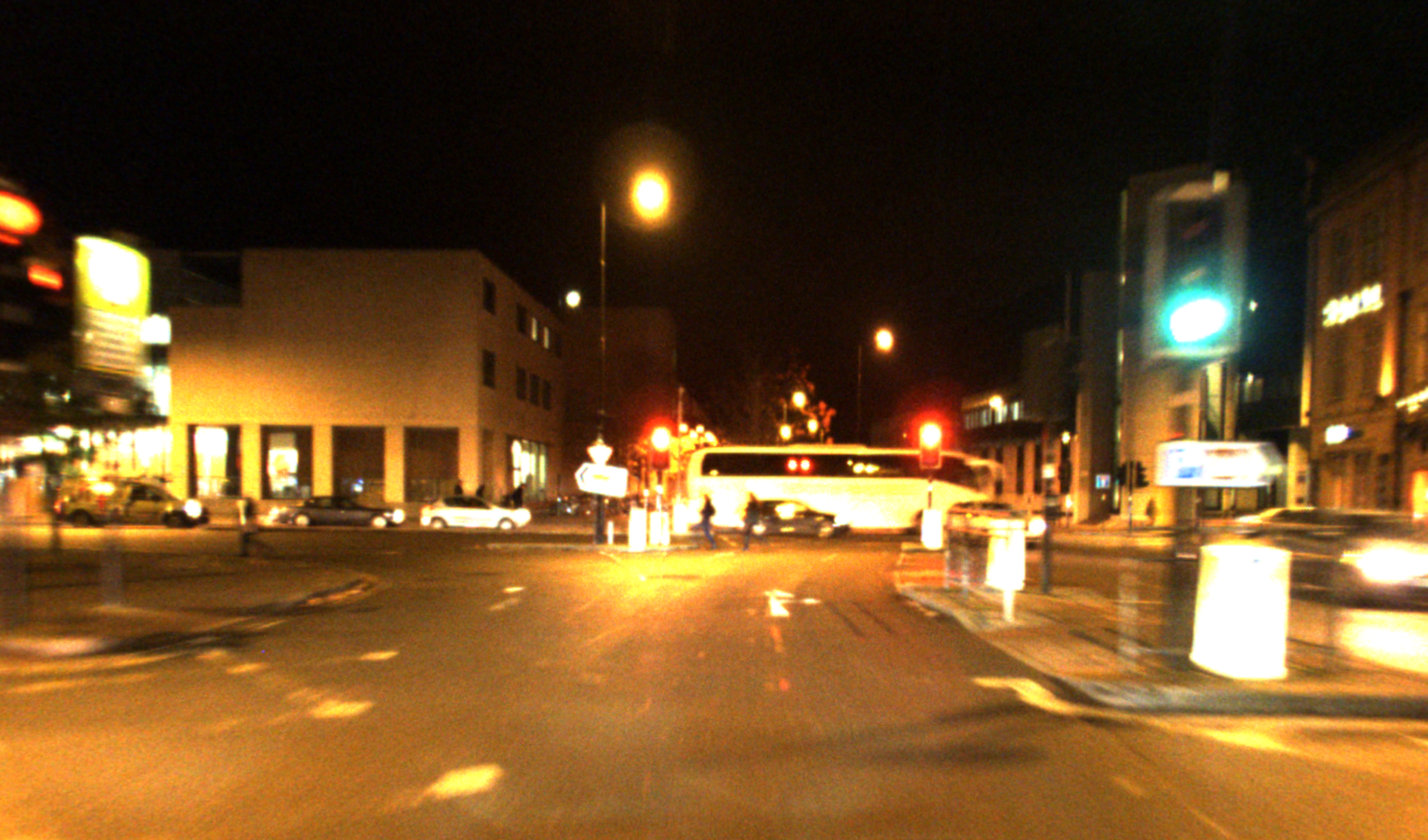} &
		\includegraphics[width=\turnheightnew,keepaspectratio]{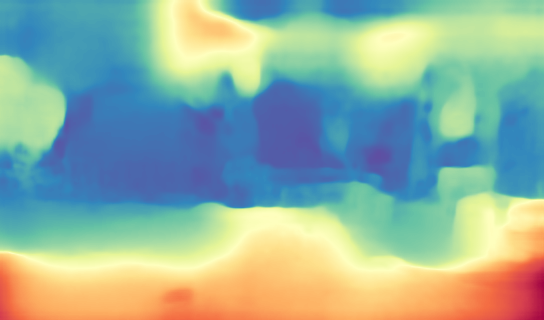} &
		\includegraphics[width=\turnheightnew,keepaspectratio]{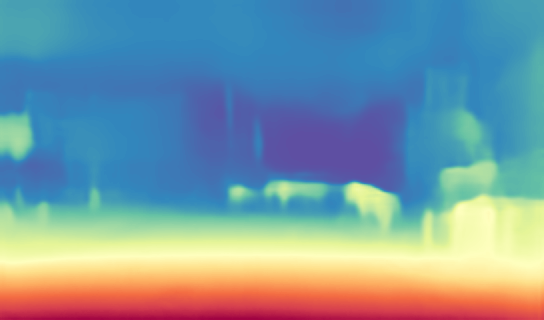} &
		\includegraphics[width=\turnheightnew,keepaspectratio]{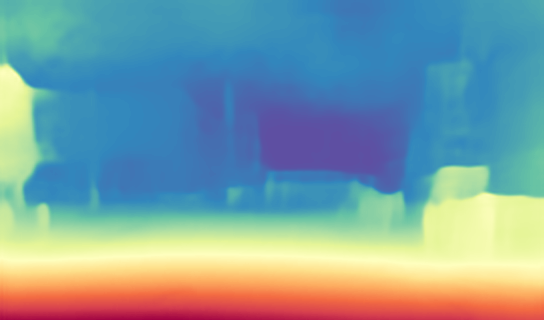} &
		\includegraphics[width=\turnheightnew,keepaspectratio]{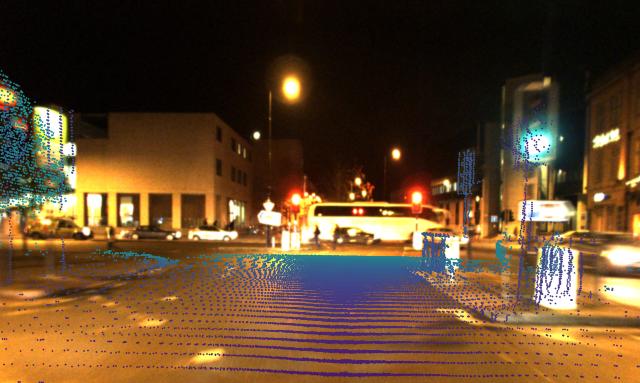} \\
		
		\vspace{-0.5mm}
		{\rotatebox{90}{\hspace{2mm}}} &
		\includegraphics[width=\turnheightnew,keepaspectratio]{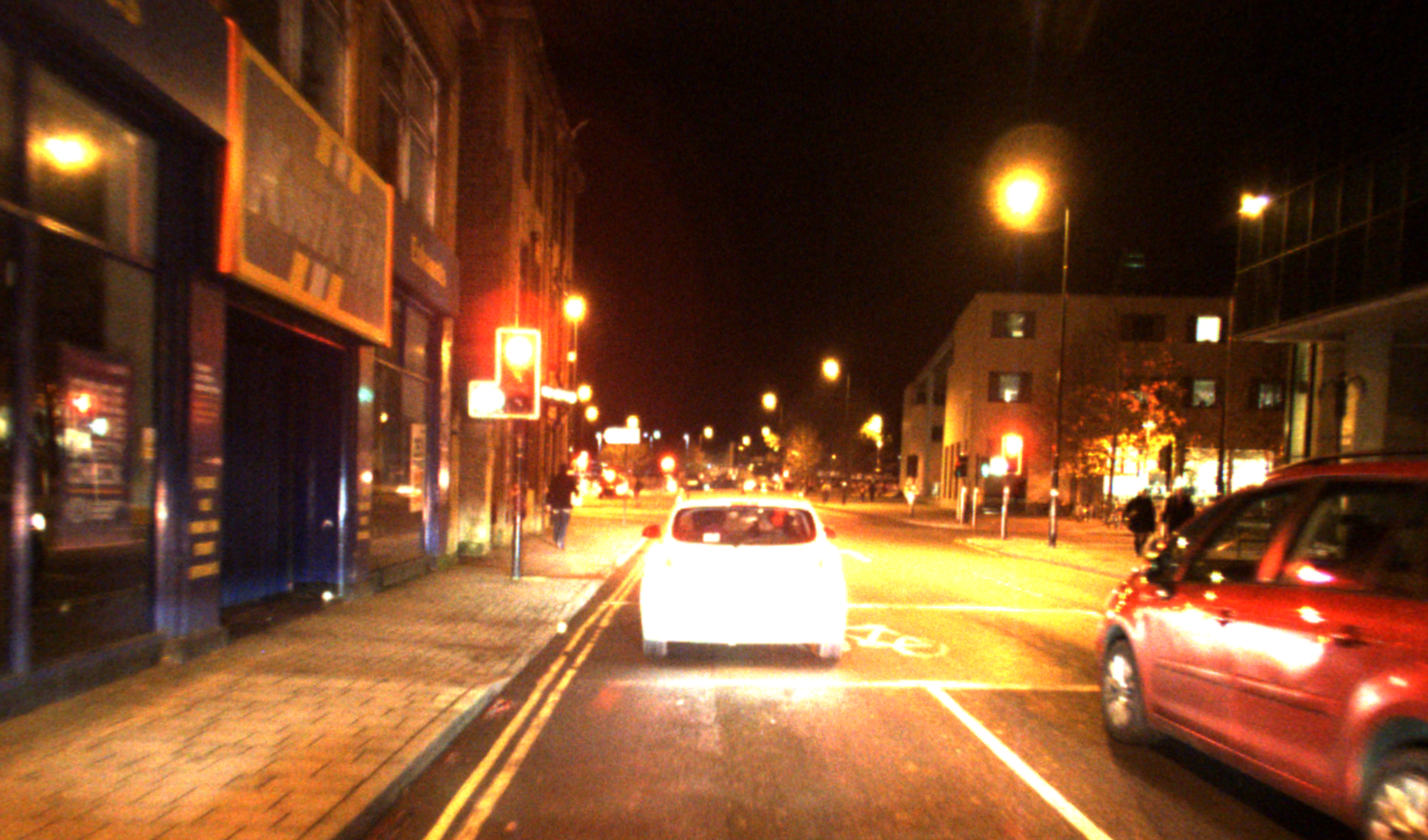} &
		\includegraphics[width=\turnheightnew,keepaspectratio]{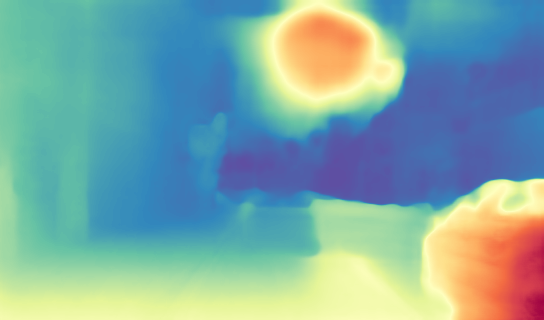} &
		\includegraphics[width=\turnheightnew,keepaspectratio]{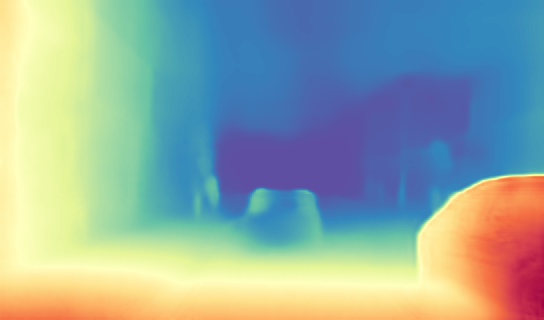} &
		\includegraphics[width=\turnheightnew,keepaspectratio]{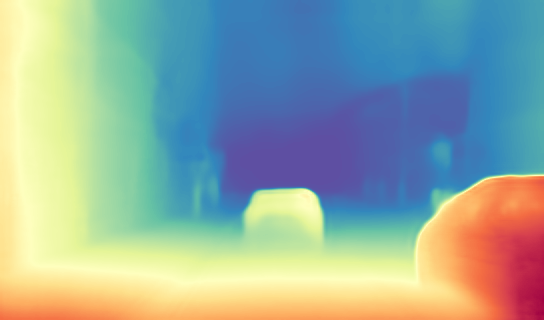} &
		\includegraphics[width=\turnheightnew,keepaspectratio]{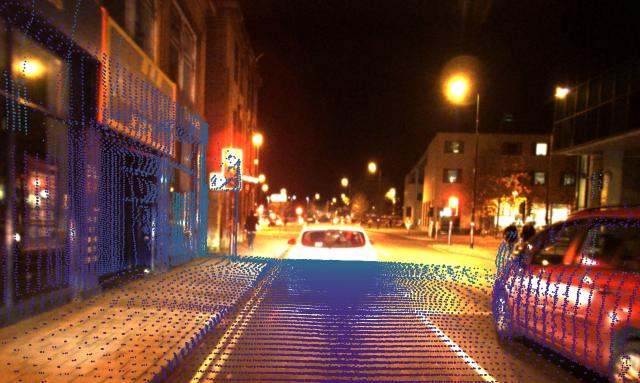} \\
		
		\vspace{-0.5mm}
		{\rotatebox{90}{\hspace{2mm}}} &
		\includegraphics[width=\turnheightnew,keepaspectratio]{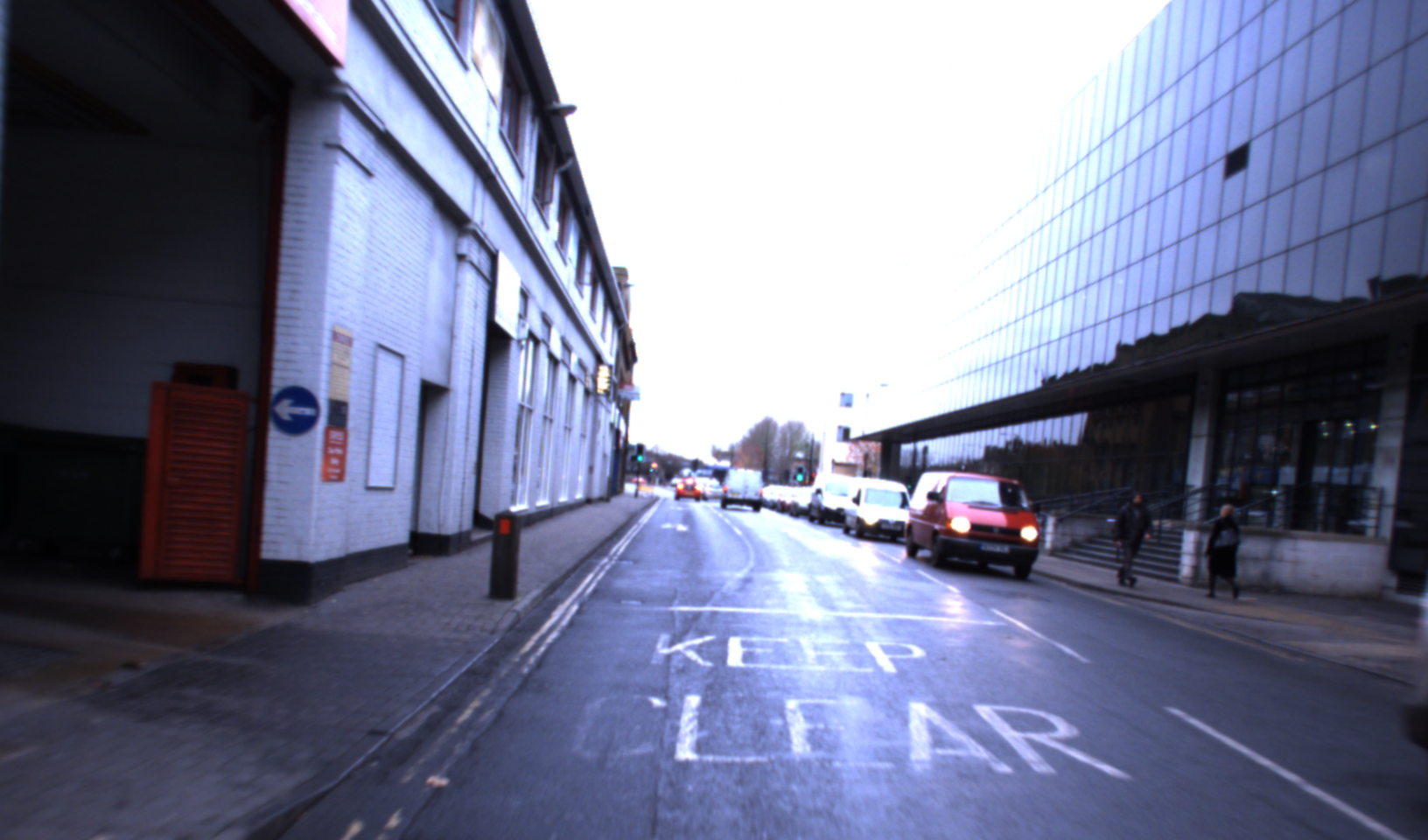} &
		\includegraphics[width=\turnheightnew,keepaspectratio]{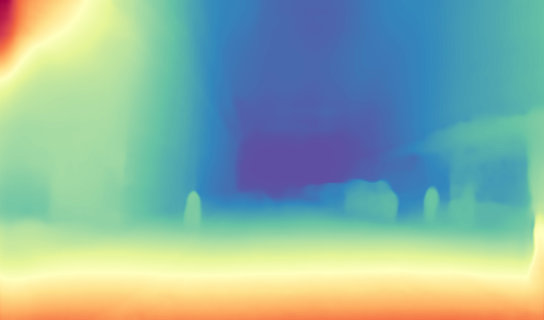} &
		\includegraphics[width=\turnheightnew,keepaspectratio]{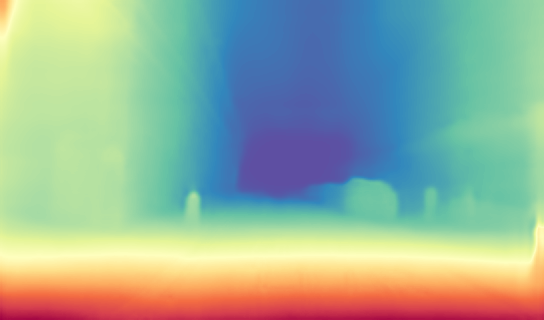} &
		\includegraphics[width=\turnheightnew,keepaspectratio]{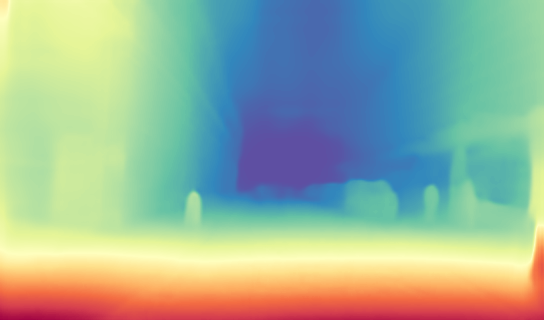} &
		\includegraphics[width=\turnheightnew,keepaspectratio]{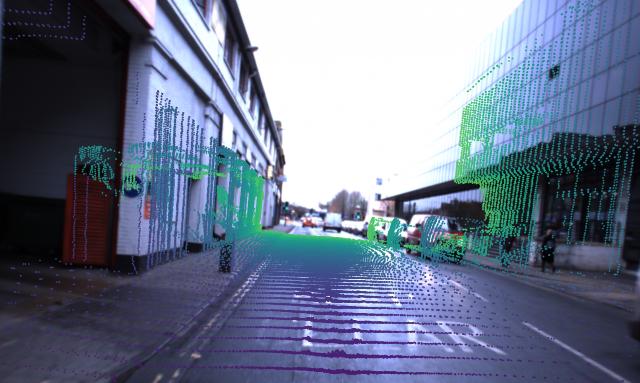} \\
		
		\vspace{-0.5mm}
		{\rotatebox{90}{\hspace{2mm}}} &
		\includegraphics[width=\turnheightnew,keepaspectratio]{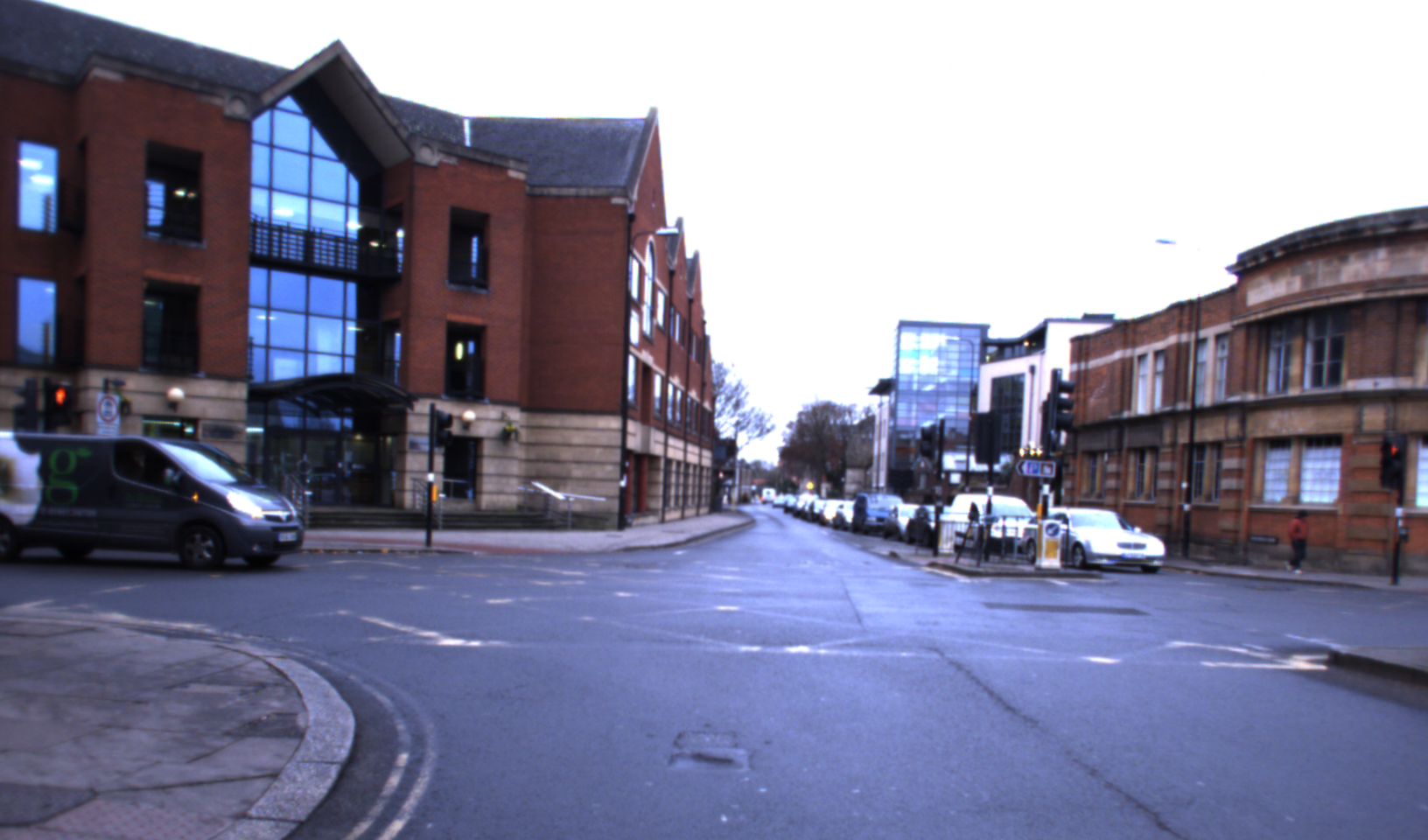} &
		\includegraphics[width=\turnheightnew,keepaspectratio]{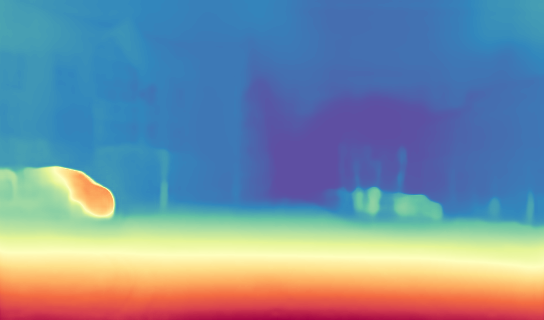} &
		\includegraphics[width=\turnheightnew,keepaspectratio]{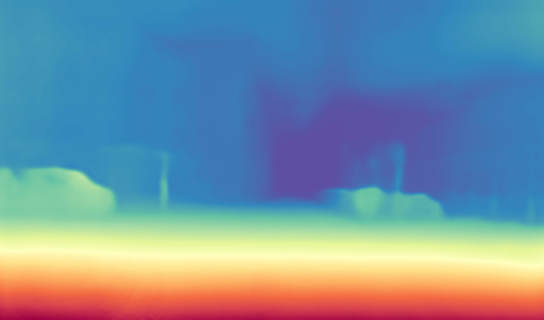} &
		\includegraphics[width=\turnheightnew,keepaspectratio]{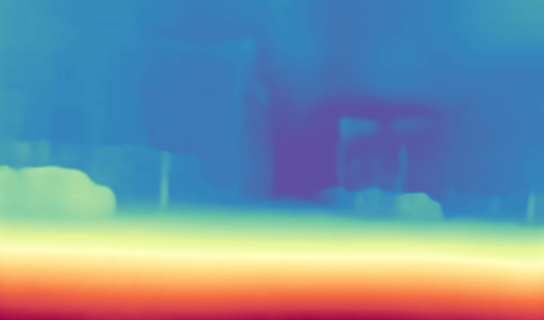} &
		\includegraphics[width=\turnheightnew,keepaspectratio]{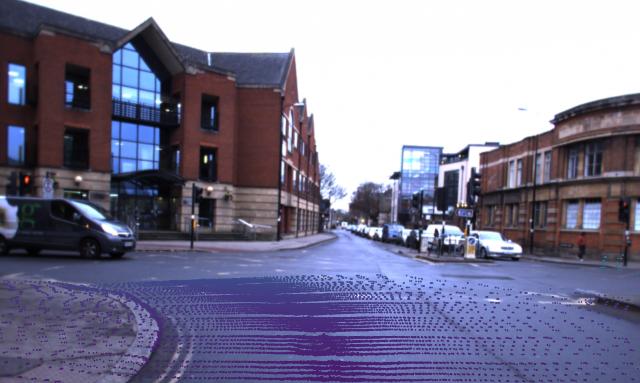} \\
		
		\vspace{-0.5mm}
		{\rotatebox{90}{\hspace{2mm}}} &
		\includegraphics[width=\turnheightnew,keepaspectratio]{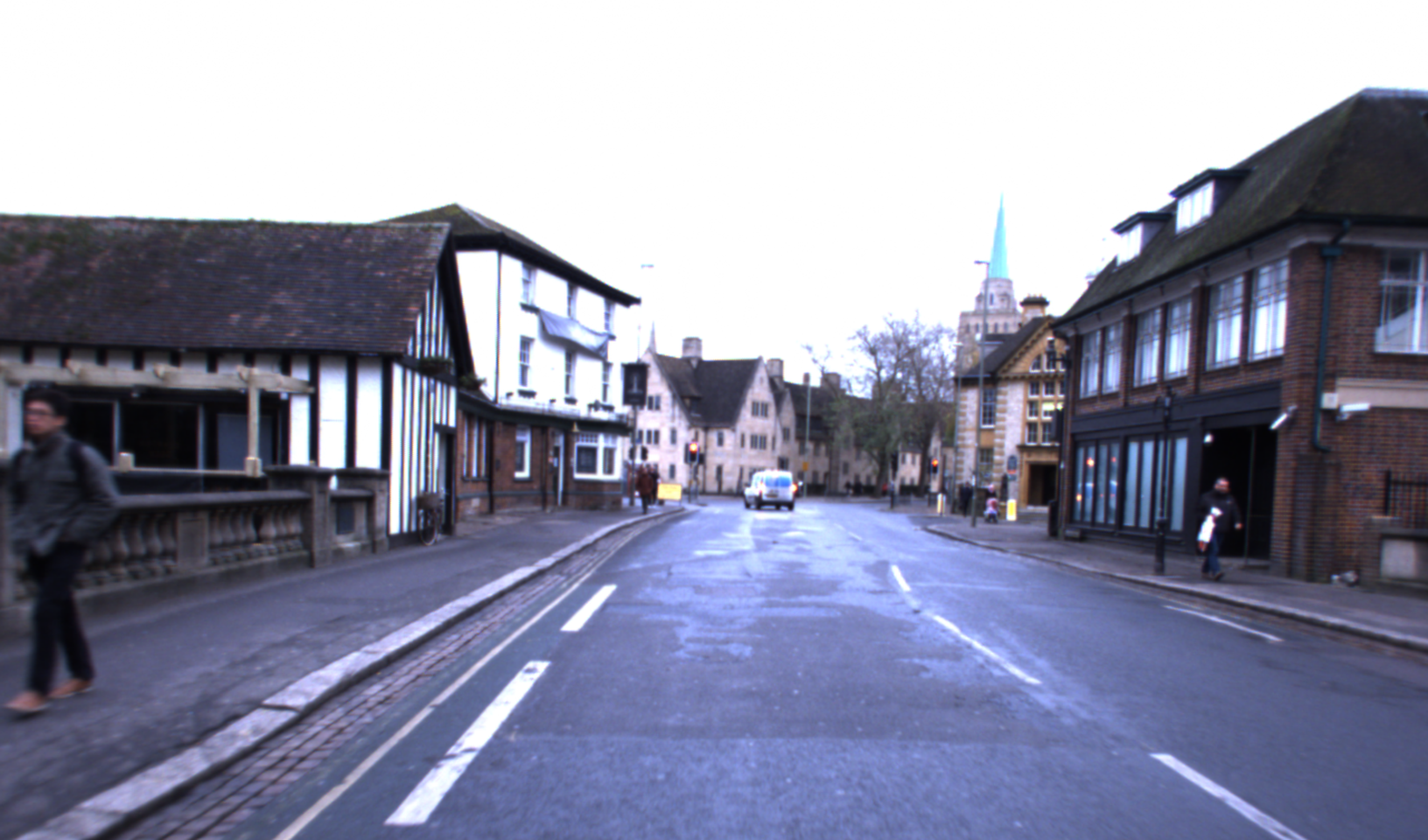} &
		\includegraphics[width=\turnheightnew,keepaspectratio]{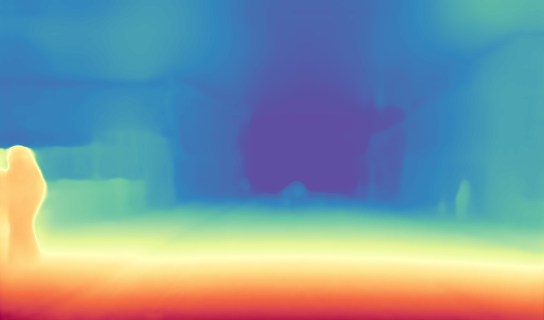} &
		\includegraphics[width=\turnheightnew,keepaspectratio]{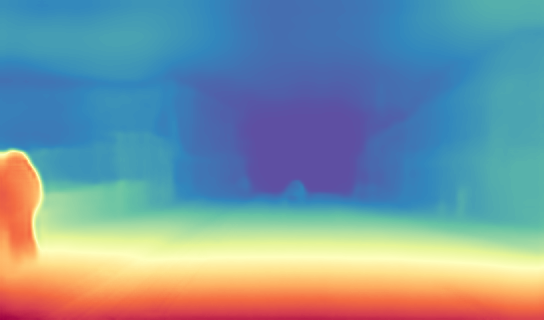} &
		\includegraphics[width=\turnheightnew,keepaspectratio]{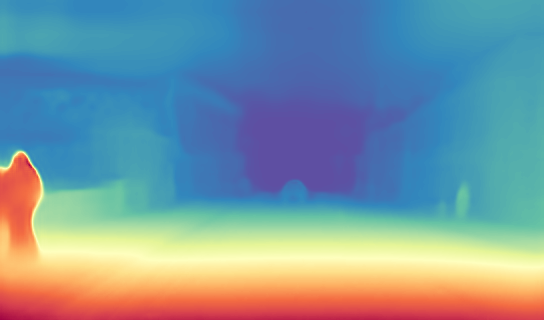} &
		\includegraphics[width=\turnheightnew,keepaspectratio]{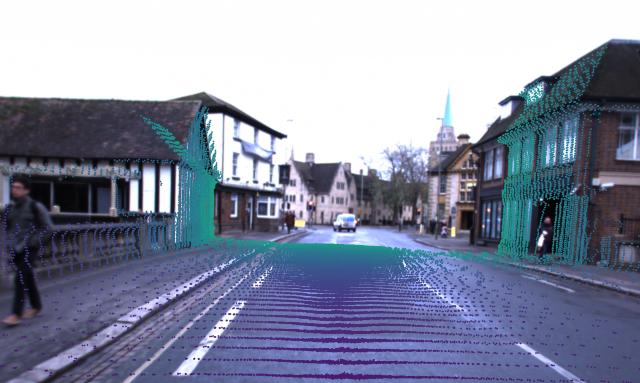} \\

		\multicolumn{1}{c}{} & 
		\multicolumn{1}{c}{Image} & 
		\multicolumn{1}{c}{Monodepth2} & 
		\multicolumn{1}{c}{md4all-DD} & 
		\multicolumn{1}{c}{ACDepth} & 
		\multicolumn{1}{c}{GT} \\
	\end{tabular}
	\caption{Qualitative results on RobotCar~\protect\cite{maddern20171}} 
	\label{fig:9}
\end{figure*}

In this section, we present a more comprehensive quantitative comparison. We selected MonoDepth2~\cite{godard2019digging}, md4all-DD~\cite{gasperini2023robust}, and our method (ACDepth) for the experiments, all of which use the same network backbone. The visual test results for nuScenes~\cite{caesar2020nuscenes} are shown in Fig.~\ref{fig:8}, and those for RobotCar~\cite{maddern20171} are shown in Fig.~\ref{fig:9}.

\end{document}